%% file: main.tex
\documentclass{article}

\PassOptionsToPackage{numbers,compress}{natbib}
\usepackage[eandd,final]{neurips_2026}

\usepackage[utf8]{inputenc}
\usepackage[T1]{fontenc}
\usepackage{hyperref}
\usepackage{url}
\usepackage{booktabs}
\usepackage{multirow}
\usepackage{amsfonts}
\usepackage{amsmath}
\usepackage{amssymb}
\usepackage{nicefrac}
\usepackage{microtype}
\usepackage{enumitem}
\usepackage{xcolor}
\usepackage{graphicx}
\usepackage{float}
\graphicspath{{figures/}}

\title{When Are Multimodal Predictions Biologically Supported? A Diagnostic Evaluation Framework}

\author{%
  Dylan Steiner$^{1,\ast}$,
  Gustavo Arango-Argoty$^{1,\dagger}$,
  Gerald Sun$^{1,\dagger}$,
  and Etai Jacob$^{1,\dagger}$ \\[0.5em]
  $^{1}$Oncology Data Science \& AI, R\&D, AstraZeneca, Boston, US
}

\begin{document}

\maketitle
\renewcommand{\thefootnote}{\fnsymbol{footnote}}
\footnotetext[1]{Contact: \texttt{dylan.steiner@astrazeneca.com}. $^{\dagger}$Equal advising.}
\renewcommand{\thefootnote}{\arabic{footnote}}
\setcounter{footnote}{0}

\begin{abstract}
Multimodal models in oncology can produce accurate predictions, but accurate prediction does not reveal whether the model has learned biology that is shared across modalities, biology confined to one modality, or spurious correlations that reflect confounders rather than genuine biology.
We introduce \textbf{DECAT}, a model-agnostic post-hoc evaluation framework that classifies multimodal representations into four diagnostic scenarios (shared, spurious, modality-specific, or no signal) for a given task and modality, using five null-referenced metrics and a rule-based decision procedure.
The framework operates on learned representations, requires no knowledge of which specific confounder is present, and returns indeterminate when the evidence is insufficient. We evaluate multimodal architecture properties using DECAT on synthetic data across four multimodal model classes spanning over 2{,}500 trained representations, and demonstrate its operational use on 8{,}979 TCGA patients with both multimodal model embeddings and five pretrained pathology foundation models. Entangled models (e.g., CLIP) detect shared signal near-perfectly on synthetic data but also falsely claim it when it is absent, doing so in the majority of such cases on real foundation model embeddings. This false claim rate increases with signal strength and sample size, so that larger cohorts and stronger representations produce more confident but still incorrect diagnoses. Applied to both multimodal TCGA embeddings and five pathology foundation models without paired RNA, DECAT detects confounding invisible to AUROC without requiring the confounder labels, as confirmed by post-hoc stratification.
\end{abstract}

\section{Introduction}
\label{sec:intro}

Multimodal foundation models are increasingly applied in oncology to characterize tumor biology and predict patient outcomes~\cite{chenPancancerIntegrativeHistologygenomic2022,vaidyaMoleculardrivenFoundationModel2025,yunEXAONEPath252025,xuMultimodalKnowledgeenhancedWholeslide2025,liuLeveragingMultimodalFoundation2026}.
These models can produce accurate predictions from a single modality, such as hematoxylin and eosin (H\&E) stained histopathology slides, even when trained with multimodal supervision.
However, accurate prediction does not reveal whether a model has learned biology that is shared across modalities, biology that is confined to a single modality, or spurious correlations that may not generalize.

\begin{figure}[t]
\centering
\includegraphics[width=\linewidth]{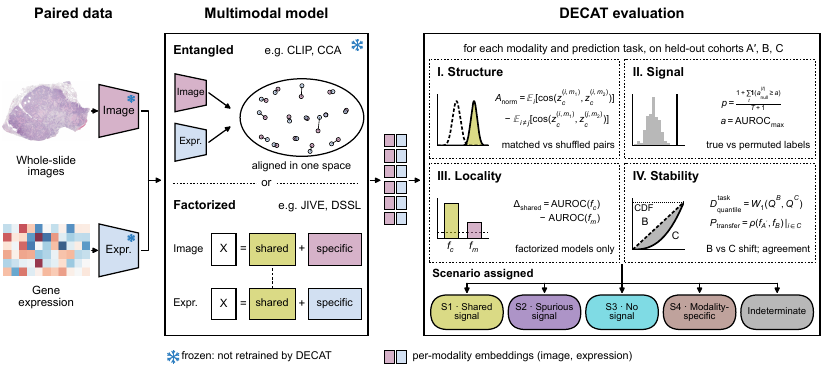}
\caption{\textbf{DECAT framework overview.} DECAT takes per-modality embeddings from any multimodal model and uses a four-stage decision tree to classify the predictive behavior of each modality, for a given task, into one of four scenarios or indeterminate (Figure~\ref{fig:app-decision-tree}, Appendix~\ref{app:decision}). Scenario classification informs downstream decisions: cross-modally supported signal consistent with shared biology (S1) supports cross-modal deployment, modality-specific signal (S4) restricts reliable predictions to the source modality, spurious signal (S2) transfers across cohorts but exhibits composition-dependent instability indicating risk of failure under distribution shift, and no signal (S3) indicates the representation is uninformative for that task.}
\label{fig:overview}
\end{figure}

Histopathology foundation models encode non-biological technical features that persist despite normalization, with better-performing models often encoding stronger features and embeddings clustering more by medical center than by tissue type~\cite{howardImpactSitespecificDigital2021,komenHistopathologicalFoundationModels2024,jongCurrentPathologyFoundation2025,komenRobustFoundationModels2025}. Scanner-induced domain shifts further mask bias behind stable prediction accuracy~\cite{thiringerScannerInducedDomainShifts2026}, and strong dependencies between biomarkers and clinicopathological variables cause models to rely on correlated features rather than the biomarker itself~\cite{dawoodConfoundingFactorsBiases2026}.
Together, these findings establish that predictive performance is a necessary but insufficient criterion for evaluating multimodal biomarker models: a model can be accurate for the wrong reasons.

Tumor biology is inherently multimodal but not uniformly shared across assays~\cite{cuiMultimodalFoundationModels2025}.
Some biological programs, such as immune infiltration, manifest across histology, transcriptomics, and proteomics.
Others are detectable primarily in specific modalities: chromosomal instability and pathway-level signaling states are robustly characterized by genomic and transcriptomic assays but may lack specific morphological correlates in histopathology.
When models are trained with multimodal supervision, predictive signal for such features may originate primarily from the molecular modality, making single-modality predictions from H\&E alone unreliable despite appearing accurate in-distribution.
Integrating complementary assays can improve clinical predictions~\cite{chenPancancerIntegrativeHistologygenomic2022,vanguriMultimodalIntegrationRadiology2022,kilimMultimodalAIBiomarker2026}, though multimodal performance gains may be partially attributable to batch effects~\cite{howardMultimodalDeepLearning2023}, and existing evaluations cannot determine which improvements are driven by shared tumor biology and which by spurious correlations.
An evaluation framework must therefore distinguish, for each modality and prediction task independently, among several distinct scenarios:
\begin{enumerate}
    \item \textbf{Shared signal.} The modality carries predictive signal that is also independently observable in other modalities and transfers stably across cohorts. DECAT treats such signal as consistent with shared biology, not as proof of a shared mechanism.
    \item \textbf{Spurious signal.} The modality carries predictive signal, but it is driven by batch effects, site-specific artifacts, or proxy features. Such predictions may remain accurate when deployment conditions match training, but are at risk of failure under distribution shift.
    \item \textbf{No signal.} The modality's learned representation carries no task-relevant information.
    \item \textbf{Modality-specific signal.} The modality carries stable predictive signal, but this signal is not independently observable in other modalities. Predictions for the same task from other modalities are not supported.
\end{enumerate}
No existing tool provides this diagnostic. Preserved-site cross-validation~\cite{howardImpactSitespecificDigital2021} and covariate stratification~\cite{dawoodConfoundingFactorsBiases2026} detect specific known confounders but require prior knowledge and do not generalize to arbitrary sources of spurious signal. Standard contrastive learning methods~\cite{radfordLearningTransferableVisual2021} conflate shared and modality-specific signal, while factorized learning methods~\cite{wangInformationCriterionControlled2025,liangFactorizedContrastiveLearning2023} separate them in principle but have not been used as diagnostic tools.

We introduce \textbf{DECAT} (\textbf{D}iagnostic \textbf{E}valuation of \textbf{C}ross-modal \textbf{A}lignment and \textbf{T}ransfer), a model-agnostic post-hoc evaluation framework that classifies the predictive behavior of multimodal representations, for a given task and modality, into the four scenarios above, without requiring knowledge of which specific confounder or artifact is present.
Rather than solely evaluating predictive accuracy on held-out data, DECAT determines whether predictive signal is shared across modalities, confined to a single modality, or driven by spurious correlations that may not generalize.
The framework operates entirely on learned representations, does not modify or retrain models, and applies to any multimodal method that produces per-modality embeddings, whether entangled (e.g., CCA~\cite{guoCanonicalCorrelationAnalysis2021}, CLIP-style contrastive learning~\cite{radfordLearningTransferableVisual2021}) or factorized (e.g., JIVE~\cite{lockJointIndividualVariation2013}, DisentangledSSL~\cite{wangInformationCriterionControlled2025}), as well as to unimodal foundation models via a subset of its metrics.
An overview of the framework is shown in Figure~\ref{fig:overview}.
We validate DECAT on synthetic data with controlled ground truth across four model classes (over 2{,}500 trained representations, nearly 600{,}000 scenario evaluations) and on real data from 8{,}979 TCGA patients, evaluating both multimodal embeddings and five pretrained pathology foundation models.
Our contributions are: (1)~a model-agnostic diagnostic framework that classifies multimodal representations into four scenarios using only learned representations, returning \emph{indeterminate} when evidence is insufficient; (2)~five null-referenced metrics that together distinguish cross-modally supported signal from spurious signal, modality-specific signal, and noise; (3)~a controlled synthetic validation demonstrating that no single architecture reliably diagnoses all scenarios; and (4)~real-data validation showing that DECAT detects cancer-type confounding invisible to AUROC on both multimodal and unimodal foundation model embeddings without requiring confounder labels. Extended related work is in Appendix~\ref{app:related}.
\section{Framework}
\label{sec:framework}

\subsection{Problem Setting}
\label{sec:problem}

We use modality to refer to an independent biological measurement of the same entity (e.g., histopathology and transcriptomics from the same tumor), as distinct from different data types that do not independently measure biology (e.g., clinical notes and billing codes).
Consider a multimodal setting with two observed modalities $x_h \in \mathbb{R}^{d_h}$ and $x_r \in \mathbb{R}^{d_r}$ (e.g., histopathology and transcriptomics), generated from unobserved latent variables representing shared biology ($z_s$), modality-specific biology ($z_h, z_r$), and batch effects ($b$), all in $\mathbb{R}^k$.
A binary outcome $y \in \{0,1\}$ depends on a subset of these latent variables through a score, with the source of predictive signal controlled by outcome coefficients $\alpha_s, \alpha_h, \alpha_r, \alpha_b$.
A representation model is trained on paired observations from a training cohort and frozen. The frozen encoder maps observations from held-out validation cohorts into a learned representation space. DECAT operates entirely on these learned representations and never accesses ground-truth latent variables, outcome labels during representation learning, or simulator parameters. The goal is to classify the learned representation, for a given outcome task and modality, into one of the four scenarios defined in Section~\ref{sec:intro} (S1--S4) or as indeterminate when the evidence is insufficient.

\subsection{Synthetic Data Generator}
\label{sec:simulator}

We construct a minimal simulator that generates paired multimodal observations with known ground-truth scenarios (Appendix~\ref{app:simulator}).
Each modality is a linear mixture of independent Gaussian latent variables representing shared biology ($z_s$), modality-specific biology ($z_h, z_r$), and batch effects ($b$), with coefficients $\beta$ controlling relative signal contributions and Gaussian noise calibrated to a fixed signal-to-noise ratio.
Four disjoint cohorts are generated per run: Cohort~A (representation training), Cohort~A$'$ (within-distribution evaluation), and Cohorts~B and~C (external, with controlled batch-axis shifts of differing magnitude).

Binary outcome labels are generated from a latent score with coefficients $\alpha_s, \alpha_h, \alpha_r, \alpha_b$ controlling the fraction of task label variance attributable to each source.
Setting different $\alpha$ coefficients to zero instantiates each scenario: Scenario~1 ($\alpha_s > 0$ only), Scenario~2 ($\alpha_b > 0$, termed S2B for direct confounding, or proxy $\gamma > 0$, termed S2A/C/D depending on proxy geometry; Appendix~\ref{app:simulator:proxy}), Scenario~3 (all $\alpha = 0$), Scenario~4 ($\alpha_h > 0$ or $\alpha_r > 0$ only).
Crucially, representation learning uses only Cohort~A and is outcome-agnostic. Outcome labels are introduced only during evaluation, enabling many biological scenarios to be evaluated per trained representation without retraining.

\subsection{Representation Models}
\label{sec:models}

We evaluate four model classes spanning a two-by-two grid (Table~\ref{tab:models}): \emph{entangled} models (CCA~\cite{guoCanonicalCorrelationAnalysis2021}, CLIP~\cite{radfordLearningTransferableVisual2021}) produce a single embedding per modality without distinguishing shared from modality-specific contributions, while \emph{factorized} models (a simplified JIVE-style decomposition~\cite{lockJointIndividualVariation2013}, DisentangledSSL~\cite{wangInformationCriterionControlled2025}) explicitly decompose representations into shared and modality-specific components.
All models are trained on Cohort~A only without outcome labels and frozen before evaluation. DisentangledSSL is evaluated at four disentanglement strengths spanning weak to strong information bottleneck pressure (Appendix~\ref{app:models}).
\begin{table}[h]
\centering
\caption{\textbf{Model classes.} Entangled models produce a single unified embedding per modality. Factorized models produce separate shared ($Z_c$) and modality-specific ($Z_s$) components.}
\label{tab:models}
\scriptsize
\begin{tabular}{lcc}
\toprule
 & \textbf{Linear} & \textbf{Nonlinear} \\
\midrule
\textbf{Entangled}   & CCA  & CLIP \\
\textbf{Factorized}  & JIVE & DisentangledSSL \\
\bottomrule
\end{tabular}
\end{table}

\subsection{Interpretability Metrics}
\label{sec:metrics}

DECAT evaluates representations through five null-referenced metrics (Table~\ref{tab:metrics}).
All comparisons are calibrated against permutation null distributions and no hard thresholds are imposed. Metrics are computed on held-out cohorts (A$'$, B, C) using outcome labels generated independently of representation learning. No single metric is sufficient for scenario classification. Full metric definitions are in Appendix~\ref{app:metrics}.

\begin{table}[t]
\centering
\caption{\textbf{DECAT evaluation metrics.} Structural metrics ($A_{\text{norm}}$, $B_{\text{norm}}$) validate representation geometry but cannot distinguish biology from spurious signal. Outcome-directed metrics ($\Delta_{\text{shared}}$, $P_{\text{transfer}}$, $D_{\text{quantile}}^{\text{task}}$) are required for scenario discrimination. Full definitions in Appendix~\ref{app:metrics}.}
\label{tab:metrics}
\scriptsize
\begin{tabular*}{\textwidth}{@{\extracolsep{\fill}}p{1.3cm}p{4.0cm}p{2.2cm}p{4.5cm}}
\toprule
\textbf{Metric} & \textbf{What it measures} & \textbf{Calibration} & \textbf{Role in scenario assignment} \\
\midrule
$A_{\text{norm}}$ & Cross-modal shared latent agreement & Permutation (sample identity) & Required for S1; guards shared-dominant localization \\
\addlinespace
$B_{\text{norm}}$ & Prototype consistency across modalities & Permutation (sample identity) & Reported but not used in scenario assignment \\
\addlinespace
$\Delta_{\text{shared}}$ & Predictive signal localization & Permutation (labels) + bootstrap CI & Distinguishes S1 from S4. Factorized models only. \\
\addlinespace
$P_{\text{transfer}}$ & Cross-cohort predictive agreement & Permutation (labels) + bootstrap CI & Necessary for S1, S2, S4 (S2 signal transfers but is composition-unstable); insufficient alone to distinguish S1 from S2 \\
\addlinespace
$D_{\text{quantile}}^{\text{task}}$ & Outcome-aligned cohort stability & Permutation (cohort B/C labels) & Primary S2 discriminator \\
\bottomrule
\end{tabular*}
\end{table}

\subsection{Decision Procedure}
\label{sec:decision}

DECAT applies a fixed, deterministic, four-stage decision procedure (Figure~\ref{fig:app-decision-tree}) with no learned parameters.
Stage~I checks whether the representation exhibits cross-modal shared structure ($A_{\text{norm}}$, $B_{\text{norm}}$).
Stage~II gates on signal presence via a label-permutation test with failure routing to Scenario~3.
Stage~III localizes signal via $\Delta_{\text{shared}}$ (factorized models only, skipped for entangled models).
Stage~IV evaluates cross-cohort stability. $D_{\text{quantile}}^{\text{task}}$ exceeding its null triggers Scenario~2 regardless of localization, because confounding can appear shared or modality-specific. Checking cross-cohort stability before signal localization is a conservative design choice that prioritizes Scenario~2 detection over Scenario~1 and Scenario~4 sensitivity so that false shared claims are minimized at the cost of routing some ambiguous cases to indeterminate.
Among stable representations, Scenario~1 requires shared-dominant localization and significant $P_{\text{transfer}}$, while  Scenario~4 requires modality-dominant localization with significant $P_{\text{transfer}}$.
Unresolved cases return \emph{indeterminate}.
Complete decision rules are in Appendix~\ref{app:decision}.

\section{Experimental Design}
\label{sec:experiments}

On synthetic data, representation learning and scenario evaluation are decoupled so that models are trained once on Cohort~A without labels and then evaluated across many prediction tasks defined by outcome coefficients that place signal in shared, modality-specific, or batch latents.
We evaluate all four model classes (Table~\ref{tab:models}) across four measurement regimes that vary the relative strength of shared, modality-specific, and batch signal (Appendix~\ref{app:simulator:measurement}), eight proxy configurations spanning weak to strong proxy entanglement with varying alignment geometry (Appendix~\ref{app:experiments:params}), and 48 prediction task configurations covering pure scenarios, mixed-modality scenarios, and transition regimes where identification is not possible (Appendix~\ref{app:experiments:outcomes}).
Training sample sizes and epoch counts were determined by pre-experiment calibration studies (Appendix~\ref{app:experiments:calibration}).
The main experiment comprises 365 synthetic runs, 2{,}555 trained representations, and 589{,}050 scenario evaluations across five evaluation sample sizes.

We additionally validate DECAT on real data using paired histopathology and transcriptomic foundation model embeddings from 8{,}979 TCGA patients spanning 32 cancer types (Section~\ref{sec:tcga}; Appendix~\ref{app:tcga}). H\&E embeddings were obtained from TITAN~\cite{dingMultimodalWholeslideFoundation2025} and RNA-seq embeddings from the Clinical Transformer~\cite{arango-argotyPretrainedTransformersApplied2025}. All four model classes were trained on Cohort~A ($N{=}4{,}938$) with latent dimensionality $K{=}50$. We evaluated two cohort structures: pooled random splits, in which patients are randomly assigned to evaluation cohorts regardless of cancer type (50 seeds, $N{\approx}1{,}347$ per cohort), and extreme-C splits, in which Cohort~C is drawn from patients with extreme values of the prediction target while A$'$/B retain the natural pan-cancer composition. The extreme pool requires no confounder labels: for continuous labels it consists of patients above the 75th percentile; for binary labels it is defined by the model's own probe score (top-10\% of evaluation patients). Task labels for the pooled evaluation include CCA variates (S1 by construction), JIVE RNA-residual PCs (S4), and random binary labels (S3). All continuous labels are binarized at the pan-cancer median (Appendix~\ref{app:tcga:labels}).
We characterize confound detection through an $\alpha$-mixture sweep that varies Cohort~C composition from unselected to fully extreme (Appendix~\ref{app:tcga:alpha}), a power curve measuring minimum cohort size for reliable detection (Appendix~\ref{app:tcga:power}), and a unimodal comparison applying DECAT Stages~II and IV directly to five H\&E-only foundation models for predicting the mutation status of 16 driver genes (Appendix~\ref{app:tcga:unimodal}).

\section{Results}
\label{sec:results}

\subsection{Synthetic Validation}
\label{sec:synthetic-results}

All synthetic results are reported on the baseline measurement regime at $N_{\text{eval}} = 1000$ unless otherwise noted. Detection rates degrade at smaller evaluation sizes in this equal-cohort setting (Appendix~\ref{app:further-results:power}). TCGA cohort size requirements are discussed in Section~\ref{sec:discussion}. Measurement regime comparisons are in Appendix~\ref{app:further-results}.

\begin{figure}[t]
\centering
\includegraphics[width=\linewidth]{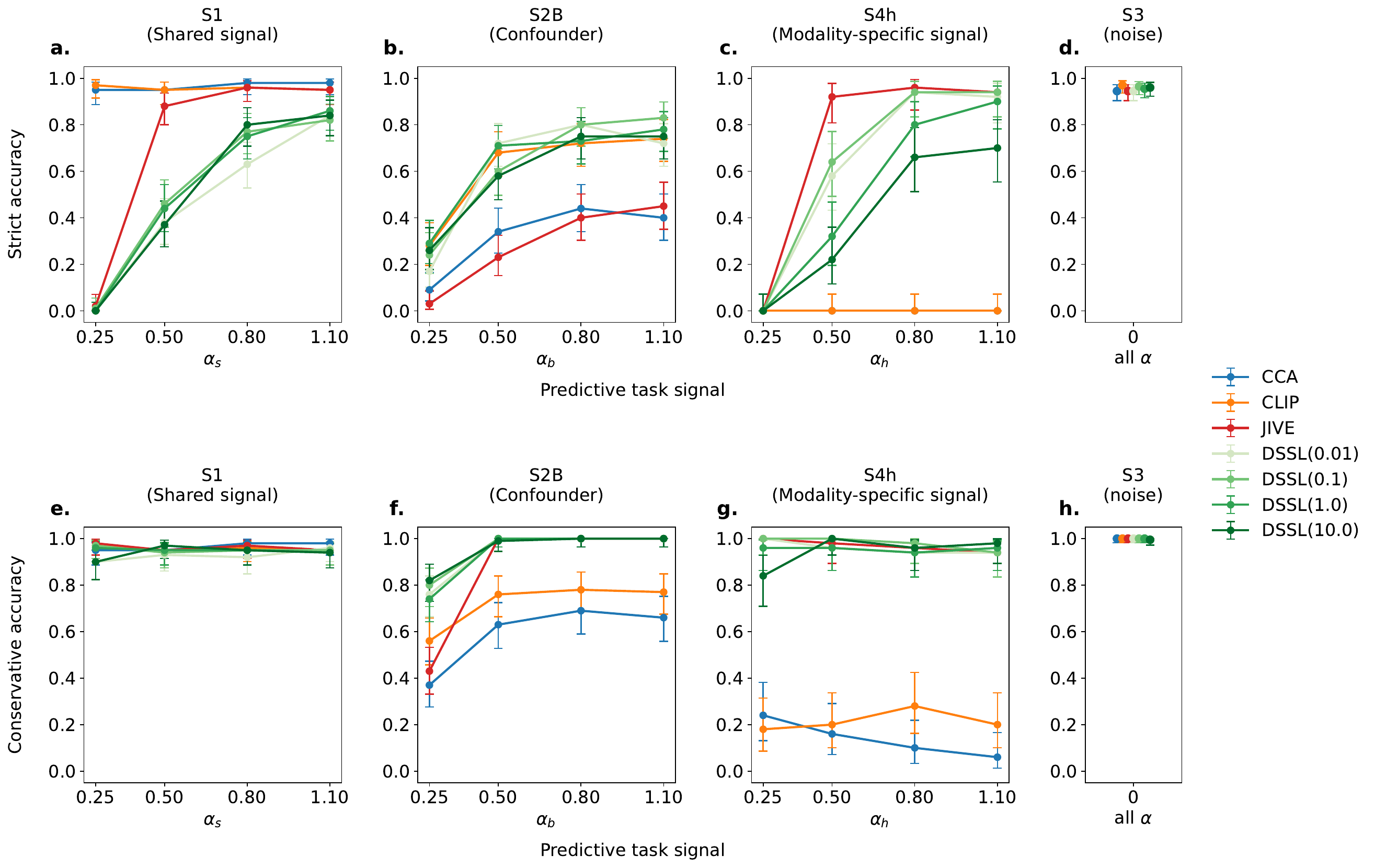}
\caption{\textbf{DECAT detection rate versus predictive task signal per model on synthetic ground truth.} Each column varies the task coefficient for one signal source while all others are zero, producing a pure scenario (S3 has all coefficients at zero). Top row: strict accuracy (correct scenario assigned). Bottom row: conservative accuracy (correct or indeterminate). Where conservative accuracy substantially exceeds strict accuracy, the framework is returning indeterminate rather than misclassifying.
$\alpha \in \{0.25, 0.50, 0.80, 1.10\}$ corresponds to approximately 20\%, 50\%, 72\%, 83\% of task label variance from the respective signal source. S1: Scenario~1, S2B: Scenario~2 (direct confounding), S4h: Scenario~4 (H\&E-specific latent), S3: Scenario~3. Baseline regime, $N_{\text{eval}} = 1000$, 95\% Clopper--Pearson confidence intervals. Each point aggregates 100 evaluations (50 runs $\times$ 2 modalities) for S1, S2B, and S3, and 50 evaluations (50 runs $\times$ H\&E modality only) for S4h.}
\label{fig:alpha_curves}
\end{figure}

\paragraph{Diagnostic accuracy is model-dependent.}
Figure~\ref{fig:alpha_curves} shows detection rates stratified by signal strength and model, reported as both strict accuracy (correct scenario assigned; top row) and conservative accuracy (correct or indeterminate, measuring how often the framework avoids misclassification; bottom row).
S3 (no signal) is identified at 94--97\% strict accuracy across all models (Figure~\ref{fig:alpha_curves}d), confirming that the permutation nulls are well-calibrated.
For S1 (shared signal), entangled models (CCA, CLIP) achieve 95--98\% strict accuracy across all signal strengths (Figure~\ref{fig:alpha_curves}a).
Factorized models have lower strict accuracy due to the additional localization requirement. At $\alpha_s \geq 0.5$, JIVE reaches 88--96\% and DSSL variants reach 37--86\%, with the spread driven by disentanglement strength. Conservative accuracy remains above 90\% for all factorized models (Figure~\ref{fig:alpha_curves}e), indicating that errors are predominantly indeterminate rather than misclassifications.
For S2B (direct confounding), CLIP and DSSL variants reach 58--83\% strict accuracy at $\alpha_b \geq 0.5$, while CCA and JIVE lag at 23--45\% (Figure~\ref{fig:alpha_curves}b). Conservative accuracy reaches 99--100\% for all factorized models (Figure~\ref{fig:alpha_curves}f), meaning the framework correctly avoids misclassifying confounded signal as shared signal even when it cannot commit to a confident S2 label.
For S4, JIVE reaches 92--96\% strict accuracy at $\alpha_h \geq 0.5$ with conservative accuracy near ceiling (Figure~\ref{fig:alpha_curves}c,g); DSSL variants range more widely (22--94\%) depending on disentanglement strength.
Entangled models (CCA, CLIP) achieve exactly zero strict S4 detection at all signal strengths as expected (Figure~\ref{fig:alpha_curves}c). Because entangled representations do not expose separate shared and modality-specific components, the localization step is unavailable and all detected signal is attributed to shared structure by default. Their low conservative accuracy (6--28\%, Figure~\ref{fig:alpha_curves}g) confirms that these are active misclassifications, not indeterminate hedges.
At weak signal ($\alpha = 0.25$, ${\approx}20\%$ outcome variance), strict detection drops substantially for all scenarios, reflecting a statistical power limitation rather than a framework design issue.

\begin{figure}[t]
\centering
\includegraphics[width=\linewidth]{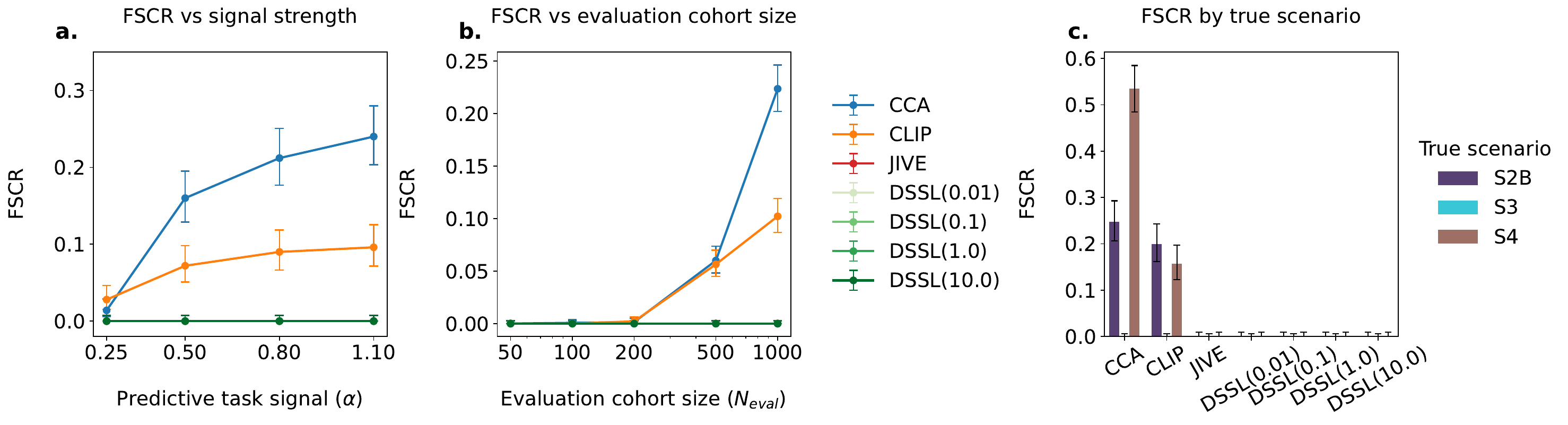}
\caption{\textbf{False shared claim rate (FSCR) on synthetic ground truth.} FSCR is the probability that DECAT assigns Scenario~1 (shared signal) when the true scenario is not S1, pooled across all non-S1 scenarios (S2B, S3, S4h, S4r). (a)~FSCR versus predictive task signal at $N_{\text{eval}} = 1000$, with $\alpha$ matched by magnitude across non-S1 scenarios. FSCR rises with signal strength for entangled models. (b)~FSCR versus evaluation sample size, pooled across all $\alpha$ values and non-S1 scenarios. (c)~FSCR decomposed by true ground-truth scenario at $N_{\text{eval}} = 1000$, pooled across all $\alpha$ values. Entangled models' FSCR is dominated by S4 to S1 misclassification (brown). Factorized models maintain near-zero FSCR across all conditions. Baseline regime, no proxy, 95\% Clopper--Pearson confidence intervals. Each point in (a) aggregates 500 evaluations per model (pooled across runs, modalities, and non-S1 scenarios at each $\alpha$). Each point in (b) aggregates 1{,}400 evaluations (pooled across all $\alpha$ values). Each bar in (c) aggregates 400 evaluations for S2B and S4, and 600 for S3.}
\label{fig:fscr}
\end{figure}

\paragraph{Entangled models are unsafe for shared-signal claims.}
The false shared claim rate (FSCR) measures how often DECAT incorrectly assigns S1 when the ground truth is not S1 (Figure~\ref{fig:fscr}).
FSCR increases with both signal strength and sample size for entangled models. CCA rises from 1\% at $\alpha = 0.25$ to 24\% at $\alpha = 1.1$ (Figure~\ref{fig:fscr}a), and from near zero at $N_{\text{eval}} = 50$ to 22\% at $N_{\text{eval}} = 1000$ (Figure~\ref{fig:fscr}b). Stronger signal and more data both provide more power to detect predictive structure, but entangled representations cannot distinguish whether that structure is shared, modality-specific, or confounded.
Decomposing FSCR by ground-truth scenario reveals that both entangled models misclassify confounded signal as shared signal at substantial rates (CCA 25\%, CLIP 20\% of S2B cases; Figure~\ref{fig:fscr}c). Confounded signal satisfies the S1 criteria for entangled models (cross-modal agreement, predictive transfer) whenever $D_{\text{quantile}}^{\text{task}}$ fails to flag it as S2, and entangled models have no secondary localization check to prevent this misclassification.
All factorized models maintain FSCR near zero across all sample sizes and ground-truth scenarios. The $\Delta_{\text{shared}}$ localization gate prevents false shared claims not by correctly routing confounded signal to S2, but by introducing sufficient uncertainty to return indeterminate rather than a confident S1 assignment.

\paragraph{The framework hedges under uncertainty and is robust to proxy contamination.}
Where conservative accuracy substantially exceeds strict accuracy in Figure~\ref{fig:alpha_curves}, the framework is returning indeterminate rather than misclassifying. For S2B at strong signal, factorized models reach 99--100\% conservative accuracy while strict accuracy remains at 45--83\%. On fundamentally non-identifiable configurations, factorized models return indeterminate in 90--95\% of cases while entangled models commit to confident but incorrect assignments in over 90\% of cases (Figure~\ref{fig:app-transition}).
On proxy-contaminated representations, S1 detection is largely preserved (drops ${<}5\%$, conservative accuracy ${>}90\%$; Figure~\ref{fig:proxy_summary}a,d), while proxy-driven S2 detection remains difficult at 5--20\% strict accuracy (conservative 60--75\% for factorized models; Figure~\ref{fig:proxy_summary}b,e). All factorized models maintain near-zero FSCR across all proxy subtypes, while dual-modality proxy induces the highest false S1 rates for entangled models (Figure~\ref{fig:proxy_summary}; Appendix~\ref{app:further-results:proxy}). When multiple asymmetric signal sources co-exist, DECAT reports the dominant mechanism per modality rather than decomposing co-occurring sources (Figure~\ref{fig:app-mixed-modality}). A consolidated sensitivity summary, including an ablation of each metric in the decision procedure, nonlinear observation processes, and cohort shift alignment, is provided in Appendix~\ref{app:further-results:sensitivity}.

\subsection{TCGA Validation}
\label{sec:tcga}

We next applied DECAT to 8{,}979 TCGA patients using H\&E embeddings from TITAN~\cite{dingMultimodalWholeslideFoundation2025} and RNA-seq embeddings from the Clinical Transformer~\cite{arango-argotyPretrainedTransformersApplied2025}, with all four multimodal model classes trained on a pan-cancer training cohort ($N{=}4{,}938$, $K{=}50$).

\paragraph{Calibration controls behave as expected on real embeddings.}
We use surrogate labels derived from the representation models (CCA variates for shared signal, RNA-residual PCs for modality-specific signal) as calibration controls that verify expected behavior on tasks of known structure, not as independent tests. We report independent validation on labels external to the models (TMB, TP53, age, and 16 driver genes) below.
On TCGA, CCA variate~0 (surrogate for shared signal) was classified as S1 at 97--98\% strict accuracy across CCA (positive control), CLIP, and JIVE (Figure~\ref{fig:tcga_detection_curves}). S3 null detection was 94--96\% across all models. RNA-residual PC~1 (surrogate for modality-specific signal) was classified as S4 at 96\% by JIVE (positive control), while entangled models detected S4 at exactly 0\% as expected. Full detection rates across all 50 variates/PCs and DSSL disentanglement strengths are in Figure~\ref{fig:tcga_detection_curves} (Appendix~\ref{app:further-tcga-results}). On S2B tasks, CCA and CLIP misclassified 94\% and 92\% as S1, respectively, confirming that false shared claims from entangled representations are systematic on real foundation model embeddings (Figure~\ref{fig:tcga_fscr}; Appendix~\ref{app:further-tcga-results:fscr}).

\begin{figure}[t]
\centering
\includegraphics[width=0.75\linewidth]{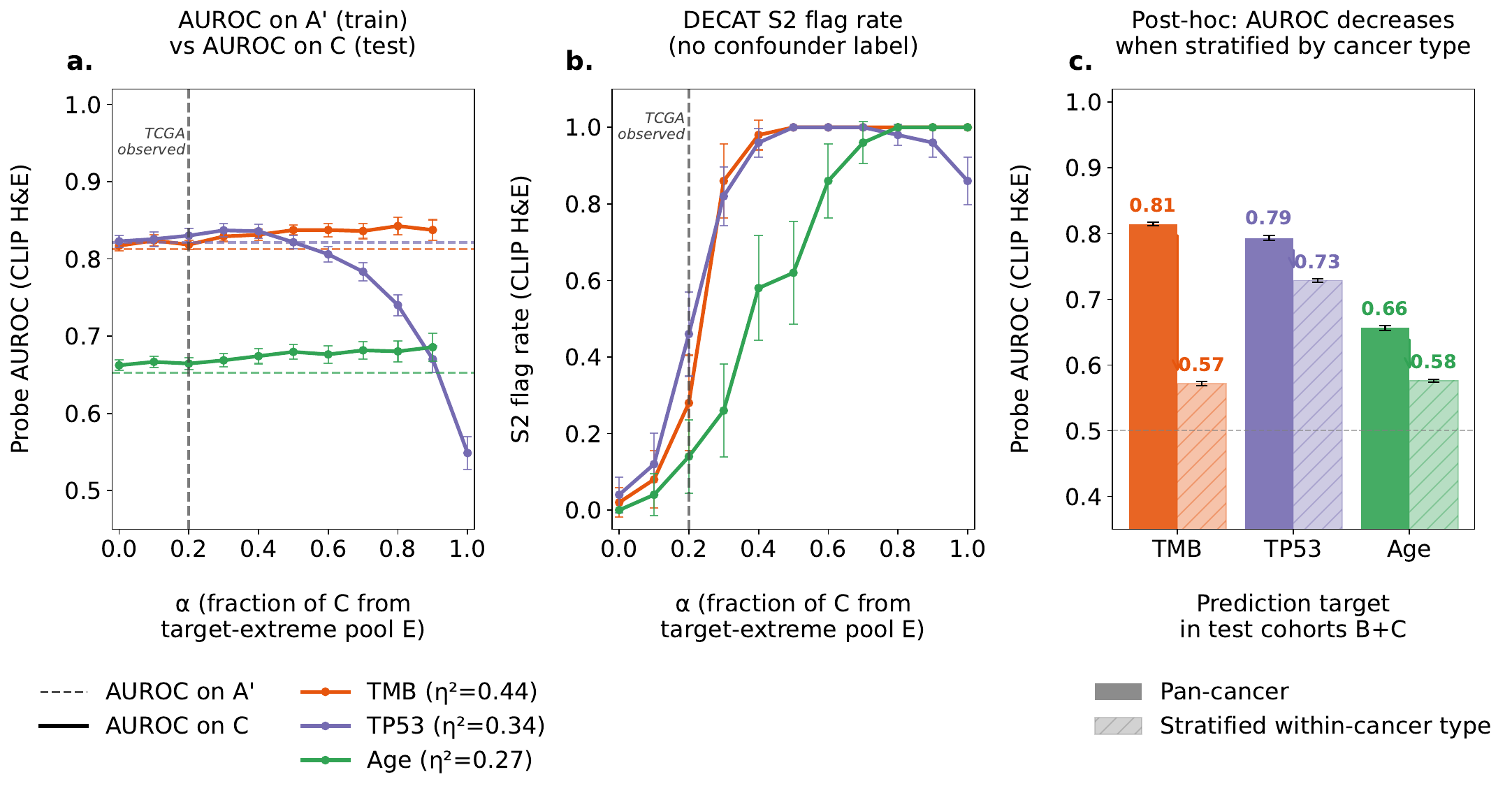}
\caption{\textbf{DECAT detects TCGA cancer-type confounding invisible to AUROC (CLIP, H\&E representation).} The $\alpha$-mixture sweep varies the fraction of Cohort~C drawn from a label-extreme pool while A$'$/B remain random pan-cancer splits; $N_C{\approx}300$ fixed. Dotted line: mean natural cohort composition across labels (${\approx}0.20$; individually TMB ${\approx}0.25$, TP53 ${\approx}0.10$, Age ${\approx}0.25$). TMB and Age are binarized at the pan-cancer median; TP53 is already binary. $\eta^2$ (fraction of label variance explained by cancer type) for each label is shown in the plot legend. 95\% CIs aggregate 50 independent cohort splits per $\alpha$.
(a)~Linear probe AUROC on A$'$ (dashed) versus C (solid). AUROC$_C$ gives false reassurance across $\alpha$. TP53 AUROC$_C$ remains close to training performance until $\alpha{\approx}0.6$, then declines steeply because E is defined by probe score (top-10\% most confident predictions), so at high $\alpha$ most patients in C have similarly high scores, preventing discrimination.
(b)~S2 strict flag rate, requiring no confounder labels. TMB saturates at $\alpha{\approx}0.4$, TP53 at $\alpha{\approx}0.5$, Age at $\alpha{\approx}0.8$.
(c)~Pan-cancer AUROC (solid) versus within-cancer-type AUROC (hatched, pair-weighted across cancer types) on identical patients. The collapse magnitude ($\Delta{=}0.24$ TMB, $0.08$ Age, $0.06$ TP53) reflects how much of the predictive signal is driven by between-cancer-type variation rather than within-type biology. DECAT detects this without cancer-type labels.}
\label{fig:tcga_alpha_mixture_curve}
\end{figure}

\paragraph{DECAT detects confounding invisible to AUROC on biologically confounded labels.}
We evaluated three labels spanning a range of confound strengths with cancer type, quantified by $\eta^2$ (fraction of label variance explained by cancer type): $\log(\text{TMB}{+}1)$ ($\eta^2{=}0.44$), TP53 somatic mutation status ($\eta^2{=}0.34$), and age at diagnosis ($\eta^2{=}0.27$). We sweep $\alpha \in [0,1]$, where $\alpha$ controls the fraction of deployment Cohort~C drawn from a label-extreme pool versus from the natural composition (Figure~\ref{fig:tcga_alpha_mixture_curve}). DECAT's detection of confounding (S2 flag rate) rises with $\alpha$ for all three labels (Figure~\ref{fig:tcga_alpha_mixture_curve}b), saturating at $\alpha{\approx}0.4$ for TMB, $\alpha{\approx}0.5$ for TP53, and $\alpha{\approx}0.8$ for age. Notably, DECAT begins to flag S2 even at $\alpha{\approx}0.20$ (the observed TCGA cancer type composition), where AUROC remains high and gives no warning of confounding (Figure~\ref{fig:tcga_alpha_mixture_curve}a). Post-hoc stratification by cancer type confirms the AUROC collapse that DECAT detected prospectively (Figure~\ref{fig:tcga_alpha_mixture_curve}c). Additional S2 detection results using CCA variates as surrogate labels are in Figure~\ref{fig:tcga_detection_curves}b/f. Because cancer types in TCGA are contributed by different institutions, this confounding is entangled with site effects: stratifying by tissue source site also collapses AUROC (by 0.19--0.24 for TMB and 0.07--0.12 for age across five FMs; Appendix~\ref{app:further-tcga-results:institution}). Within BRCA alone, DECAT flags composition-dependent instability for TP53, CDH1, and ER (S2 in up to 68\% of splits) without access to the candidate confounders (Appendix~\ref{app:further-tcga-results:within-brca}).

\begin{figure}[t]
\centering
\includegraphics[width=\linewidth]{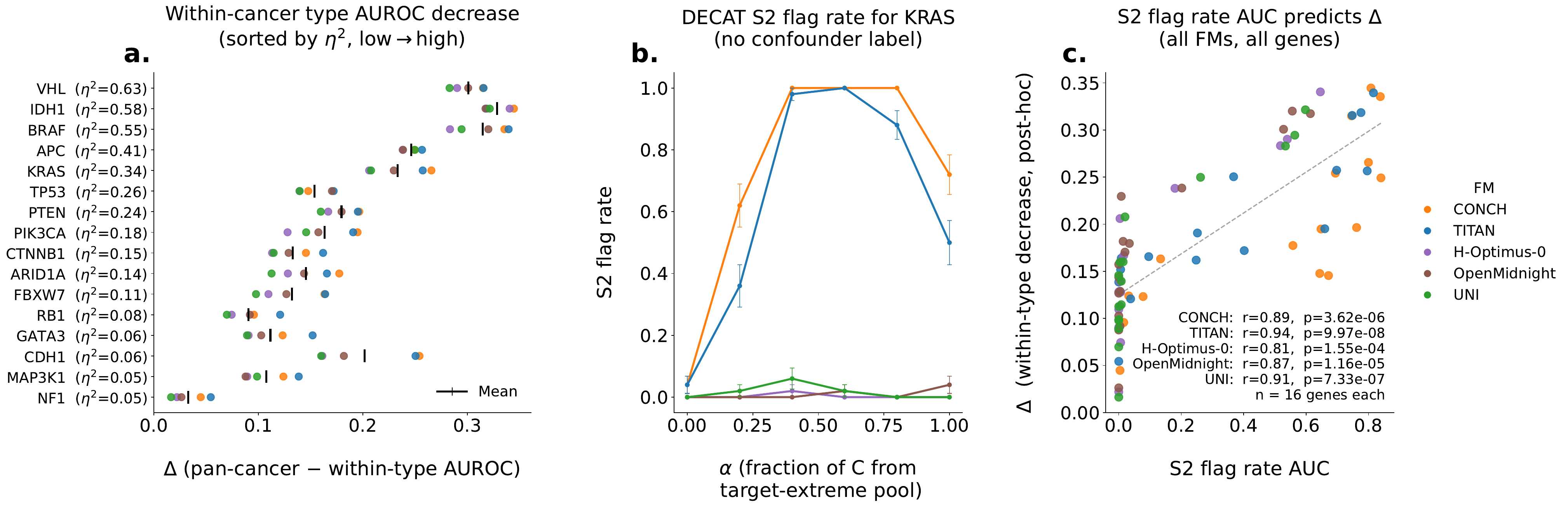}
\caption{\textbf{DECAT's S2 flag rate predicts within-type AUROC collapse across driver genes (unimodal H\&E, Stages~II and IV only).}
(a)~Within-type AUROC collapse $\Delta = \text{AUROC}_{\text{pan}} - \text{AUROC}_{\text{within}}$ (median across cancer types) per FM (colored dots) for 16 driver genes sorted by $\eta^2$ (fraction of mutation prevalence variance explained by cancer type).
(b)~S2 flag rate versus $\alpha$ for KRAS (50 splits). Extreme pool E defined per FM from probe scores on A$'$ (top 10\%), requiring no cancer-type labels.
(c)~S2 AUC (area under the flag rate curve) versus $\Delta$ across all gene$\times$FM combinations. The concordance (Spearman $r$, shown) indicates that the S2 flag rate tracks the same FM--gene combinations that exhibit confounding-driven overestimation under post-hoc stratification.}
\label{fig:tcga_driver_gene}
\end{figure}

\paragraph{Unimodal H\&E foundation model comparison.}
Applied to five H\&E foundation models without paired RNA (TITAN~\cite{dingMultimodalWholeslideFoundation2025}, CONCHv1.5~\cite{luVisuallanguageFoundationModel2024}, UNIv2~\cite{chenGeneralpurposeFoundationModel2024}, H-Optimus-0~\cite{saillardHoptimus02024}, OpenMidnight~\cite{kaplanHowTrainStateoftheArt2025,karasikovTrainingStateoftheartPathology2025}) using Stages~II and IV only, DECAT's S2 flag rate tracks within-type AUROC collapse across 16 driver genes without requiring confounder labels (Figure~\ref{fig:tcga_driver_gene}). The S2 flag rate rises with $\alpha$ for FMs with strong probe directions (CONCH, TITAN), while weaker FMs remain near zero (Figure~\ref{fig:tcga_driver_gene}b). The S2 AUC (area under the flag rate vs $\alpha$ curve) shows strong concordance with the post-hoc within-type AUROC collapse $\Delta$ across all gene$\times$FM combinations (Spearman $\rho = 0.81$--$0.94$, $p < 0.001$ for all FMs) (Figure~\ref{fig:tcga_driver_gene}c), indicating that the S2 flag rate identifies the same FM--gene combinations that exhibit confounding-driven overestimation under post-hoc stratification.

\section{Discussion}
\label{sec:discussion}

\paragraph{Architecture determines diagnostic reliability.}
Entangled models (CCA, CLIP) reliably detect shared signal (S1 at 95--98\% in the synthetic validation) but structurally cannot detect modality-specific signal (S4 at 0\%), and their FSCR (10--22\% synthetic, 83--84\% on TCGA) means shared-signal claims from entangled representations should not be trusted without independent validation. Counterintuitively, FSCR increases with both signal strength and sample size meaning more power to detect signal leads to more false shared claims when the representation lacks factorized components.
Factorized models enable S4 detection (up to 96\%) and substantially reduce false shared claims (FSCR ${\approx}$0\% on synthetic data; strong disentanglement reduced FSCR to 12\% on TCGA, Figure~\ref{fig:tcga_fscr}), but sacrifice S1 sensitivity due to the localization requirement. Within DSSL, the disentanglement strength $\beta_{\text{ssl}}$ behaves differently on synthetic and real data. In the synthetic validation, weaker $\beta_{\text{ssl}}$ improves S4 sensitivity, because stronger disentanglement pressure partially attributes modality-specific signal to the shared component, and all values maintain FSCR${=}$0 (Figure~\ref{fig:fscr}). On TCGA, however, $\beta_{\text{ssl}} \geq 1.0$ gives both the lowest FSCR (0--0.3\% on S4r tasks; Figure~\ref{fig:tcga_fscr}) and the highest S4 detection (92--100\%, Figure~\ref{fig:tcga_detection_curves}), whereas weaker values detect no S4 signal. As a practical heuristic, we recommend $\beta_{\text{ssl}} \geq 1.0$ as the default. This is not a universal optimum, and users should report sensitivity over a small grid of values when possible. The trade-off between localization and S1 sensitivity is also configurable: skipping the localization stage would recover entangled-level S1 sensitivity at the cost of losing S4 detection and the FSCR safety guarantee. Note that the inability of entangled representations to avoid confident misclassification of confounded or modality-specific signal as shared signal is a property of the representations themselves, not a limitation of the evaluation.

\paragraph{Power limits interpretability, but errors are conservative.}
Detection rates for all scenarios improve with signal strength and evaluation sample size. For factorized models, the gap between strict and conservative accuracy confirms that errors are predominantly indeterminate rather than misclassifications. On TCGA, indeterminate rates range from 4--5\% for entangled models to 51--69\% for DSSL, and the reported metric values identify the cause (Appendix~\ref{app:further-tcga-results:indeterminate}).
The TCGA power curve (Figure~\ref{fig:tcga_power_curve}) reveals that the primary bottleneck is probe quality in the reference cohorts (A$'$/B), not the size of the deployment cohort. Under the vary-C design, TMB is reliably flagged with as few as $N_C{\approx}50$ patients when A$'$/B are held at full size, whereas equal-N designs require $N{\approx}500$. Since the extreme pool comprises 25--36\% of evaluation patients (Appendix~\ref{app:further-tcga-results:power}), populating C requires an evaluation pool of approximately $3$--$4 \times N_C$ patients. This implies that DECAT can evaluate smaller deployment populations if a sufficiently large reference cohort is available for stable probe training.

\paragraph{Confounding is detectable only along the cohort shift.}
$D_{\text{quantile}}^{\text{task}}$ detects instability along the outcome-predictive direction, so confounding is detectable only when the cohort shift between B and C has a component along the confounding direction. When the shift is orthogonal to it ($\rho_b = 0$), S2B detection falls to 4--5\% for both model families (Table~\ref{tab:app-rhob}). Entangled models then misclassify confounded signal as shared signal more often (from 22\% under aligned shift to 72\%), whereas factorized models never assign S1 and route 86\% of these cases to indeterminate.
A confounder orthogonal to the shift does not threaten generalization to cohorts shifted in that direction, but a prediction that appears safe for one deployment cohort is not necessarily safe for all deployment contexts. On TCGA, the extreme-C design aligns the shift with the label-predictive direction by construction.

\paragraph{Proxy detection is the hardest problem.}
In the synthetic validation, while direct confounding is detectable at moderate-to-strong signal, proxy-driven spurious signal remains difficult at 5--20\% strict accuracy, with conservative accuracy reaching 60--75\% for factorized models.
This reflects a limitation of the linear probe used by $D_{\text{quantile}}^{\text{task}}$ because proxy signal that enters the representation along directions not aligned with the cohort shift between B and C remains hidden to the stability test (Appendix~\ref{app:further-results:proxy}).
Improving proxy detection likely requires either larger evaluation cohorts, multiple shifted cohorts that probe different directions of variation, or nonlinear probes that can detect instability along directions inaccessible to linear projections.

\section{Limitations}
\label{sec:limitations}

DECAT addresses whether predictive signal is supported by cross-modally observable biology without requiring prior knowledge of which confounder is present. This complements rather than replaces existing methods like preserved-site cross-validation~\cite{howardImpactSitespecificDigital2021} and covariate stratification~\cite{dawoodConfoundingFactorsBiases2026} that detect specific known confounders. 
DECAT requires per-modality embeddings. Most multimodal models in computational pathology use late fusion, reflecting the scarcity of paired multimodal samples relative to unimodal data, and DECAT applies to all such architectures. For early-fusion architectures, where modalities attend to each other during encoding, Stages~II and IV could be applied to the single fused representation to detect confounding, as in our unimodal FM evaluation. However, the S1 versus S4 distinction is unavailable because per-modality signal localization is undefined after joint encoding, and resolving it would require alternative approaches to localization.
The unimodal application (Stages~II and IV only) can detect confounding but cannot assign S1, which by definition requires agreement across at least two independent modalities.
Real-data validation is limited to oncology (TCGA), and confounding structures in other disease areas may differ. The synthetic simulator makes several simplifying assumptions (Appendix~\ref{app:simulator-limitations}).
Finally, DECAT classifies the category of predictive signal but does not identify specific mechanisms or establish causal relationships. An S1 classification indicates cross-modally supported and stable predictive signal consistent with shared biology, not proof of a shared biological mechanism or that the representation has recovered a causal biological process. Similarly, an S4 classification indicates stable signal confined to one modality, not necessarily modality-specific biology.

\section{Conclusion}
\label{sec:conclusion}

We introduced DECAT, a model-agnostic evaluation framework that classifies multimodal representations into four diagnostic scenarios using five null-referenced metrics and a rule-based decision procedure.
No single architecture reliably classifies all scenarios, with entangled models assigning shared signal when it is absent (FSCR 83--84\% on real embeddings) and factorized models enabling safer localization at the cost of reduced sensitivity. On TCGA, DECAT detects cancer-type and within-cancer-type confounding invisible to AUROC on both multimodal embeddings and frozen H\&E foundation models, without requiring confounder labels. These distinctions were previously unavailable from predictive accuracy alone.

\bibliographystyle{unsrtnat}
\bibliography{references}

\newpage
\input{appendix}

\newpage

\end{document}

%% file: appendix.tex
\appendix
\section*{Appendix}

\section{Additional Simulator Limitations}
\label{app:simulator-limitations}

The measurement model  is linear (Appendix~\ref{app:simulator:measurement}), while real biological assays involve highly nonlinear transformations. This likely advantages linear models (CCA, JIVE) because their inductive bias matches the data-generating process, and does not fully challenge nonlinear encoders. The TCGA validation (Section~\ref{sec:tcga}) confirms that key findings transfer despite this mismatch.
Next, latent variables are independent Gaussians, whereas real biology has complex dependencies between shared structure, modality-specific effects, and batch.
The orthogonality of $z_s$ and $b$ by construction means that S2 detection in our simulator represents an optimistic upper bound. In real data, biology and batch may be correlated along the same directions, making separation harder. The TCGA validation partially addresses this concern, demonstrating that $D_{\text{quantile}}^{\text{task}}$ detects confounding on real embeddings where label prevalence varies substantially across cancer types.
$D_{\text{quantile}}^{\text{task}}$ detects spurious signal only when the cohort shift has a component along the outcome-predictive direction. If the shift is orthogonal to the spurious signal, Scenario~2 is undetectable, though in this case the spurious signal does not constitute a generalization risk for the specific evaluation cohorts. In the simulator this alignment is controlled by parameters ($\rho_b$, $\rho_p$); in the TCGA validation, the extreme-C design constructs cohort shifts along the label-predictive direction by enriching C for label-extreme patients, ensuring alignment without requiring knowledge of the confounder direction. A variant that discovers the cohort-shift direction from data and measures its alignment with the outcome-predictive direction could address this limitation for deployment cohorts where the shift structure is unknown.
Finally, the synthetic experiment uses a single latent dimensionality ($k = 50$) and sensitivity to latent complexity is not characterized.

\section{Broader Impacts}
\label{app:broader-impacts}

DECAT is an evaluation framework for multimodal biomedical representations and does not directly make clinical decisions, generate predictions for patients, or process identifiable patient data. Its intended use is as a model diagnostic tool to assess whether learned representations support biological interpretation before downstream use.

\paragraph{Positive impacts.}
By making the distinction between transferable cross-modally supported signal and spurious signal explicit, DECAT may help prevent the deployment of multimodal biomarkers that rely on batch effects or proxy features, reducing the risk of clinical decisions based on models that appear accurate but do not generalize. The framework returns an indeterminate output to provide an explicit signal that the available evidence does not support reliable scenario classification, rather than forcing a potentially incorrect label.

\paragraph{Potential risks.}
A false sense of security could arise if practitioners treat DECAT's Scenario~1 classification as a guarantee of biological validity. Scenario~1 indicates that predictive signal is cross-modally supported and stable across the evaluation cohorts, but it does not establish a causal relationship between the representation and the underlying biology. DECAT's detection power is limited by evaluation sample size and proxy geometry, and the framework may fail to detect spurious signal under conditions not covered by the evaluation cohorts. Practitioners should interpret DECAT's output as one component of a broader model validation strategy, not as a standalone certification of biological validity.

\paragraph{Disclosure.}
All authors are employees of AstraZeneca and have stock ownership and/or stock options or interests in the company.

\section{Related Work}
\label{app:related}

\paragraph{Multimodal representation learning in pathology.}
Recent multimodal foundation models integrate histopathology with molecular data using contrastive objectives, demonstrating that molecular supervision provides a powerful inductive bias for learning clinically relevant histology representations~\cite{jaumeTranscriptomicsguidedSlideRepresentation2024a,chenPancancerIntegrativeHistologygenomic2022,vaidyaMoleculardrivenFoundationModel2025,yunEXAONEPath252025,xuMultimodalKnowledgeenhancedWholeslide2025,liuLeveragingMultimodalFoundation2026,chenVisualOmicsFoundation2025}.
These models optimize cross-modal alignment rather than factorization and their embedding spaces may entangle shared biological signal with modality-specific effects, cohort structure, and correlated proxy features.
As a result, proximity in an entangled embedding space is difficult to attribute to a specific biological process, and it remains unclear when predictions derived from a single modality, such as H\&E alone, are biologically justified.
DECAT does not compete with these models but instead evaluates entangled representations post hoc to determine when predictions are biologically supported.

\paragraph{Disentangled multimodal learning.}
Contrastive methods such as CLIP implicitly assume multi-view redundancy, i.e., that each modality contains overlapping information about the underlying signal.
Factorized contrastive learning~\cite{liangFactorizedContrastiveLearning2023} demonstrates that private, modality-specific information can be predictive, but does not provide a mechanism to determine whether predictive signal is supported by cross-modally shared biology.
Recent approaches explicitly separate shared and modality-specific representations using mutual information objectives~\cite{wangInformationCriterionControlled2025,fischerConditionalEntropyBottleneck2020,zhangDisentanglementVariationsMultimodal2025,guiIndiSeekLearnsInformationguided2025,wuLearningOptimalMultimodal2025} and architectural decompositions with partially shared latent spaces~\cite{zhangPartiallySharedMultimodal2026}.
Prior work primarily evaluates disentanglement in terms of representation quality, reconstruction, or downstream predictive performance.
A complementary information-theoretic perspective uses partial information decomposition to quantify synergistic versus redundant contributions of each modality, providing a criterion for when multimodal integration adds information beyond the strongest unimodal representation~\cite{richterAlignmentSynergisticIntegration2026}. However, synergistic information may still reflect spurious correlations rather than biology.
In contrast, DECAT treats disentangled representations as one input to a diagnostic framework that determines when predictions from partial data are biologically supported, a question that has not been studied in prior disentangled multimodal learning work.

\paragraph{Batch effects and confounding in computational pathology.}
Site-specific digital signatures in histology~\cite{howardImpactSitespecificDigital2021}, persistent medical center signatures in foundation model embeddings~\cite{komenHistopathologicalFoundationModels2024,jongCurrentPathologyFoundation2025}, and scanner-induced domain shifts that are masked by stable prediction accuracy~\cite{thiringerScannerInducedDomainShifts2026} are well-documented sources of spurious predictive signal. PathoROB~\cite{komenRobustFoundationModels2025} provides a systematic robustness benchmark measuring the degree to which FM embeddings encode medical center signatures rather than biology, finding substantial robustness deficits across 20 evaluated FMs. Strong dependencies between biomarkers and clinicopathological features further cause models to rely on correlated features rather than the biomarker itself~\cite{dawoodConfoundingFactorsBiases2026}.
These findings extend to the multimodal setting, where approximately 30\% of multimodal model performance in one study was attributed to site or institution batch effects rather than biology~\cite{howardMultimodalDeepLearning2023}.
Existing mitigations include preserved-site cross-validation~\cite{howardImpactSitespecificDigital2021}, color normalization, and domain adaptation, but these are detection or correction methods that identify or reduce specific confounders without providing a general diagnostic for whether learned representations support biological interpretation.
DECAT addresses a complementary problem: given a trained representation, determine whether its predictive signal is consistent with shared biology or reflects spurious signal (which may transfer across cohorts but exhibits composition-dependent instability), modality-specific effects, or noise, without requiring knowledge of which specific confounder is present.

\paragraph{Identifiability and causal representation learning.}
Unsupervised disentanglement without inductive biases is theoretically impossible~\cite{locatelloChallengingCommonAssumptions2019}, motivating our approach of evaluating representations post hoc rather than attempting to recover ground-truth latent structure.
Recent work connects representation learning to causal structure discovery, arguing that representation learning fails at causal tasks without proper structure~\cite{uhlerCausalStructureRepresentation2025}.
DECAT classifies the category of predictive signal (shared, spurious, noise, modality-specific) without identifying the underlying mechanism or establishing causal relationships. The framework returns \emph{indeterminate} when statistical evidence is insufficient for confident classification.

\section{Simulator Details}
\label{app:simulator}

This appendix specifies the synthetic data generator in full. The generator is intentionally minimal. Its goal is not to mimic any specific biological assay, but to construct the smallest latent structure sufficient to instantiate each of the four diagnostic scenarios (S1--S4) with controllable ground truth. All downstream metrics operate only on learned representations. Ground-truth latents, weight vectors, and scenario labels are never observed by representation learning or by the decision procedure.

\subsection{Latent Variables}
\label{app:simulator:latents}

For each patient $i$, we sample low-dimensional latents $z_s, z_h, z_r, b \in \mathbb{R}^k$ representing shared biology, H\&E-specific biology, RNA-specific biology, and a shared nuisance factor (interpreted operationally as ``batch'' but covering any shared non-biological source such as institution, cohort, or correlated biological confounders such as tumor purity):
\begin{equation}
z_s,\ z_h,\ z_r,\ b \;\overset{\text{i.i.d.}}{\sim}\; \mathcal{N}(\mathbf{0},\,\mathbf{I}).
\label{eq:app-latents}
\end{equation}
Independence is imposed by construction so that cross-latent dependence in learned representations reflects the measurement process or model, not the generator. We evaluated $k \in \{5,10,50,100\}$ during pre-experiment calibration (Appendix~\ref{app:experiments:calibration}) and use $k=50$ throughout the main experiment.

\subsection{Measurement Model}
\label{app:simulator:measurement}

Each modality is a linear mixture of latents plus Gaussian noise:
\begin{align}
    x_h &= \beta_s A_s z_s + \beta_h A_h z_h + \beta_b A_b b + \epsilon_h, \qquad x_h \in \mathbb{R}^{d_h} \label{eq:app-xh} \\
    x_r &= \beta_s B_s z_s + \beta_r B_r z_r + \beta_b B_b b + \epsilon_r, \qquad x_r \in \mathbb{R}^{d_r} \label{eq:app-xr}
\end{align}
with default $d_h = d_r = 1024$. The matrices $A_\ast, B_\ast$ are sampled once per synthetic run and held fixed across all patients in that run, representing the unknown measurement process.

\paragraph{Block normalization.}
To prevent accidental scaling artifacts from random matrices, each block matrix $M \in \mathbb{R}^{d \times k}$ is rescaled so that every latent block contributes unit expected variance per observed dimension before the $\beta$ scalars are applied:
\begin{equation}
    \tfrac{1}{d}\,\mathrm{tr}(M M^\top) = 1.
\end{equation}
We draw $M$ from a standard Gaussian and multiply by $\sqrt{d / \mathrm{tr}(M M^\top)}$. This decouples observed variance from the random matrix magnitude, so the $\beta$ coefficients fully control the relative contribution of each latent block.

\paragraph{Fixed-SNR noise calibration.}
Under block normalization, the per-dimension signal variance for each modality is
\begin{equation}
    \mathrm{Var}_{\mathrm{signal}}(x_h) \approx \beta_s^2 + \beta_h^2 + \beta_b^2,
    \qquad
    \mathrm{Var}_{\mathrm{signal}}(x_r) \approx \beta_s^2 + \beta_r^2 + \beta_b^2.
\end{equation}
We fix the signal-to-noise ratio at $\mathrm{SNR} = 3$ across all measurement regimes by setting
\begin{equation}
    \sigma_h^2 = \tfrac{1}{3}(\beta_s^2 + \beta_h^2 + \beta_b^2),
    \qquad
    \sigma_r^2 = \tfrac{1}{3}(\beta_s^2 + \beta_r^2 + \beta_b^2),
\end{equation}
so that differences across regimes reflect signal composition rather than overall task difficulty. Table~\ref{tab:app-regimes} reports the four measurement regimes used in the main experiment.

\begin{table}[h]
\centering
\small
\caption{\textbf{Measurement regimes.} Each non-baseline regime doubles exactly one coefficient from its baseline value; noise is re-calibrated to preserve $\mathrm{SNR}=3$. Preliminary calibration identified $\beta_b = 0.75$ as the baseline operating point because at $\beta_b = 1.0$, S2B detection becomes trivial.}
\label{tab:app-regimes}
\begin{tabular}{lcccccc}
\toprule
\textbf{Regime} & $\beta_s$ & $\beta_h$ & $\beta_r$ & $\beta_b$ & $\sigma_h^2$ & $\sigma_r^2$ \\
\midrule
Baseline            & 1.0 & 1.0 & 1.0 & 0.75 & 0.854 & 0.854 \\
Shared-dominant     & 2.0 & 1.0 & 1.0 & 0.75 & 1.92  & 1.92  \\
Batch-dominant      & 1.0 & 1.0 & 1.0 & 1.50 & 1.42  & 1.42  \\
Modality-dominant   & 1.0 & 2.0 & 2.0 & 0.75 & 1.92  & 1.92  \\
\bottomrule
\end{tabular}
\end{table}

\subsection{Outcome Generation}
\label{app:simulator:outcome}

Binary outcomes $y_i \in \{0,1\}$ are generated from a latent score:
\begin{equation}
r_i = \alpha_s \langle \bar{w}_s,\, z_{s,i}\rangle
    + \alpha_h \langle \bar{w}_h,\, z'_{h,i}\rangle
    + \alpha_r \langle \bar{w}_r,\, z'_{r,i}\rangle
    + \alpha_b \langle \bar{w}_b,\, b_i\rangle
    + \epsilon_{y,i},
\qquad y_i = \mathbf{1}(r_i > 0),
\label{eq:app-outcome}
\end{equation}
where $\bar{w}_\ast = w_\ast / \|w_\ast\|_2$ are unit-norm weight vectors sampled once per run and $\epsilon_{y,i} \sim \mathcal{N}(0, \sigma_y^2)$ tunes task difficulty ($\sigma_y = 0.5$ in all main experiments). Because $\bar{w}$ is unit-norm and $z \sim \mathcal{N}(\mathbf{0},\mathbf{I})$, each term $\langle \bar{w}, z\rangle$ has unit variance, so the coefficients $\alpha_\ast$ directly control the fractional variance contribution of each latent source to the outcome.

\paragraph{Scenario-to-$\alpha$ mapping.}
Setting different coefficients to zero instantiates each scenario (Table~\ref{tab:app-scenarios}). Scenarios are defined \emph{per task and per modality}; for example, a task with $\alpha_s > 0$ alone yields Scenario~1 for both modalities, whereas $\alpha_h > 0$ alone yields Scenario~4 for H\&E and Scenario~3 for RNA.

\begin{table}[h]
\centering
\small
\caption{\textbf{Canonical scenario definitions.} Scenario~2 encompasses two distinct mechanisms. Outcome-level confounding ($\alpha_b > 0$, Scenario~2B in the main text) is controlled via the scenario-to-$\alpha$ mapping described above. Representation-level proxy ($\gamma > 0$) operates at the feature level and is described in Appendix~\ref{app:simulator:proxy}. These two mechanisms are observationally similar but arise from distinct generative sources.}
\label{tab:app-scenarios}
\begin{tabular}{lll}
\toprule
\textbf{Scenario} & \textbf{Outcome signal location} & \textbf{Coefficients} \\
\midrule
S1 (shared biology)     & Shared latent only            & $\alpha_s > 0,\ \alpha_h = \alpha_r = \alpha_b = 0$ \\
S2 (spurious signal)  & Confounding or proxy          & $\alpha_b > 0$ \emph{or} ($\gamma > 0$ and $\alpha_m > 0$), $\alpha_s = 0$ \\
S3 (no signal)          & None                          & $\alpha_s = \alpha_h = \alpha_r = \alpha_b = 0$ \\
S4 (modality-specific)  & Modality $m$ only             & $\alpha_m > 0,\ \alpha_s = \alpha_{m'} = \alpha_b = 0,\ \gamma = 0$ \\
\bottomrule
\end{tabular}
\end{table}

\subsection{Proxy Entanglement}
\label{app:simulator:proxy}

To model representation-level proxy signal (e.g., site-specific staining artifacts correlated with outcome), we optionally entangle modality-specific biology with batch structure using a modality-specific contamination coefficient $\gamma_m$ :
\begin{equation}
z'_{h,i} = z_{h,i} + \gamma_h\,\text{proxy}_i,
\qquad
z'_{r,i} = z_{r,i} + \gamma_r\,\text{proxy}_i,
\label{eq:app-proxy}
\end{equation}
where the proxy direction is
\begin{equation}
\text{proxy}_i = \frac{(1-\eta)\,b_i + \eta\,\tilde{b}_i}{\sqrt{(1-\eta)^2 + \eta^2}},
\qquad \tilde{b}_i \sim \mathcal{N}(\mathbf{0},\mathbf{I}).
\end{equation}
The coefficient $\gamma_m$ controls per-modality proxy strength and the misalignment parameter $\eta \in [0,1]$ controls alignment of proxy signal with the cross-modal batch axis. These two parameters are orthogonal: $\gamma$ governs magnitude, $\eta$ governs geometric alignment across cohorts.

\paragraph{Proxy regimes.}
\emph{Single-modality proxy} ($\gamma_m > 0$ for one modality, zero for the other) models artifacts confined to a single assay (e.g., scanner or staining contamination of H\&E with $\gamma_h > 0, \gamma_r = 0$); \emph{dual-modality proxy} ($\gamma_h = \gamma_r > 0$) models site-level batch effects that affect both assays. Separately, \emph{aligned proxy} ($\eta = 0$) concentrates along the batch axis, which shifts between cohorts. \emph{Misaligned proxy} ($\eta > 0$) spreads off the batch axis. Because the off-axis component does not shift between cohorts it is therefore harder to detect.

\paragraph{Terminology: confounding vs.\ proxy.}
We distinguish two conceptually distinct sources of spurious predictive signal. \emph{Outcome-level confounding} ($\alpha_b$) occurs when a latent directly influences both the observed features and the outcome (e.g., smoking affecting both tumor morphology and survival). \emph{Representation-level proxy signal} ($\gamma_h, \gamma_r$) arises when modality-specific latents mix with batch-aligned variation at the feature level, creating label-correlated directions in the learned representation that reflect batch rather than the biology of interest (e.g., hospital-of-care effects that correlate with patient response via treatment differences). Both mechanisms produce high in-distribution AUROC because the spurious signal is genuinely correlated with the outcome in the training population, making them indistinguishable from real biology by standard evaluation metrics.

\subsection{Cohort Structure and Shift Alignment}
\label{app:simulator:cohorts}

Each run produces four disjoint cohorts that differ only in the distribution of the batch latent $b$:
\begin{align}
\text{Cohort A}   &:\;\; b \sim \mathcal{N}(\mathbf{0},\mathbf{I}) && \text{(representation training)} \\
\text{Cohort A}'  &:\;\; b \sim \mathcal{N}(\mathbf{0},\mathbf{I}) && \text{(within-distribution evaluation)} \\
\text{Cohort B}   &:\;\; b \sim \mathcal{N}(\mu_b,\,\mathbf{I}) && \text{(external, shifted)} \\
\text{Cohort C}   &:\;\; b \sim \mathcal{N}((1+\varepsilon)\mu_b,\,\mathbf{I}) && \text{(external, scaled shift)}
\end{align}
Here $\mu_b \in \mathbb{R}^k$ is the cohort-level batch mean shift. For Cohort C we sample using the \emph{same mean direction} $\mu_b$ as Cohort B with a relative scaling factor ($\varepsilon = 0.2$ by default). This ensures that both external cohorts differ from A along the same batch direction, so any patient ordering instability (Appendix~\ref{app:metrics:dq}) observed between B and C is only due to sensitivity to batch strength rather than to a mean shift in a different direction. Two external cohorts are necessary because a single shifted cohort cannot distinguish changes in patient ordering due to batch strength from those due to sampling variability.

\paragraph{Batch shift alignment ($\rho_b$).}
By default, $\mu_b$ is drawn independently of the outcome-predictive batch direction $\bar{w}_b$ (Eq.~\ref{eq:app-outcome}), so the direction that shifts between cohorts is unrelated to the direction of $b$ that predicts the outcome. This does not model outcome-relevant confounding, where batch variation aligns with predictive signal. Therefore, we add a batch alignment parameter $\rho_b \in [0,1]$ that controls the degree of alignment between the cohort shift direction $\mu_b$ and the outcome-predictive batch direction $\bar{w}_b$:
\begin{equation}
\mu_b = \frac{(1-\rho_b)\,\hat{n} + \rho_b\,\bar{w}_b}{\|(1-\rho_b)\,\hat{n} + \rho_b\,\bar{w}_b\|} \cdot M,
\label{eq:app-batchalign}
\end{equation}
where $\hat{n}$ is a random direction (drawn from a standard Gaussian and normalized to unit length) and $M$ is the batch shift magnitude, held fixed (we use $M=2.0$, corresponding to Cohen's $d \approx 2$). The magnitude $\|\mu_b\| = M$ is preserved regardless of $\rho_b$, so batch alignment controls geometry only.

\paragraph{Proxy shift alignment ($\rho_p$).}
In realistic proxy settings, the site-level artifact that contaminates a modality is typically the same axis that differs between institutions (e.g., the staining protocol that varies across hospitals is the one that creates outcome-correlated morphological features). By default, the cohort shift $\mu_b$ is unrelated to the proxy-contaminated direction $\bar{w}_m$, which is unrealistic for this mechanism. The proxy shift alignment parameter $\rho_p \in [0,1]$ controls the degree of alignment between the cohort shift and the proxy direction, extending Eq.~\ref{eq:app-batchalign} with a third term:
\begin{equation}
\mu_b = \frac{(1-\rho_b-\rho_p)\,\hat{n} + \rho_b\,\bar{w}_b + \rho_p\,\bar{w}_m}{\|(1-\rho_b-\rho_p)\,\hat{n} + \rho_b\,\bar{w}_b + \rho_p\,\bar{w}_m\|} \cdot M,
\end{equation}
where the target modality $m \in \{h, r\}$ is the proxy-contaminated modality, with the constraint $\rho_b + \rho_p \leq 1$. When $\rho_p = 0$ this reduces to Eq.~\ref{eq:app-batchalign}; $\rho_p = 1$ with $\rho_b = 0$ corresponds to a site signature in which the staining difference between institutions is the one that mimics outcome-predictive morphology.

\paragraph{Shift configuration used in experiments.}
Table~\ref{tab:app-shift} summarizes the shift alignment used per scenario class. All scenarios use $M = 2.0$ and alignment is scenario-dependent so that the cohort shift either probes the confounding direction (S2B), the proxy direction (S2A/C/D), or is independent of both (S1, S3, S4 baselines).

\begin{table}[h]
\centering
\small
\caption{\textbf{Cohort shift alignment per scenario class.} The magnitude $M = 2.0$ is fixed; $\rho_b$ and $\rho_p$ control geometry only.}
\label{tab:app-shift}
\begin{tabular}{llll}
\toprule
\textbf{Scenario class} & $\rho_b$ & $\rho_p$ & \textbf{Proxy target} \\
\midrule
Baseline (S1, S3, S4)               & 0.0 & 0.0 & --- \\
S2B (outcome confounding)           & 1.0 & 0.0 & --- \\
S2A/C/D ($\gamma_h > 0$)            & 0.0 & 1.0 & \texttt{'h'} \\
S2A/C/D ($\gamma_r > 0$)            & 0.0 & 1.0 & \texttt{'r'} \\
Transition B (confound + proxy)     & 0.5 & 0.5 & dominant proxy modality \\
\bottomrule
\end{tabular}
\end{table}

\subsection{Ambiguous Configurations}
\label{app:simulator:identifiability}

We do not assume that a single scenario label is always appropriate. Three configurations are explicitly designed to be ambiguous for scenario classification. \emph{Transition~A} ($\alpha_s \approx \alpha_h \approx \alpha_r > 0$) introduces competing biological signal sources with no dominant mechanism. \emph{Transition~B} adds confounding and proxy contamination on top of Transition~A ($\alpha_b > 0$, $\gamma > 0$), creating a configuration where multiple signal sources and spurious structure co-exist. \emph{Aligned proxy} ($\eta = 0$ with $\gamma > 0$) places proxy perfectly along the shared batch axis, making it indistinguishable from true shared biology via structural geometry alone so that only $D_{\text{quantile}}^{\text{task}}$ with $\rho_p > 0$ can detect it, by targeting the outcome-probe direction rather than an arbitrary geometric axis. The correct framework behavior in these configurations is to return \emph{indeterminate}, not a confident scenario label. These configurations provide ground truth for evaluating whether the decision procedure returns indeterminate when the underlying signal structure is genuinely ambiguous.

\subsection{Structural Assumptions}
\label{app:simulator:assumptions}

The generator makes several simplifying assumptions whose implications are discussed in the main Limitations section:
\begin{enumerate}
    \item Observations are linear mixtures of latents plus Gaussian noise. Real assays involve nonlinear mixing, which the linear simulator does not capture.
    \item Latents are independent Gaussians. Real biology has complex dependencies and imposed independence is an optimistic assumption for $z_s$/$b$ separation.
    \item Measurement matrices $A_\ast, B_\ast$ are fixed across cohorts. Real cohorts differ in scanner, protocol, and pre-processing and we do not model measurement-level shifts.
    \item Confounding is shared across modalities via a single latent $b$. Modality-specific artifacts uncorrelated with the shared batch axis are not modeled as a separate latent, though single-modality proxy ($\gamma_h > 0, \gamma_r = 0$) captures a related mechanism. Such artifacts do not generate shared-looking signal and are therefore not the hardest failure mode for S1 misclassification.
    \item Outcome weight vectors $\bar{w}_\ast$ are fixed across patients and dense (every latent dimension contributes). Real biomarkers may be sparse, with phenotype depending on a low-dimensional mechanism embedded in a higher-dimensional latent space.
    \item Modalities are treated as symmetric (similar SNR and contamination). In practice, one modality may capture shared biology more reliably than the other.
    \item All outcome-relevant structure is encoded in $(z_s, z_h, z_r, b)$ with no unmodeled confounders. In real data, additional unmeasured factors may influence both the outcome and the observed features.
    \item All modalities observe the same patient's latent state with perfect spatial and temporal correspondence. In real data, different assays may sample different tissue regions or time points, introducing intratumor heterogeneity and sampling variation that the simulator does not model.
\end{enumerate}

\section{Model Architectures and Training}
\label{app:models}

We evaluate four model classes spanning a two-by-two grid of linear vs.\ nonlinear and entangled vs.\ factorized. All models are trained exclusively on paired observations from Cohort~A without access to outcome labels and encoders are frozen after training. This appendix section specifies each model's representation structure, training procedure, and hyperparameters.

\subsection{Shared Architecture for Nonlinear Models}
\label{app:models:shared}

To isolate the effect of factorization among nonlinear models, CLIP and DisentangledSSL share identical encoder and projection architectures. Each modality is mapped to a latent via a two-layer MLP encoder (hidden width $4 \times d_m$ ($d_m = 1024$ for both modalities), ReLU activation, L2-normalized output). Architecture, normalization, temperature, initialization, optimizer, and learning rate schedule are held fixed across models. This design ensures that differences in learned representations reflect training objectives (contrastive alignment vs.\ conditional entropy bottleneck) rather than architectural capacity. Linear baselines (CCA, JIVE) do not share this architecture because they are closed-form.

\paragraph{Training details.}
Both nonlinear models are optimized with Adam ($\text{lr} = 10^{-3}$, constant schedule, no weight decay). Contrastive losses use a fixed temperature $\tau = 0.07$. For the synthetic experiment, batch size is 1024 and CLIP trains for 200 epochs. DisentangledSSL trains in two phases: a shared-latent alignment phase (75--125 epochs, depending on regularization strength) followed by a modality-specific independence phase (50 epochs), each with a separate Adam optimizer at the same learning rate. For the TCGA validation experiment, batch size is 256 and epoch counts are unchanged. No learning rate warmup, weight decay, or gradient clipping is applied.

\subsection{CCA (Linear Entangled)}
\label{app:models:cca}

Canonical correlation analysis learns paired linear projections $W_h \in \mathbb{R}^{d_h \times K}, W_r \in \mathbb{R}^{d_r \times K}$ that maximize correlation between modalities:
\begin{equation}
z_{\text{HE}} = W_h^\top x_{\text{HE}}, \qquad z_{\text{RNA}} = W_r^\top x_{\text{RNA}}.
\end{equation}
The resulting representations are entangled because CCA does not separate shared from modality-specific structure. We retain the top $K = 50$ canonical components. CCA has no trainable hyperparameters.

\subsection{CLIP (Nonlinear Entangled)}
\label{app:models:clip}

The CLIP baseline is a standard symmetric contrastive model. Positive pairs are matched patient samples across modalities; negative pairs are mismatched patients within the batch. Training minimizes the symmetric InfoNCE loss without any factorization-inducing regularizer. CLIP produces a single joint embedding $z_m \in \mathbb{R}^K$ per modality, $K = 50$. This embedding conflates shared biology, modality-specific biology, batch effects, and proxy features, and these components are not separable by construction.

\subsection{JIVE (Linear Factorized)}
\label{app:models:jive}

Our linear factorized model is a simplified variant inspired by JIVE (Joint and Individual Variation Explained)~\cite{lockJointIndividualVariation2013}. Rather than the full iterative JIVE procedure with permutation-based rank selection, we use CCA for joint subspace estimation followed by residual PCA for individual subspaces. This provides a conservative estimate of factorized model performance, since a more sophisticated factorization algorithm would likely improve separation quality. Training is two-step:
\begin{enumerate}
    \item \textbf{Joint subspace:} compute the SVD of the cross-covariance $C = X_h^\top X_r / (n-1)$ with the top $K$ left and right singular vectors $W_h, W_r$ defining the joint projection directions. This step is identical to CCA, so JIVE's joint component matches CCA's shared representation exactly.
    \item \textbf{Individual subspaces:} project out the joint subspace from each modality ($X_h^{\text{res}} = X_h - X_h W_h W_h^\top$, analogously for $r$), then apply PCA to each residual to obtain $K$ individual components per modality.
\end{enumerate}
For evaluation cohorts, the encoder returns the concatenation $[Z^{(h)}_{\text{joint}} \mid Z^{(h)}_{\text{indiv}}] \in \mathbb{R}^{n \times 2K}$. The first $K$ dimensions are joint (shared) and the remaining $K$ are individual (modality-specific), which $\Delta_{\text{shared}}$ uses to separate the two components for localization.

The original JIVE formulation decomposes each data block as $X_i = J_i + A_i + E_i$ (joint + individual + noise) and uses a permutation test on the noise residual to select ranks. Our implementation fixes $K{=}50$ for both joint and individual components based on pre-experiment calibration (Appendix~\ref{app:experiments:calibration}) rather than data-driven rank selection. The noise residual $E_i$ is implicitly discarded after projection and does not enter the DECAT evaluation, which operates only on the projected scores.

\subsection{DisentangledSSL (Nonlinear Factorized)}
\label{app:models:dssl}

DisentangledSSL~\cite{wangInformationCriterionControlled2025} factorizes each modality's representation into a shared latent $z_c^{(m)} \in \mathbb{R}^{d_c}$ and a modality-specific latent $z_m \in \mathbb{R}^{d_m}$ via a two-stage training procedure with information-theoretic regularization.

\paragraph{Stage I: cross-modal alignment with redundancy penalty.}
A contrastive alignment objective encourages agreement of shared latents across modalities, and a redundancy penalty (weight $\lambda$) discourages overlap between the shared latent $z_c$ and modality-specific latents $z_m$.

\paragraph{Stage II: independence enforcement.}
The shared latent $z_c$ is frozen. Modality-specific latents are trained to be independent of $z_c$ via a conditional entropy bottleneck penalty of strength $\beta_{\text{ssl}}$.

\paragraph{Hyperparameter grid.}
We sweep a 4-point diagonal $\beta_{\text{ssl}} = \lambda \in \{0.01, 0.1, 1.0, 10.0\}$, spanning near-zero disentanglement (where DSSL behavior approaches CLIP) to strong disentanglement (approaching representational collapse). The diagonal $\beta_{\text{ssl}} = \lambda$ is the natural symmetric choice because both modalities operate on the same information scale (Table~\ref{tab:app-dssl}).

\begin{table}[h]
\centering
\small
\caption{\textbf{DisentangledSSL hyperparameter grid.} There is no single optimal $(\lambda, \beta_{\text{ssl}})$ across biological settings. Reporting the full sweep reveals how diagnostic conclusions depend on factorization strength.}
\label{tab:app-dssl}
\begin{tabular}{llll}
\toprule
\textbf{Config} & $\beta_{\text{ssl}}$ & $\lambda$ & \textbf{Interpretation} \\
\midrule
DSSL(0.01)  & 0.01  & 0.01  & Near-zero disentanglement (behavior approaches CLIP) \\
DSSL(0.1)   & 0.10  & 0.10  & Weak disentanglement \\
DSSL(1.0)   & 1.00  & 1.00  & Moderate disentanglement \\
DSSL(10.0)  & 10.00 & 10.00 & Strong disentanglement \\
\bottomrule
\end{tabular}
\end{table}

\subsection{Training Protocol}
\label{app:models:training}

All representation models are trained on Cohort~A only ($N_{\text{train}} = 1500$ patients, $k = 50$ latent dimensions, $K_{\text{model}} = 50$ representation dimensions). Training sample size was calibrated via an alignment saturation study (Pre Step~A, Appendix~\ref{app:experiments:calibration}) to ensure representation learning is not sample-limited. Nonlinear models (CLIP, DisentangledSSL) were validated for epoch-level convergence via Pre Step~A.2, which tracks $A_{\text{norm}}^{*}$ on held-out Cohort~A$'$ at intermediate checkpoints. Closed-form models (CCA, JIVE) require no convergence validation.

\section{Metric Definitions}
\label{app:metrics}

All metrics operate only on learned representations and are calibrated against permutation null distributions. We use $i, j$ to index patients, $m \in \{\text{HE}, \text{RNA}\}$ for modality, $z_c^{(i,m)}$ for the shared representation inferred from modality $m$, and $z_m^{(i)}$ for the modality-specific latent when the model exposes one.

\subsection{Standardization Convention}
\label{app:metrics:standardization}

For structural cosine-based metrics ($A_{\text{norm}}, B_{\text{norm}}$), representations are z-scored per latent dimension within Cohort~A$'$, and the same A$'$ statistics are applied to both modalities. This prevents arbitrary scale differences (e.g., in CCA canonical variates) from biasing cosine computations.

For cross-cohort stability metrics ($P_{\text{transfer}}, D_{\text{quantile}}^{\text{task}}$), representations are projected without per-cohort z-scoring. Quantile-based comparisons are anchored to a fixed reference cohort (Cohort~A$'$ for $D_{\text{quantile}}^{\text{task}}$). Per-cohort z-scoring is not used because it removes the cohort mean-shift signal, collapsing detection power for $D_{\text{quantile}}^{\text{task}}$ to near zero. This distinction is central to $D_{\text{quantile}}^{\text{task}}$'s ability to discriminate S1 from S2.

\subsection{$A_{\text{norm}}$: Cross-Modal Shared Latent Agreement}
\label{app:metrics:anorm}

\paragraph{Diagnostic question.}
Does the model recover patient-specific shared structure consistently across modalities?

\paragraph{Definition.}
For patients with at least two modalities and modality pair $(m_1, m_2)$,
\begin{equation}
\text{Agreement}_{\text{obs}} = \mathbb{E}_i\!\left[\cos\!\left(z_c^{(i,m_1)}, z_c^{(i,m_2)}\right)\right],
\qquad
\text{Agreement}_{\text{null}} = \mathbb{E}_{i \ne j}\!\left[\cos\!\left(z_c^{(i,m_1)}, z_c^{(j,m_2)}\right)\right],
\end{equation}
and $A_{\text{norm}} = \text{Agreement}_{\text{obs}} - \text{Agreement}_{\text{null}}$.

\paragraph{Null calibration.}
A 95\% two-sided permutation null interval $[\text{null}_{\text{lower}}, \text{null}_{\text{upper}}]$ is estimated by randomly permuting patient correspondence across modalities.

\paragraph{Decision rule.}
Patient-level shared structure is supported when $A_{\text{norm}} > \text{null}_{\text{upper}}$ for all modality pairs.

\paragraph{Applicability.}
All models. The shared representation $z_c^{(i,m)}$ is defined per architecture: CLIP's contrastive embeddings, CCA's canonical variates, JIVE's joint subspace projections, and DisentangledSSL's shared latents. Note that for modality-anchored architectures where a data-rich modality (e.g., H\&E) servers as the reference space, high $A_{\text{norm}}$ may reflect the training objective rather than independently observable shared biology.

\subsection{$B_{\text{norm}}$: Prototype Consistency Across Modalities}
\label{app:metrics:bnorm}

\paragraph{Diagnostic question.}
Is biological interpretation invariant to which modality was used to infer the shared representation?

\paragraph{Definition.}
Let $p$ be a prototype in shared space (constructed via $k$-means with $k=2$ on Cohort~A shared latents, anchored to a fixed modality). For each patient,
\begin{equation}
\Delta_p(i) = \left| \cos(z_c^{(i,m_1)}, p) - \cos(z_c^{(i,m_2)}, p) \right|,
\end{equation}
with a null computed by breaking patient correspondence:
\begin{equation}
\Delta_{p,\text{null}} = \mathbb{E}_{i \ne j}\!\left[\left| \cos(z_c^{(i,m_1)}, p) - \cos(z_c^{(j,m_2)}, p) \right|\right],
\end{equation}
and $B_{\text{norm}} = \mathbb{E}_i[\Delta_p(i)] - \Delta_{p,\text{null}}$.

\paragraph{Null calibration.}
A 95\% two-sided permutation null interval $[\text{null}_{\text{lower}}, \text{null}_{\text{upper}}]$ is estimated by randomly permuting patient correspondence across modalities, matching the procedure for $A_{\text{norm}}$.

\paragraph{Decision rule.}
$B_{\text{norm}} < \text{null}_{\text{lower}}$ indicates modality-invariant prompting and $B_{\text{norm}} > \text{null}_{\text{upper}}$ indicates modality-dependent prompting. We use a two-sided null interval with two one-sided tests.

\paragraph{Applicability.}
All models. $B_{\text{norm}}$ is used as a diagnostic of interpretability stability but is not a gate in the decision tree. A representation may encode transferable biological signal while still yielding modality-dependent prompts.

\subsection{$\Delta_{\text{shared}}$: Predictive Signal Localization}
\label{app:metrics:delta}

\paragraph{Diagnostic question.}
Does outcome-relevant signal reside in the shared or modality-specific component?

\paragraph{Definition.}
For each modality $s$, train three linear probes on Cohort~A$'$ using fixed stratified 5-fold CV: $f_c^{(s)}$ on the shared component, $f_m^{(s)}$ on the modality-specific component, and $f_{\text{both}}^{(s)}$ on their concatenation. Define $\text{AUROC}_{\max}^{(s)} = \max\{\text{AUROC}(f_c), \text{AUROC}(f_m), \text{AUROC}(f_{\text{both}})\}$ and
\begin{equation}
\Delta_{\text{shared}}^{(s)} = \text{AUROC}(f_c^{(s)}) - \text{AUROC}(f_m^{(s)}).
\end{equation}
CV fold indices are precomputed once and held fixed so that the null differs from the observed statistic only in label assignment.

\paragraph{Null calibration.}
A signal-presence gate is applied first. Estimate a null by permuting outcome labels $T=200$ times, retraining all three probes using the same CV folds, and recording $\text{AUROC}_{\max,\text{null}}^{(s,t)}$. The permutation $p$-value uses the standard $+1$ correction:
\begin{equation}
p = \frac{1 + \sum_{t=1}^{T} \mathbf{1}\!\left(\text{AUROC}_{\max,\text{null}}^{(s,t)} \geq \text{AUROC}_{\max}^{(s)}\right)}{T + 1}.
\end{equation}
This calibration is essential at high $d/n$ where chance AUROC can exceed 0.5 substantially. If $p \geq 0.05$, no reliable signal exists and localization is not attempted. For $\Delta_{\text{shared}}$ itself, a permutation null (from label permutation with fixed CV folds) provides $[\text{null}_{\text{lower}}, \text{null}_{\text{upper}}]$.

\paragraph{Decision rule.}
Inference uses both a bootstrap CI (from $B=200$ bootstrap replicates of the full CV procedure) and the permutation null. Classification requires the same CI bound to satisfy two conditions. The result is \emph{shared-dominant} when $\text{CI}_{\text{lower}}(\Delta_{\text{shared}}^{(s)}) > \text{null}_{\text{upper}}$ and $\text{CI}_{\text{lower}}(\Delta_{\text{shared}}^{(s)}) > 0$, and \emph{modality-dominant} when $\text{CI}_{\text{upper}}(\Delta_{\text{shared}}^{(s)}) < \text{null}_{\text{lower}}$ and $\text{CI}_{\text{upper}}(\Delta_{\text{shared}}^{(s)}) < 0$. Otherwise localization is \emph{indeterminate}.

\paragraph{Applicability.}
Factorized models only (JIVE, DisentangledSSL). For entangled models (CCA, CLIP), $\Delta_{\text{shared}}$ is not computed and the localization stage is skipped.

\subsection{$P_{\text{transfer}}$: Cross-Cohort Predictive Agreement}
\label{app:metrics:ptransfer}

\paragraph{Diagnostic question.}
Do independently trained outcome predictors induce a consistent patient ordering on a third, held-out cohort?

\paragraph{Definition.}
Train probes independently: $f_{A'}^{(s)}$ on Cohort~A$'$ and $f_B^{(s)}$ on Cohort~B, both using the signal-carrying representation component $z_{\text{sig}}^{(i,s)}$ selected by Stage~III localization outcome (shared $z_c$ as conservative fallback if indeterminate). Evaluate both on Cohort~C and compute
\begin{equation}
P_{\text{transfer}}^{(s)} = \text{Spearman}\!\left(f_{A'}^{(s)}(z_{\text{sig}}^{(i,s)}),\; f_B^{(s)}(z_{\text{sig}}^{(i,s)})\right)\Big|_{i \in C}.
\end{equation}

\paragraph{Null calibration.}
A one-sided permutation null is estimated by permuting outcome labels within Cohorts~A$'$ and B, retraining probes, and recomputing $P_{\text{transfer}}^{(s)}$ on Cohort~C. The 97.5th percentile defines $\text{null}_{\text{upper}}$, and a bootstrap CI ($B=200$, stratified by outcome on C) captures estimation uncertainty.

\paragraph{Decision rule.}
Predictive transfer is supported when $\text{CI}_{\text{lower}}(P_{\text{transfer}}^{(s)}) > \text{null}_{\text{upper}}$.

\paragraph{Applicability.}
All models. The representation component used for probe training is selected by Stage~III localization outcome (shared $z_c$ if shared-dominant or indeterminate, modality-specific $z_{\text{ms}}$ if modality-dominant, and full representation for non-factorized models).

\subsection{$D_{\text{quantile}}^{\text{task}}$: Outcome-Aligned Cohort Stability}
\label{app:metrics:dq}

\paragraph{Diagnostic question.}
Does the outcome-predictive ordering of patients remain stable across cohorts under distributional shift?

\paragraph{Definition.}
$D_{\text{quantile}}^{\text{task}}$ reuses probes already fitted in Stages~II/III so that no additional probe training is needed.
\begin{enumerate}[nosep]
    \item \textbf{Probe direction} $\hat{w}$. From the logistic regression coefficients $W_{\text{LR}}$ of the selected probe and per-dimension standard deviations $\hat{\sigma}$ estimated on A$'$:
    \begin{equation}
        \hat{w} = \frac{W_{\text{LR}} / \hat{\sigma}}{\|W_{\text{LR}} / \hat{\sigma}\|_2}.
    \end{equation}
    Selection follows Stage~III: shared-dominant leads to using probe on $z_c$; modality-dominant leads to using probe on $z_{\text{ms}}$; indeterminate leads to using probe on $z_c$ (conservative fallback); non-factorized leads to using probe on full representation.
    \item \textbf{Scores.} Project representations onto $\hat{w}$ without per-cohort standardization: $s_i^{(X)} = \langle z_{\text{sig}}^{(i,s)}, \hat{w}\rangle$ for $X \in \{A', B, C\}$. Each patient receives a scalar representing their position along the outcome-predictive direction.
    \item \textbf{A$'$-referenced quantiles.}
    \begin{equation}
        q_i^{(X)} = F_{A'}(s_i^{(X)}) = \frac{|\{j \in A' : s_j^{(A')} \leq s_i^{(X)}\}|}{|A'|}, \qquad X \in \{B, C\}.
    \end{equation}
    A$'$ provides a fixed reference scale so that B and C patients are expressed in the same units. Comparing B to C (rather than either to A$'$) isolates sensitivity to shift magnitude along the probe direction, since both cohorts share the same shift direction but differ in strength.
    \item \textbf{Wasserstein-1 distance.} $D_{\text{quantile}}^{\text{task}(s)} = W_1(Q^B, Q^C)$, where $Q^X = \{q_i^{(X)}\}_{i \in X}$. A large $W_1$ indicates that the cohort shift moves patients differentially along the probe direction, suggesting composition-dependent ordering.
\end{enumerate}

\paragraph{Null calibration.}
A one-sided permutation null is estimated by shuffling B/C cohort labels (preserving sizes); the probe direction $\hat{w}$ is held fixed across permutations because it is defined from A$'$, which is not permuted. With $N_{\text{perm}} = 200$, $\text{null}_{\text{upper}}$ is the 97.5th percentile. Instability is detected when $D_{\text{quantile}}^{\text{task}(s)} > \text{null}_{\text{upper}}$.

\paragraph{Decision rule.}
$D_{\text{quantile}}^{\text{task}}$ is computed unconditionally and not gated on the signal-presence permutation test. Its permutation null controls false positives independently, and cohort-level shift can be real even when individual-level AUROC is marginal.

\paragraph{Applicability.}
All models. $D_{\text{quantile}}^{\text{task}}$ detects instability when the cohort mean shift has a substantial component in the outcome-predictive direction. The metric measures deployment risk for the evaluation cohorts, not abstract confounding severity.

\section{Decision Procedure}
\label{app:decision}

\begin{figure}[H]
\centering
\includegraphics[width=0.6\linewidth,trim=0 20 0 40,clip]{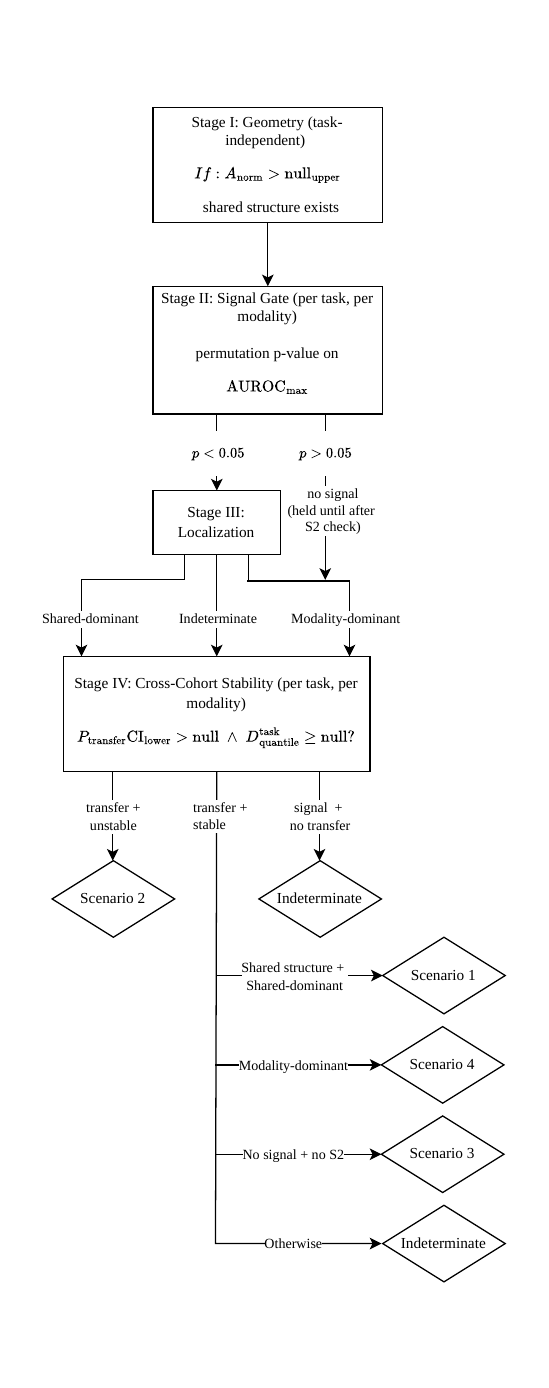}
\caption{\textbf{DECAT decision procedure.} Four stages are applied sequentially per task and per modality. Stage~I checks structural geometry (task-independent). Stage~II gates on signal presence. Stage~III localizes signal to shared or modality-specific components (factorized models only). Stage~IV evaluates cross-cohort stability via $P_{\text{transfer}}$ and $D_{\text{quantile}}^{\text{task}}$, with Scenario~2 checked first. Terminal nodes are the four diagnostic scenarios plus indeterminate.}
\label{fig:app-decision-tree}
\end{figure}

The decision procedure is a fixed, rule-based classifier with no learned parameters. It is applied \emph{per task and per modality} $s \in \{\text{HE}, \text{RNA}\}$ after metric computation. All decisions are expressed as directional comparisons to permutation null intervals (2.5th and 97.5th percentiles) or bootstrap confidence intervals, and no scenario-specific effect-size thresholds are used. Figure~\ref{fig:app-decision-tree} provides an overview of the four-stage flow and scenario assignment logic. This section specifies the exact decision rules in full.

\subsection{Stage I: Structural Geometry (Task-Independent)}
\label{app:decision:stage1}

Computed once per modality on Cohort~A$'$. $A_{\text{norm}} > \text{null}_{\text{upper}}$ indicates that patient-specific shared structure exists across modalities (necessary but not sufficient for interpreting predictive signal as shared biology). $B_{\text{norm}}$ is used as a diagnostic of interpretability stability but is not used in scenario assignment. Stage~I outputs a single flag of shared structure for Stages~III and IV.

\subsection{Stage II: Predictive Signal Existence (Per Task, Per Modality)}
\label{app:decision:stage2}

Compute the signal-presence gate (Appendix~\ref{app:metrics:delta}, Step~1): the permutation $p$-value over $\text{AUROC}_{\max}$ across the three probes. If $p \geq 0.05$ and Scenario~2 conditions (Stage~IV) are not met, assign \textbf{Scenario~3}; otherwise proceed to Stage~III.

\textbf{Ordering note:} $D_{\text{quantile}}^{\text{task}}$ (Stage~IV) is computed unconditionally, and Scenario~2 is evaluated before the signal-gate routes to Scenario~3. This is necessary because cohort-level shift can be real even when individual-level AUROC is marginal. $D_{\text{quantile}}^{\text{task}}$'s own permutation null controls false positives independently of the signal gate.

\subsection{Stage III: Predictive Localization (Factorized Models Only)}
\label{app:decision:stage3}

For factorized models (JIVE, DisentangledSSL), compute $\Delta_{\text{shared}}^{(s)}$ with bootstrap CI and permutation null. Localization requires the same CI bound to satisfy two conditions. The result is \emph{shared-dominant} when $\text{CI}_{\text{lower}}(\Delta_{\text{shared}}^{(s)}) > \text{null}_{\text{upper}}$ and $\text{CI}_{\text{lower}}(\Delta_{\text{shared}}^{(s)}) > 0$, and \emph{modality-dominant} when $\text{CI}_{\text{upper}}(\Delta_{\text{shared}}^{(s)}) < \text{null}_{\text{lower}}$ and $\text{CI}_{\text{upper}}(\Delta_{\text{shared}}^{(s)}) < 0$. If shared-dominant, an additional guard requires $A_{\text{norm}} > \text{null}_{\text{upper}}$. Otherwise the shared-dominant signal is rejected as a factorization artifact and localization is recorded as \emph{indeterminate}. For non-factorized models (CCA, CLIP), Stage~III is skipped.

\subsection{Stage IV: Cross-Cohort Stability (Per Task, Per Modality)}
\label{app:decision:stage4}

Stage~IV reuses probes and component selections from Stages~II/III.

\paragraph{Representation component.} Selected by Stage~III outcome. Shared-dominant uses $z_c$, modality-dominant uses $z_{\text{ms}}$, and indeterminate falls back to $z_c$ as a conservative default. Non-factorized models use the full representation.

\paragraph{Scenario assignment rules.} Applied in order:

\begin{enumerate}
    \item \textbf{Scenario~2 (checked first).} If $\text{CI}_{\text{lower}}(P_{\text{transfer}}^{(s)}) > \text{null}_{\text{upper}}$ \emph{and} $D_{\text{quantile}}^{\text{task}(s)} \geq \text{null}_{\text{upper}}$, predictive signal transfers functionally but predictive ordering is unstable across cohorts. \textbf{Assign S2.}
    \item \textbf{Scenario~1.} If not S2, \emph{and} $A_{\text{norm}} > \text{null}_{\text{upper}}$, \emph{and} signal-gate $p < 0.05$, \emph{and} (for factorized models) shared-dominant localization, \emph{and} $\text{CI}_{\text{lower}}(P_{\text{transfer}}^{(s)}) > \text{null}_{\text{upper}}$, \emph{and} $D_{\text{quantile}}^{\text{task}(s)} < \text{null}_{\text{upper}}$: \textbf{Assign S1.}
    \item \textbf{Scenario~4.} If not S2 or S1, \emph{and} signal-gate $p < 0.05$, \emph{and} (for factorized models) modality-dominant localization, \emph{and} $\text{CI}_{\text{lower}}(P_{\text{transfer}}^{(s)}) > \text{null}_{\text{upper}}$, \emph{and} $D_{\text{quantile}}^{\text{task}(s)} < \text{null}_{\text{upper}}$: \textbf{Assign S4.}
    \item \textbf{Scenario~3.} If signal-gate $p \geq 0.05$ and Scenario~2 not assigned: \textbf{Assign S3.}
    \item \textbf{Indeterminate.} Any case not matching the above: \textbf{Assign $\varnothing$.}
\end{enumerate}

\subsection{Cross-Modality Scenario Interactions}
\label{app:decision:cross}

Scenarios are assigned per modality. The framework permits mixed outcomes across modalities for the same task: S1/S1 (both modalities carry cross-modally supported, transferable signal), S1/S3 (shared signal exists but one assay does not capture it), S4/S4 (modality-specific signal transfers within each modality), S1/S2 (shared signal in one modality, cohort-specific artifact in the other), S1/S4 (shared signal in one modality, additional modality-specific signal in the other). For non-factorized models, S1 and S4 cannot co-occur across modalities because localization is not available. The procedure distinguishes them operationally via $A_{\text{norm}}$, which S1 requires and S4 does not.

\section{Synthetic Experimental Details}
\label{app:experiments}

This section specifies the full synthetic experimental protocol including measurement regimes, proxy configurations, outcome configurations, the pre-experiment calibration, and sample size selection.

\subsection{Two-Phase Experiment Design}
\label{app:experiments:phases}

Each synthetic run comprises two decoupled phases:
\begin{enumerate}
    \item \textbf{Representation learning} (Cohort~A only, unsupervised, outcome-agnostic).
    \item \textbf{Outcome-scenario evaluation} (linear probes on Cohorts~A$'$, B, C across multiple outcome configurations).
\end{enumerate}
This separation enables evaluating many biological scenarios on a single frozen representation without retraining, improving computational efficiency.

\subsection{Representation-Generating Parameter Space}
\label{app:experiments:params}

Each synthetic run is defined by a joint configuration of measurement regime $\beta$ and proxy configuration $(\gamma_h, \gamma_r, \eta)$. Proxy parameters affect data generation (and therefore representation learning) while outcome parameters $\alpha$ are introduced only at evaluation.

\paragraph{Measurement regimes.}
Four regimes are defined (Appendix~\ref{app:simulator:measurement}, Table~\ref{tab:app-regimes}). Each non-baseline regime doubles exactly one coefficient and the single-parameter-change design isolates the effect of each signal source.

\paragraph{Proxy configurations.}
Eight configurations spanning $(\gamma_h, \gamma_r, \eta)$ (Table~\ref{tab:app-proxy-grid}). Six use single-modality proxy ($\gamma_r = 0$, H\&E contamination only) and two use dual-modality proxy ($\gamma_h = \gamma_r > 0$) to test Scenario~2D.

\begin{table}[h]
\centering
\small
\caption{\textbf{Proxy configuration grid.} $\gamma$ controls proxy magnitude; $\eta$ controls geometric alignment with the shared batch axis ($\eta=0$: aligned, non-identifiable via geometry; $\eta > 0$: misaligned, detectable via $D_{\text{quantile}}^{\text{task}}$ under aligned cohort shift).}
\label{tab:app-proxy-grid}
\begin{tabular}{lllll}
\toprule
$\gamma_h$ & $\gamma_r$ & $\eta$ & \textbf{Scenario coverage} & \textbf{Proxy regime} \\
\midrule
0.3 & 0.0 & 0.3 & 2A weak                  & single-modality, misaligned \\
0.5 & 0.0 & 0.0 & 2C moderate              & single-modality, aligned \\
0.5 & 0.0 & 0.3 & 2A moderate              & single-modality, misaligned \\
0.5 & 0.0 & 0.6 & 2A moderate              & single-modality, strongly misaligned \\
1.0 & 0.0 & 0.0 & 2C strong                & single-modality, aligned \\
1.0 & 0.0 & 0.3 & 2A strong                & single-modality, misaligned \\
0.5 & 0.5 & 0.0 & 2D moderate              & dual-modality, aligned \\
1.0 & 1.0 & 0.0 & 2D strong                & dual-modality, aligned \\
\bottomrule
\end{tabular}
\end{table}

\subsection{Outcome Configurations}
\label{app:experiments:outcomes}

Outcome configurations ($\alpha$ values) are organized into four groups covering pure global scenarios, transition (indeterminate) regimes, mixed-modality scenarios, and proxy calibration sweeps. All baseline-runs use 27 configurations while proxy-entangled runs use 21 (17 proxy-specific plus 4 Scenario~1 configurations to measure false shared claim rate under proxy contamination).

\paragraph{Pure global scenarios.}
Scenario~1: $\alpha_s \in \{0.25, 0.5, 0.8, 1.1\}$ with all other $\alpha = 0$ and $\gamma = 0$ (4 configs; explained variance 20--83\%). Scenario~2B (outcome-level confounding): $\alpha_b \in \{0.25, 0.5, 0.8, 1.1\}$ with all other $\alpha = 0$ (4 configs). Scenario~2A (feature-level proxy, misaligned): $\alpha_m > 0$ with $\gamma = 0.5, \eta = 0.3$ (4 configs, 2 per modality). Scenario~2C (aligned proxy): $\alpha_m = 0.8$ with $\gamma \in \{0.5, 1.0\}, \eta = 0$ (2 configs). Scenario~2D (dual-modality proxy): $\alpha_h = \alpha_r = 0.8$ with $\gamma_h = \gamma_r \in \{0.5, 1.0\}, \eta = 0$ (2 configs). Scenario~4: $\alpha_m \in \{0.25, 0.5, 0.8, 1.1\}$ with $\gamma = 0$ (4 configs, 2 per modality). Scenario~3: $\alpha = 0$ (2 resamples).

\paragraph{Transition regimes.}
Transition~A (biological ambiguity): $\alpha_s \approx \alpha_h \approx \alpha_r > 0$ with $\alpha_b = 0, \gamma = 0$; magnitudes from $\{0.4, 0.6\}$ (3 configs). Transition~B (confounded ambiguity): Transition~A plus $\alpha_b = 0.3$ and $\gamma = 0.3, \eta = 0.3$ (3 configs). These configurations test whether DECAT correctly returns indeterminate when ground-truth interpretation is impossible.

\paragraph{Mixed-modality scenarios.}
Four mixed types with dominant/secondary signal levels from $\{0.8, 1.1\}$ and $\{0.3, 0.5\}$ (8 configs total): Mixed~A (shared + modality-specific; $\alpha_s + \alpha_h$), Mixed~B (two modality-specific; $\alpha_h + \alpha_r$), Mixed~C (shared + proxy; $\alpha_s + \alpha_h$ with $\gamma = 0.5, \eta = 0.3$), Mixed~D (shared + confounding; $\alpha_s + \alpha_b$). These induce cross-modal scenario mismatches (e.g., S1 for RNA, S4 for H\&E) that test DECAT's per-modality discrimination.

\paragraph{Proxy calibration sweep.}
Four configurations holding $\alpha_m = 0.8$ fixed while varying $(\gamma, \eta)$: $(0, 0), (0.5, 0), (0.5, 0.3), (0.5, 0.6)$, isolating the effect of proxy alignment from proxy strength.

\subsection{Pre-Experiment Calibration}
\label{app:experiments:calibration}

Two pre-experiments establish training protocols before the main experiment.

\paragraph{Pre Step~A: alignment saturation.}
Determine the minimal $N_{\text{train}}$ for which $A_{\text{norm}}^{*}$ (normalized cross-modal alignment on held-out Cohort~A$'$) stabilizes. Sweep $k \in \{5, 10, 50, 100\}$ and $N_{\text{train}} \in \{10k, 20k, 30k, 40k, 50k\}$ across all 7 model configurations (4 DSSL, CLIP, JIVE, CCA) with 5 seeds each. Cohort~A is generated once at $N_{\text{train}}^{\max} = 50k$ and smaller sizes are obtained by subsampling without replacement. Saturation is assessed by checking whether consecutive $N_{\text{train}}$ values yield overlapping 95\% bootstrap CIs of $A_{\text{norm}}^{*}$. Most models across most regimes satisfy this criterion by $N_{\text{train}} = 30k$, with the modality-dominant regime showing the slowest convergence (see Section~\ref{app:further-results:calibration}). For the main experiment ($k=50$), we use $N_{\text{train}} = 30k = 1500$.

\paragraph{Pre Step~A.2: epoch convergence.}
Validate that fixed epoch counts for nonlinear models (CLIP, DisentangledSSL) fall in the converged regime at the $N_{\text{train}}$ selected above. Train 5 configurations (CLIP + 4 DSSL diagonal) for 5 seeds, tracking $A_{\text{norm}}^{*}$ every 25 epochs. Default epoch counts are validated if $A_{\text{norm}}^{*}$ plateaus before the final epoch. Closed-form models (CCA, JIVE) require no convergence validation.

\subsection{Sample Sizes and Scenario Identifiability Metrics}
\label{app:experiments:sizes}

\paragraph{Evaluation sample size.} We sweep $N_{\text{eval}} \in \{50, 100, 200, 500, 1000\}$ for each cohort (A$'$, B, C). Cohorts are generated at $\max(N_{\text{eval}}) = 1000$ and smaller sizes are obtained by stratified subsampling. All other parameters are held fixed.

\paragraph{Reporting metrics.} Let $S^{*}$ denote the ground-truth scenario and $\hat{S}$ the DECAT-assigned scenario. We report the following as functions of $N_{\text{eval}}$: \emph{strict accuracy} $\mathbb{P}(S^{*} \in \hat{S})$; \emph{conservative accuracy} $\mathbb{P}(S^{*} \in \hat{S}\; \text{or}\; \hat{S} = \varnothing)$, in which indeterminate is not penalized; \emph{false shared claim rate} (FSCR) $\mathbb{P}(\hat{S} = \{1\} \wedge S^{*} \ne 1)$, the key safety metric capturing how often the framework assigns S1 when the ground truth is not S1; \emph{indeterminate sensitivity} $\mathbb{P}(\hat{S} = \varnothing \mid S^{*} = \varnothing)$, measuring whether the framework correctly hedges on fundamentally ambiguous configurations; \emph{cross-modality resolution accuracy}, the probability that both modalities are simultaneously classified correctly on tasks where the two modalities have different ground-truth scenarios.

\subsection{Total Experimental Scale}
\label{app:experiments:scale}

Main experiment: $R = 365$ synthetic runs (125 baseline + 240 proxy-entangled). Each run trains 7 representations (4 DSSL + CLIP + JIVE + CCA), for a total of 2555 representation trainings. Evaluation covers 27 configurations for baseline runs and 21 for proxy runs, across 5 evaluation sizes, yielding 294{,}525 evaluation calls (589{,}050 modality-task records).

\section{TCGA Validation Experimental Details}
\label{app:tcga}

\subsection{Dataset and Embeddings}
\label{app:tcga:data}

We applied DECAT to paired histopathology and transcriptomic embeddings from 8{,}979 TCGA patients spanning 32 cancer types. Only diagnostic (FFPE) slides were used and for patients with multiple diagnostic slides, one was selected at random. Patients without both modalities available were excluded, yielding the 8{,}979-patient paired cohort used throughout.

\paragraph{H\&E image preprocessing.}
For each H\&E slide, tissue regions were identified using HSV-based colour thresholding and non-tissue tiles were excluded based on a minimum tissue coverage threshold of 10\% per tile. Tile sizes were matched to each model's pretraining resolution: $512 \times 512$ pixels at 0.5~$\mu$m/pixel for TITAN, and $224 \times 224$ pixels at 0.5~$\mu$m/pixel for the four additional pathology foundation models used in the unimodal comparison (Section~\ref{app:tcga:unimodal}): CONCHv1.5, UNIv2, H-Optimus-0, and OpenMidnight. For multimodal experiments, H\&E embeddings were obtained from TITAN~\cite{dingMultimodalWholeslideFoundation2025}, a slide-level multimodal foundation model whose native aggregation head produces a single 768-dimensional slide-level embedding per patient. For patch-level models (CONCHv1.5, UNIv2, H-Optimus-0, OpenMidnight), per-tile embeddings were mean-pooled across all retained tissue tiles to yield a single per-patient embedding.

\paragraph{RNA-seq preprocessing.}
Bulk RNA-seq profiles were processed to $\log_2(\text{TPM}+1)$-normalized expression values across 19{,}177 protein-coding genes (Ensembl gene IDs). Gene expression values were converted to sample-level embeddings using the Clinical Transformer~\cite{arango-argotyPretrainedTransformersApplied2025} (vnBERT architecture, a value-based transformer that encodes each gene as a token--value pair; 12 layers, 16 attention heads, hidden size 1{,}024, ${\sim}$96M parameters). For each sample, genes with zero expression were removed before tokenization, and the remaining values were normalized using a robust z-score $((x - \text{median}) / (\text{MAD} + \varepsilon))$, then passed through a dual embedding layer (combining a learned token embedding with a linear projection of the scalar value) and a standard transformer encoder, with the [CLS] token's final hidden state serving as the 1{,}024-dimensional sample embedding.

\subsection{Representation Learning}
\label{app:tcga:repr}

Cohort~A (55\% of paired patients, $N = 4{,}938$) was used for unsupervised representation learning. The remaining 45\% ($N = 4{,}041$) were reserved as held-out evaluation patients and never used during training. All four model classes (CCA, CLIP, JIVE, and DisentangledSSL at four disentanglement strengths) were trained on Cohort~A using the same architecture and hyperparameters as the synthetic experiment, with the only structural difference being that input dimensions are no longer equal across modalities. H\&E embeddings have $d_h = 768$ (TITAN) and RNA embeddings have $d_r = 1{,}024$ (Clinical Transformer), compared to $d_h = d_r = 1{,}024$ in the synthetic experiment. For nonlinear models (CLIP, DisentangledSSL), each modality is encoded by a two-layer MLP with hidden width $4 \times d_m$, ReLU activation, and L2-normalized output, yielding hidden dimensions of 3{,}072 for H\&E and 4{,}096 for RNA. The output latent dimension is $K = 50$ for all models and modalities. CCA and JIVE are applied directly to the input embeddings and their projections adapt to the asymmetric input dimensions automatically. Epoch counts, batch size (256), learning rate ($10^{-3}$), and temperature ($\tau = 0.07$) are unchanged from the synthetic experiment. Encoders were frozen before evaluation. DECAT operates on the $K = 50$ (or $2K$ for factorized models) learned latent representations output by each model. The raw foundation model embeddings are used only as input to the representation learning step and are never accessed by the diagnostic metrics.

\subsection{Pooled Random Splits}
\label{app:tcga:pooled}

Fifty independent splits were drawn from the held-out 45\%, each partitioned into Cohort~A$'$ (15\%), Cohort~B (15\%), and Cohort~C (15\%) by random sampling from the pan-cancer held-out pool, retaining the natural cancer-type composition. These splits evaluate whether representations learned on Cohort~A generalize to independent evaluation cohorts. For the multimodal experiment, expected verdicts are S1 (CCA variates), S4 (JIVE RNA-residual PCs), and S3 (random labels). For the unimodal experiment (Stages~II and IV only), task-relevant labels are expected to return Transferable or S2 depending on confound strength.

\subsection{Extreme-C Splits}
\label{app:tcga:extreme}

To test S2B (confounding signal) detection, we designed an extreme-C cohort structure. Cohorts~A$'$ and B are drawn randomly from the full held-out pool, retaining the natural pan-cancer composition. Cohort~C is drawn from a label-extreme pool $E$ defined without requiring knowledge of cancer type or any other confounder. For continuous labels (e.g., TMB, Age), $E$ is defined by thresholding the prediction variable itself at the patient-level 75th percentile (e.g., patients with $\log(\text{TMB}+1)$ above p75). For binary labels (e.g., TP53), $E$ is defined as the top-10\% of held-out patients ranked by a logistic regression probe score trained on A$'$ representations. Fifty splits were generated per label. Cohort~A (training) is held fixed across all splits and only A$'$/B/C assignments vary across seeds.

\subsection{Task Labels}
\label{app:tcga:labels}

All continuous task labels are binarized at the pan-cancer median to produce binary classification targets for the linear probes. For pooled splits, task labels include CCA variates 1--50 binarized at their respective pan-cancer medians (S1 ground truth by construction), JIVE RNA-residual principal components 1--50 similarly binarized (S4 ground truth by construction), and ten random binary labels (S3). For extreme-C splits, biologically confounded labels include $\log(\text{TMB}+1)$ binarized at the pan-cancer median, TP53 somatic mutation status, and age at diagnosis binarized at the pan-cancer median. Labels are introduced only at evaluation. $\eta^2$ quantifies the fraction of label variance explained by cancer type, serving as a naive estimate of confound strength. MC3 mutation calls were used for TP53 and TMB was calculated as somatic non-synonymous mutations per megabase.

\subsection{$\alpha$-Mixture Sweep}
\label{app:tcga:alpha}

To characterize the S1-to-S2 transition as a function of cohort composition, we swept the fraction $\alpha \in \{0.0, 0.1, \ldots, 1.0\}$ of Cohort~C drawn from the extreme pool $E$ versus the non-extreme pool. The three-way split assigns one-third of held-out patients to A$'$, one-third to B, and one-third to a C-pool from which C is sampled. At $\alpha = 0$, C is drawn entirely from the non-extreme pool (patients not in $E$), which deliberately excludes label-extreme patients and is thus slightly more conservative than a fully random draw from the observed pan-cancer cohort composition. $N_C = \min(|E \cap \text{C-pool}|, 300)$ is held fixed across $\alpha$ to equate statistical power. Fifty seeds were used per (label, $\alpha$) combination.

\subsection{Power Curve}
\label{app:tcga:power}

We measured S2 detection rate as a function of evaluation cohort size $N$ under two complementary designs that address distinct practitioner questions. In the \emph{equal-N} design, A$'$ = B = C = $N$ are varied together, modeling a prospective study where all cohorts are resource-limited. In the \emph{vary-C} design, A$'$ and B are held at their full size while only C = $N$ is varied, modeling a setting where a large reference cohort exists and only the deployment cohort is limited. $N \in \{25, 50, 100, 150, 200, 300, 500, 750, 1000\}$ with 50 seeds per $N$.

\subsection{Unimodal H\&E Foundation Model Comparison}
\label{app:tcga:unimodal}

To assess DECAT's applicability to unimodal H\&E models without paired RNA, we evaluated five pretrained pathology foundation models: TITAN~\cite{dingMultimodalWholeslideFoundation2025} (slide-level, 768-dim), CONCHv1.5~\cite{luVisuallanguageFoundationModel2024} (mean-pooled over patch-level, 768-dim), UNIv2~\cite{chenGeneralpurposeFoundationModel2024} (mean-pooled over patch level, 1536-dim), H-Optimus-0~\cite{saillardHoptimus02024} (mean-pooled over patch-level, 1536-dim), and OpenMidnight~\cite{kaplanHowTrainStateoftheArt2025,karasikovTrainingStateoftheartPathology2025} (mean-pooled over patch-level, 1536-dim). Note that although CONCHv1.5 is a vision-language pretrained foundation model, and the TITAN architecture uses the CONCHv1.5 patch encoder, we refer to these as "unimodal" since they only encode one direct measurement (H\&E) of the biological sample at inference. 

Unlike the multimodal pipeline, the unimodal comparison applies DECAT directly to foundation model embeddings without an intermediate representation learning step. Stages~II and IV are applied to FM embeddings standardized using A$'$ statistics (mean and variance estimated on Cohort~A$'$ and applied to all cohorts), and raw unstandardized embeddings are passed to $D_{\text{quantile}}^{\text{task}}$, which uses A$'$-referenced quantile positions to preserve cohort mean-shift information. Logistic regression probes are trained on Cohort~A$'$ embeddings to define the probe direction for $P_{\text{transfer}}$ and $D_{\text{quantile}}^{\text{task}}$. This tests DECAT as a practitioner might use it: take an existing frozen FM, apply logistic regression probes, and run the diagnostic metrics with no training overhead beyond the probe itself. Stage~I ($A_{\text{norm}}$, cross-modal agreement) and Stage~III ($\Delta_{\text{shared}}$, signal localization) require paired modalities and are omitted. The unimodal decision maps to S2 ($P_{\text{transfer}}$ and $D_{\text{quantile}}^{\text{task}}$ both significant), Transferable ($P_{\text{transfer}}$ significant, $D_{\text{quantile}}^{\text{task}}$ stable), S3 (signal gate fails), or Indeterminate.

We evaluated 16 recurrently mutated driver genes from the MC3 consensus somatic mutation callset, selected by requiring pan-cancer prevalence ${\geq}3\%$ and a minimum of 40 patients with the minority label in at least one cancer type. For each gene, the extreme pool $E$ is defined per FM using that FM's own linear probe trained on A$'$. $E$ consists of the top-10\% of evaluation patients by probe score, requiring no knowledge of cancer type. The $\alpha$-mixture sweep varies the fraction of Cohort~C drawn from $E$ across $\alpha \in \{0.0, 0.2, 0.4, 0.6, 0.8, 1.0\}$ with $N_C = 300$ fixed. Fifty seeds were used per (gene, FM, $\alpha$) combination. For each split, the S2 flag rate is recorded across $\alpha$ values, and the S2 AUC (area under the flag rate vs $\alpha$ curve) provides a single summary statistic per gene$\times$FM combination. Post-hoc within-cancer-type AUROC collapse ($\Delta = \text{AUROC}_{\text{pan}} - \text{AUROC}_{\text{within}}$, median across cancer types) is computed independently to validate that the S2 flag rate tracks the same FM--gene combinations affected by confounding.

\section{Further Synthetic Experimental Results}
\label{app:further-results}

This section contains synthetic experimental results that complement the main paper's focus on scenario classification.

\subsection{Simulator and Metric Validation}
\label{app:further-results:validation}

Before evaluating DECAT on scenario classification, we validate that the structural metrics ($A_{\text{norm}}$, $B_{\text{norm}}$) respond correctly to controlled variation in the simulator's signal composition. In these experiments, observations are projected into the ground-truth shared latent space using the simulator's known projection matrices ($A_s^\top$ and $B_s^\top$), providing an ideal-case baseline independent of any learned representation. Each $\beta$ coefficient is swept independently while all others are held at zero.

\begin{figure}[H]
\centering
\includegraphics[width=0.85\linewidth]{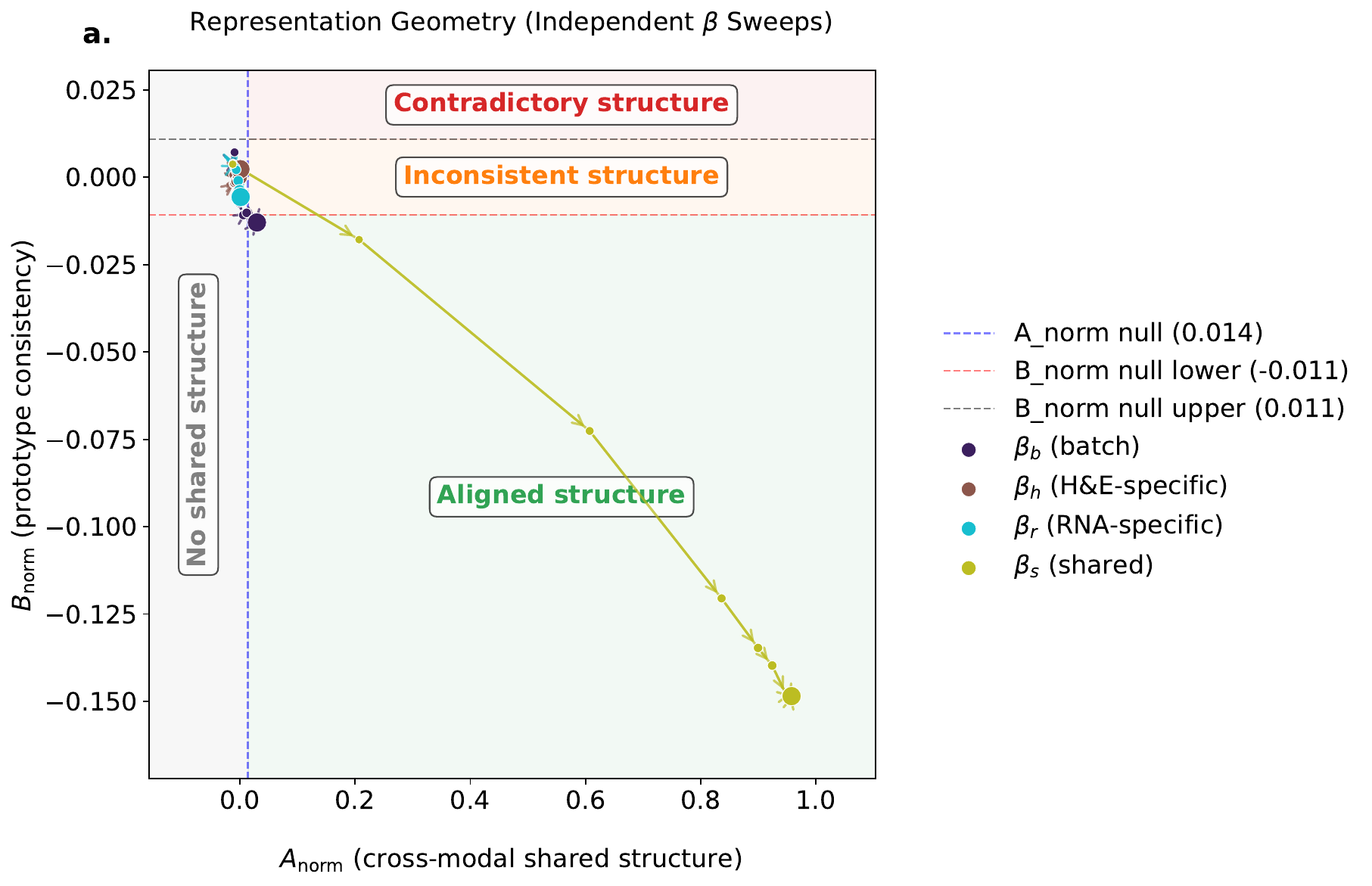}
\caption{\textbf{Representation geometry under independent $\beta$ sweeps.} Each curve varies one $\beta$ coefficient while holding others at zero; marker size increases with $\beta$ value. Null thresholds (dashed lines) are the most conservative boundaries across all $\beta$ conditions (max of per-condition 97.5th percentiles from 200 permutations). Only shared signal ($\beta_s$) drives the representation strongly into the aligned-structure region. Modality-specific signal ($\beta_h$, $\beta_r$) remains in the no-shared-structure cluster. Batch signal ($\beta_b$) at high values enters the edge of the aligned-structure region, confirming that strong batch effects can mimic shared structure in these metrics.}
\label{fig:app-geometry-sweeps}
\end{figure}

\begin{figure}[H]
\centering
\includegraphics[width=\linewidth]{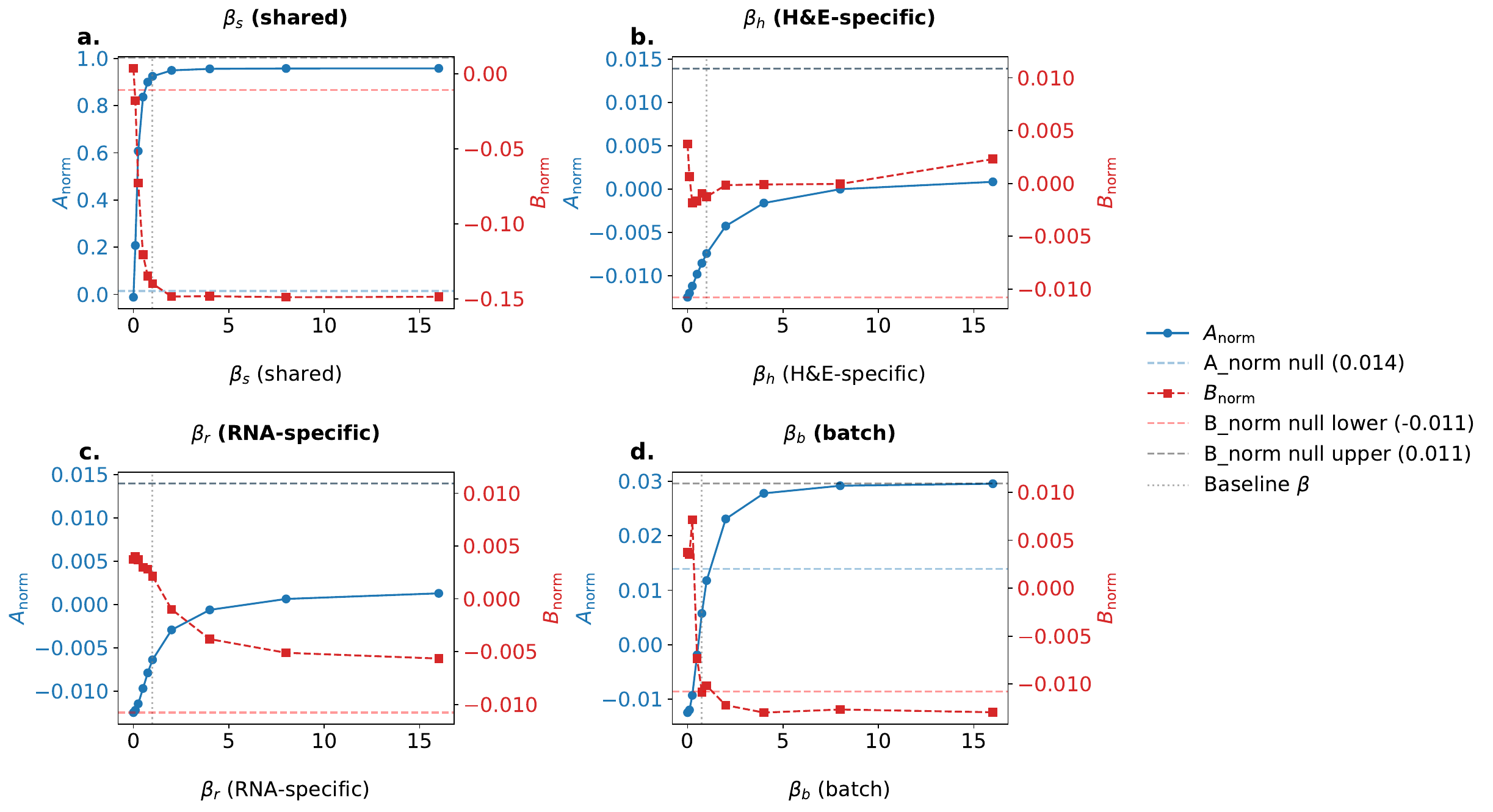}
\caption{\textbf{Per-$\beta$ sensitivity of $A_{\text{norm}}$ and $B_{\text{norm}}$.} Each panel sweeps one $\beta$ coefficient independently while holding all others at zero. Dashed lines indicate the most conservative null boundary across all $\beta$ conditions (max/min of per-condition percentiles from 200 permutations). (a)~$\beta_s$ produces strong $A_{\text{norm}}$ response and negative $B_{\text{norm}}$. (b/c)~$\beta_h$ and $\beta_r$ do not inflate $A_{\text{norm}}$. The $B_{\text{norm}}$ curve for $\beta_h$ is noisier than for $\beta_r$ because prototypes are defined from $z_c^{(h)}$, whose projections shift as $\beta_h$ varies. (d)~$\beta_b$ inflates $A_{\text{norm}}$ above null at high values.}
\label{fig:app-sensitivity}
\end{figure}

Figure~\ref{fig:app-geometry-sweeps} shows the representation geometry landscape. Shared signal ($\beta_s$) drives the representation strongly into the aligned-structure region, modality-specific signal ($\beta_h$, $\beta_r$) remains in the no-shared-structure region, and batch signal ($\beta_b$) at high values enters the edge of the aligned-structure region, mimicking shared structure. Figure~\ref{fig:app-sensitivity} shows the per-$\beta$ sensitivity curves. Two findings are notable. First, $A_{\text{norm}}$ rises above its null threshold at high $\beta_b$ (Figure~\ref{fig:app-sensitivity}d): both shared biology and batch signal produce statistically significant cross-modal agreement, though at very different magnitudes. Since the decision tree uses a binary null test rather than a magnitude threshold (which would not be comparable across model architectures), this confirms that structural metrics alone cannot reliably distinguish Scenario~1 from Scenario~2, motivating the outcome-directed metrics ($\Delta_{\text{shared}}$, $P_{\text{transfer}}$, $D_{\text{quantile}}^{\text{task}}$). Second, modality-specific signal ($\beta_h$, $\beta_r$) does not inflate $A_{\text{norm}}$ even at large values (Figure~\ref{fig:app-sensitivity}b/c), meaning the S1/S4 distinction cannot be made from geometry alone and requires predictive localization via $\Delta_{\text{shared}}$.

\begin{figure}[H]
\centering
\includegraphics[width=\linewidth]{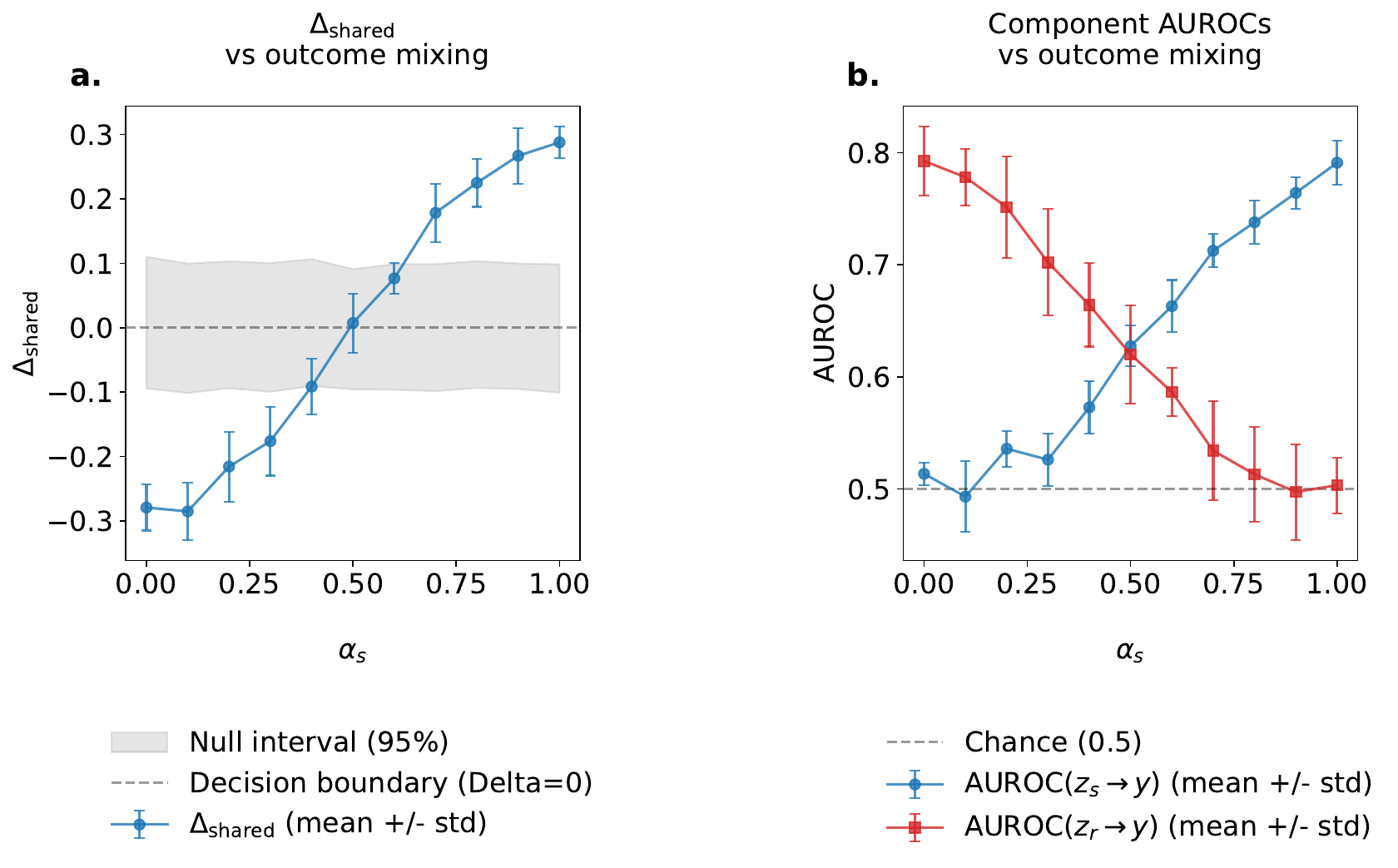}
\caption{\textbf{$\Delta_{\text{shared}}$ validation via outcome mixing sweep.} $\alpha_s$ controls the fraction of outcome signal from the shared latent ($\alpha_r = 1 - \alpha_s$). (a)~$\Delta_{\text{shared}}$ transitions from modality-dominant (negative) to shared-dominant (positive) as shared signal increases. Gray band: 95\% permutation null interval (200 permutations, averaged across seeds). Error bars: mean $\pm$ std across 5 seeds. (b)~Component AUROCs cross over, confirming the mechanism.}
\label{fig:app-delta-alpha}
\end{figure}

Figure~\ref{fig:app-delta-alpha} validates $\Delta_{\text{shared}}$ by sweeping the outcome mixing parameter $\alpha_s$ from 0 (outcome driven entirely by modality-specific signal) to 1 (outcome driven entirely by shared signal), with $\alpha_r = 1 - \alpha_s$. Cross-validated linear probes are trained on the ground-truth latents ($z_s$ and $z_r$) from Cohort~A$'$, in a clean two-signal setting ($\beta_s = \beta_r = 1$, $\beta_h = \beta_b = 0$). As $\alpha_s$ increases, $\Delta_{\text{shared}}$ transitions smoothly from negative (modality-specific dominant) through the null interval to positive (shared dominant), and the component AUROCs cross over, confirming that the metric correctly tracks the source of predictive signal.

\begin{figure}[H]
\centering
\includegraphics[width=\linewidth]{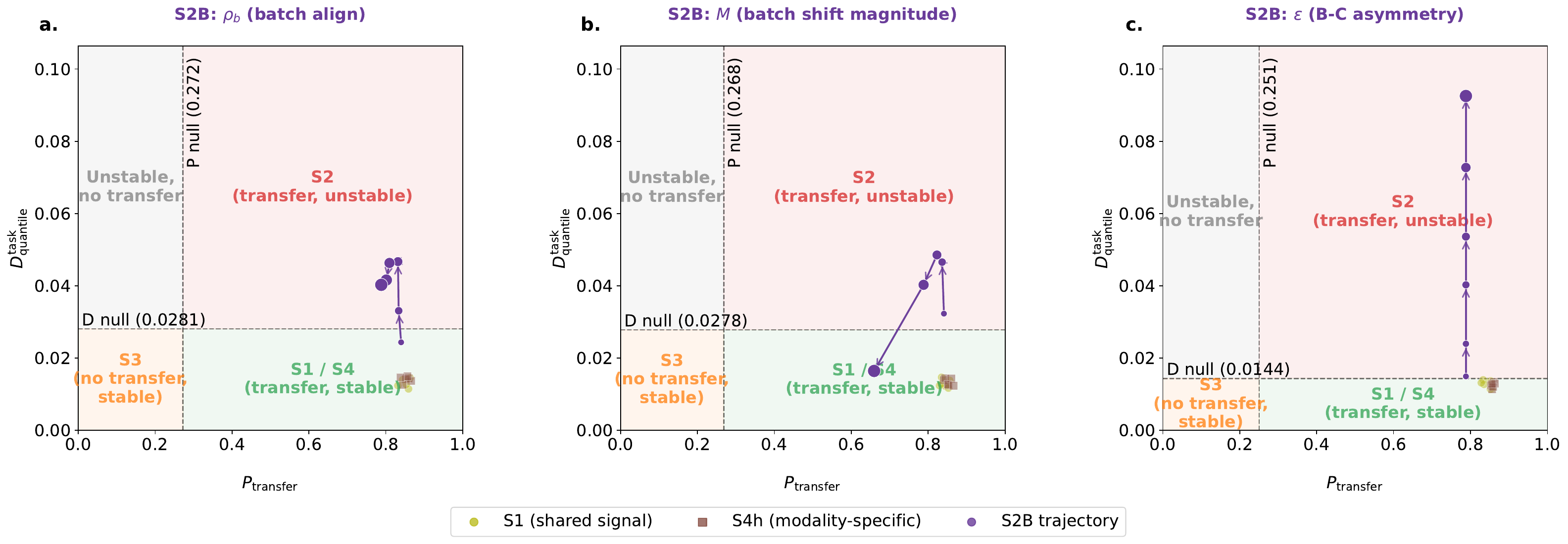}
\caption{\textbf{$D_{\text{quantile}}^{\text{task}}$ and $P_{\text{transfer}}$ validation on ground-truth latents.} S2B trajectories through the metric space under three parameter sweeps. Quadrant boundaries are the most conservative permutation null thresholds (max across conditions, 200 permutations per condition). S1 (olive) and S4h (brown) remain in the stable-transfer region. S2B (purple) enters the unstable-transfer region as cohort-shift parameters increase, with non-monotonic behavior at extreme values due to CDF saturation. Mean across 20 seeds per point.}
\label{fig:app-dq-trajectories}
\end{figure}

Figure~\ref{fig:app-dq-trajectories} validates $D_{\text{quantile}}^{\text{task}}$ and $P_{\text{transfer}}$ jointly on ground-truth latents by sweeping three cohort-shift parameters that control S2B detectability. Each panel shows a trajectory through the $P_{\text{transfer}}$ vs.\ $D_{\text{quantile}}^{\text{task}}$ metric space, with S1 and S4h as stable reference points. As batch alignment ($\rho_b$; Figure~\ref{fig:app-dq-trajectories}a), shift magnitude ($M$; Figure~\ref{fig:app-dq-trajectories}b), or B--C asymmetry ($\varepsilon$; Figure~\ref{fig:app-dq-trajectories}c) increase, S2B moves from the S1/S4 region (transfer, stable) into the S2 region (transfer, unstable). The batch alignment and magnitude sweeps (Figure~\ref{fig:app-dq-trajectories}a,b) exhibit non-monotonic behavior at extreme values due to CDF saturation: when both B and C are pushed far into the tail of A$'$, the quantile transform compresses both to ${\approx}\,1$ and $W_1$ collapses. This regime corresponds to unrealistically large shifts outside the operating range of the framework.
Proxy-driven S2 (S2A) cannot be validated on ground-truth latents because proxy entanglement ($\gamma_h$) contaminates the observations ($x_h$) rather than the latents ($z_h$) directly. On ground-truth latents, the probe sees clean $z_h$ with no cohort shift, making S2A metrically indistinguishable from S4h. S2A validation therefore requires learned representations (Figure~\ref{fig:proxy_summary}).

\begin{figure}[H]
\centering
\includegraphics[width=\linewidth]{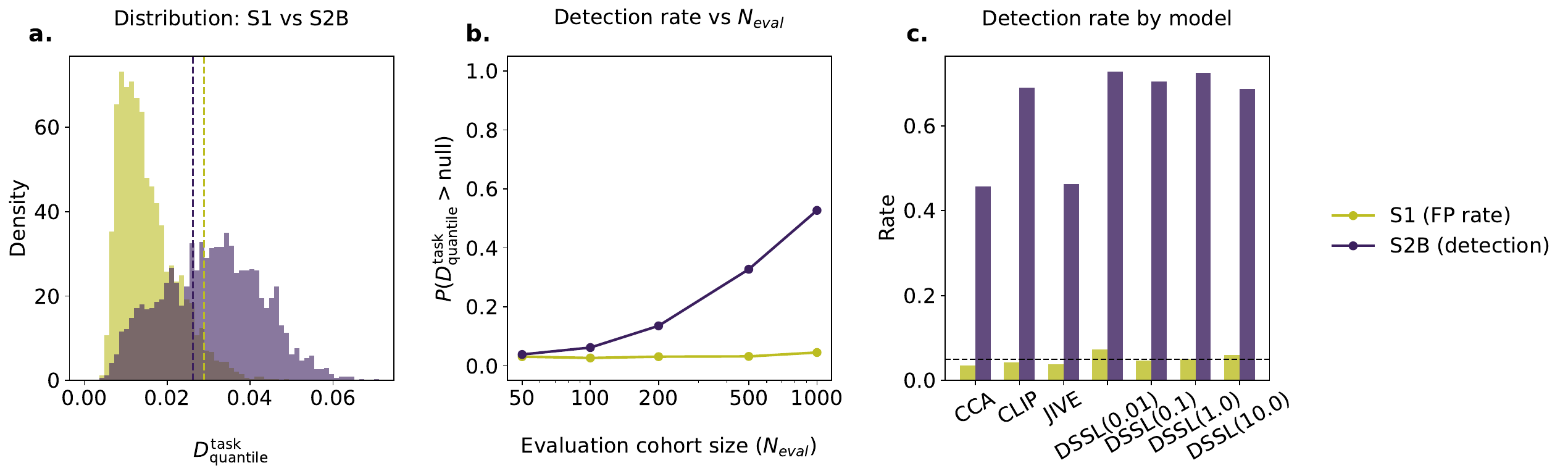}
\caption{\textbf{$D_{\text{quantile}}^{\text{task}}$ discrimination between S1 and S2B on learned representations.} (a)~Distribution of $D_{\text{quantile}}^{\text{task}}$ values for S1 and S2B at $N_{\text{eval}} = 1000$, all models pooled. Dashed lines: per-scenario null upper bounds. The bimodal S2B distribution reflects pooling across signal strengths ($\alpha_b \in \{0.25, 0.5, 0.8, 1.1\}$). (b)~Detection rate (P($D_{\text{quantile}}^{\text{task}} >$ null)) versus $N_{\text{eval}}$. S1 stays near 5\% (well-calibrated null), S2B rises with sample size. (c)~Per-model detection rates at $N_{\text{eval}} = 1000$.}
\label{fig:app-dq-discrimination}
\end{figure}

$D_{\text{quantile}}^{\text{task}}$ is most informatively validated on learned representations for the full S1/S2B discrimination, where the separation is imperfect. Figure~\ref{fig:app-dq-discrimination} provides this validation from the main experiment. Figure~\ref{fig:app-dq-discrimination}a shows that S1 and S2B produce overlapping but distinguishable distributions, with S2B exhibiting a heavier right tail past the null threshold. Figure~\ref{fig:app-dq-discrimination}b confirms that the S1 false-positive rate remains near the expected 5\% across all evaluation sample sizes while S2B detection rises with $N_{\text{eval}}$, demonstrating that the permutation null is well-calibrated. Figure~\ref{fig:app-dq-discrimination}c shows per-model detection rates: S1 false positives stay below 5\% for all models, while S2B detection varies by architecture (40--60\% at $N_{\text{eval}} = 1000$).

\begin{figure}[H]
\centering
\includegraphics[width=\linewidth]{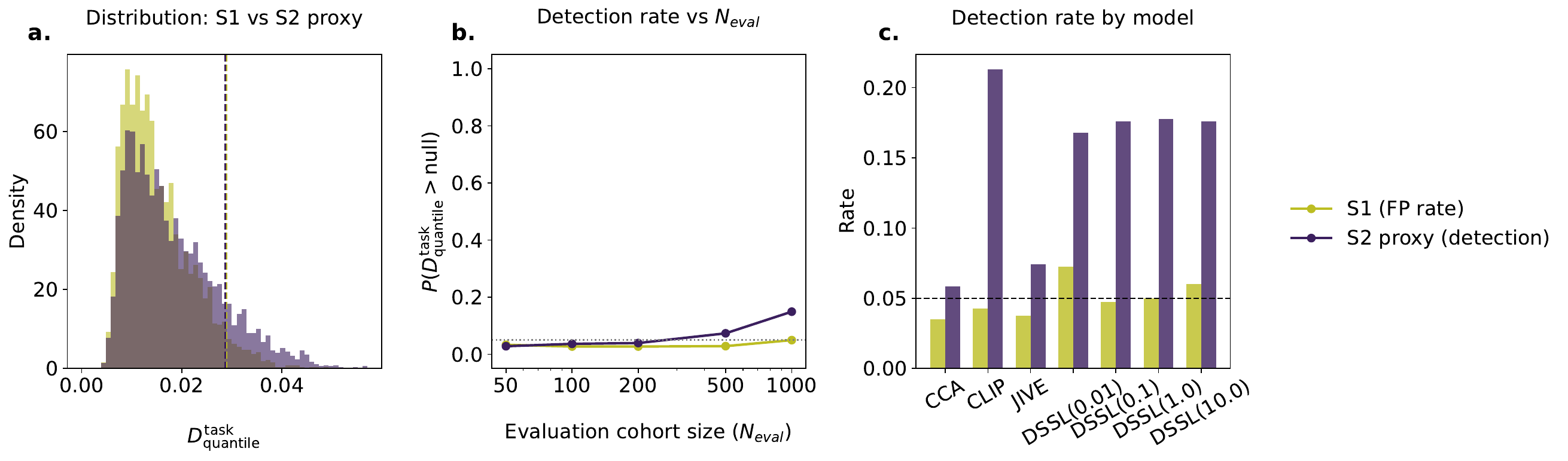}
\caption{\textbf{$D_{\text{quantile}}^{\text{task}}$: S1 vs.\ proxy S2 on learned representations.} Same format as Figure~\ref{fig:app-dq-discrimination}. (a)~Distributions overlap substantially because proxy-induced instability has a smaller absolute effect size than direct confounding. Detection relies on per-run permutation null sensitivity. Pooled across all proxy configurations.}
\label{fig:app-dq-s2a}
\end{figure}

Figure~\ref{fig:app-dq-s2a} extends this analysis to proxy-driven S2 on learned representations (pooled across all proxy configurations). Unlike direct confounding, proxy S2 produces small absolute $D_{\text{quantile}}^{\text{task}}$ values with substantial distribution overlap with S1 (Figure~\ref{fig:app-dq-s2a}a). However, the per-run permutation null is sensitive enough to detect the small but systematic proxy-induced instability and the detection rate (Figure~\ref{fig:app-dq-s2a}c) exceeds 5\% for most models. This confirms that proxy S2 detection operates through per-run null sensitivity rather than absolute effect size separation, and that the effect size is smaller than for direct confounding by design.

\begin{figure}[H]
\centering
\includegraphics[width=\linewidth]{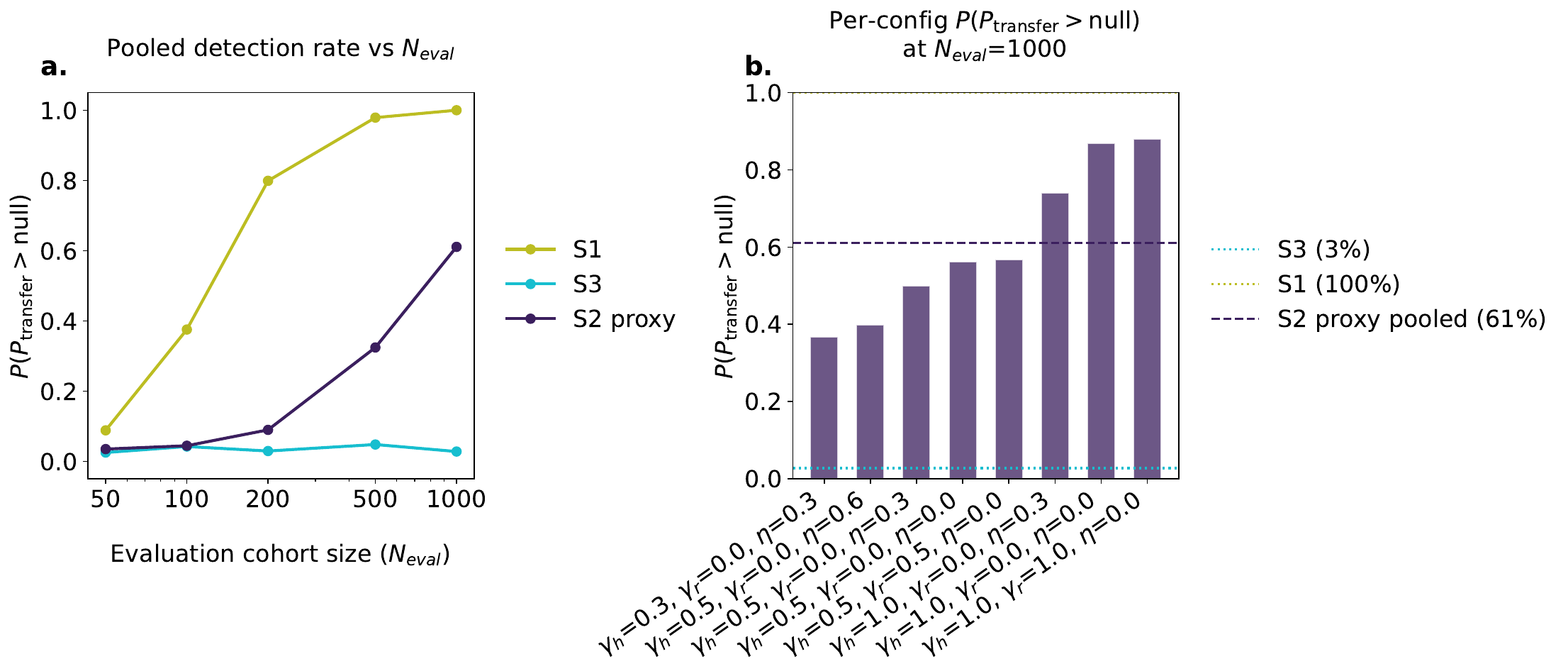}
\caption{\textbf{$P_{\text{transfer}}$ detection sensitivity for proxy S2.} (a)~Detection rate versus $N_{\text{eval}}$ for S1, S3, and proxy S2 (pooled). (b)~Detection rate at $N_{\text{eval}} = 1000$ by proxy configuration, sorted by detectability. Baseline regime, all models pooled.}
\label{fig:app-ptransfer-proxy}
\end{figure}

Figure~\ref{fig:app-ptransfer-proxy} shows $P_{\text{transfer}}$ detection sensitivity for proxy S2. At $N_{\text{eval}} = 1000$, $P_{\text{transfer}}$ is significant for 61\% of proxy S2 runs pooled across all configurations, compared to ${\approx}5\%$ for S3 (Figure~\ref{fig:app-ptransfer-proxy}a). Detectability varies substantially by proxy configuration (Figure~\ref{fig:app-ptransfer-proxy}b): strong aligned proxy ($\gamma_h = 1.0$, $\eta = 0$) reaches 87--88\%, while weak misaligned proxy ($\gamma_h = 0.3$, $\eta = 0.3$) reaches only 37\%. Runs where $P_{\text{transfer}}$ is not significant represent proxy contamination that is insufficiently coherent to produce reliable cross-cohort predictive transfer and the framework correctly returns indeterminate or S3 in these cases.

\clearpage
\subsection{Pre-Experiment Calibration Results}
\label{app:further-results:calibration}

\begin{figure}[H]
\centering
\includegraphics[width=\linewidth]{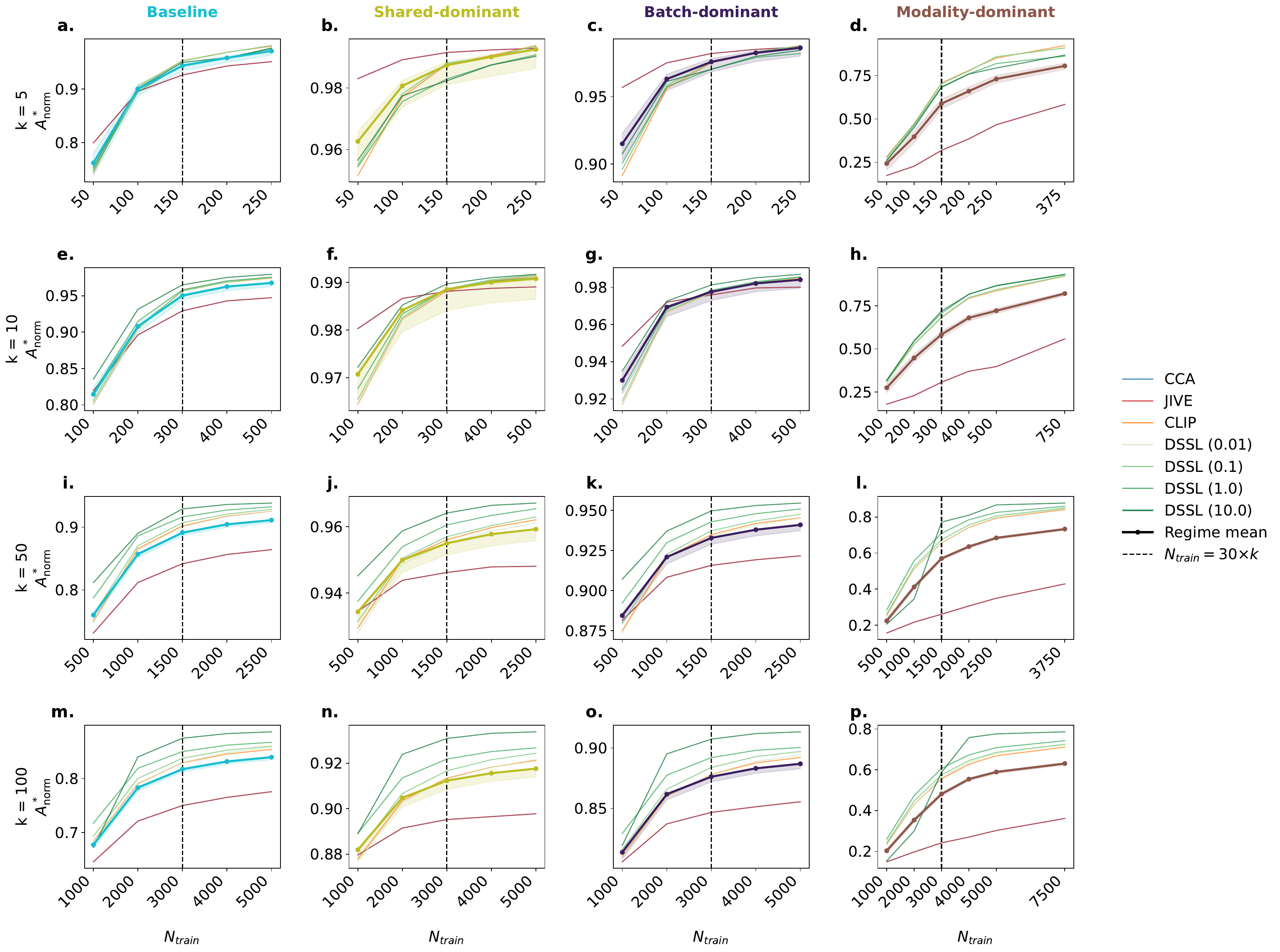}
\caption{\textbf{Pre Step~A: $A_{\text{norm}}^{*}$ saturation across measurement regimes and latent dimensionalities.} Thin lines: individual models. Thick lines: mean across all 7 models. Black dashed: $N_{\text{train}} = 30k$ recommendation used in the main experiment. All models saturate before this threshold across all regimes.}
\label{fig:app-saturation-regimes}
\end{figure}

Figure~\ref{fig:app-saturation-regimes} shows $A_{\text{norm}}^{*}$ saturation curves from Pre Step~A across all four measurement regimes and latent dimensionalities ($k \in \{5, 10, 50, 100\}$). Most models saturate by $N_{\text{train}} = 30k$ (black dashed line). The modality-dominant regime ($\beta_h = \beta_r = 2.0$) shows the slowest saturation due to weaker shared signal relative to modality-specific signal, and some models have not fully plateaued at $30k$ in this regime. We select $N_{\text{train}} = 30k$ ($= 1500$ for $k = 50$) as a practical compromise because further increasing training size yields diminishing returns for most models and regimes and the modality-dominant regime is designed to stress-test the framework.

Pre Step~A.2 validated epoch convergence for CLIP and all four DisentangledSSL configurations across all measurement regimes, confirming that $A_{\text{norm}}^{*}$ plateaus before the selected epoch counts. The final epoch counts used in the main experiment are: CLIP 200 epochs; DSSL(0.01) and DSSL(0.1) 125 shared + 50 specific epochs; DSSL(1.0) and DSSL(10.0) 75 shared + 50 specific epochs. CCA and JIVE are closed-form and require no epoch selection.

\clearpage
\subsection{Measurement Regime Effects}
\label{app:further-results:power}

\begin{figure}[H]
\centering
\includegraphics[width=\linewidth]{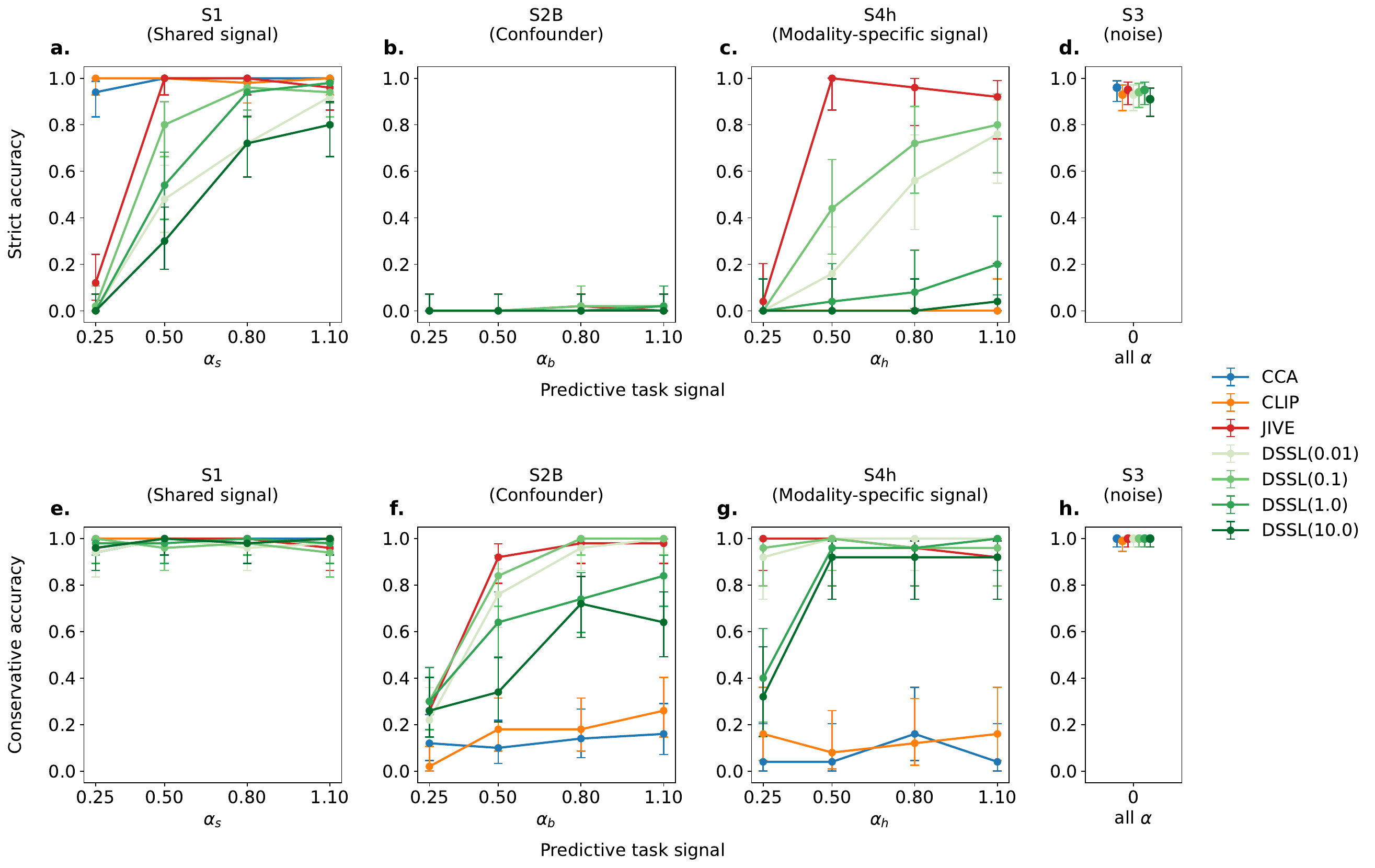}
\caption{\textbf{Detection rate versus predictive task signal, shared-dominant regime} ($\beta_s = 2.0$, $\beta_b = 0.75$). Same panel layout as Figure~\ref{fig:alpha_curves}. S1 detection improves for factorized models; S2B detection is suppressed because the stronger shared signal overwhelms the batch contribution. S1: Scenario~1, S2B: Scenario~2 (direct confounding), S4h: Scenario~4 (H\&E-specific latent), S3: Scenario~3. $N_{\text{eval}} = 1000$, 95\% Clopper--Pearson confidence intervals, 100 evaluations per point (S1, S2B, S3) and 50 per point (S4h).}
\label{fig:app-shared-dominant}
\end{figure}

\begin{figure}[H]
\centering
\includegraphics[width=\linewidth]{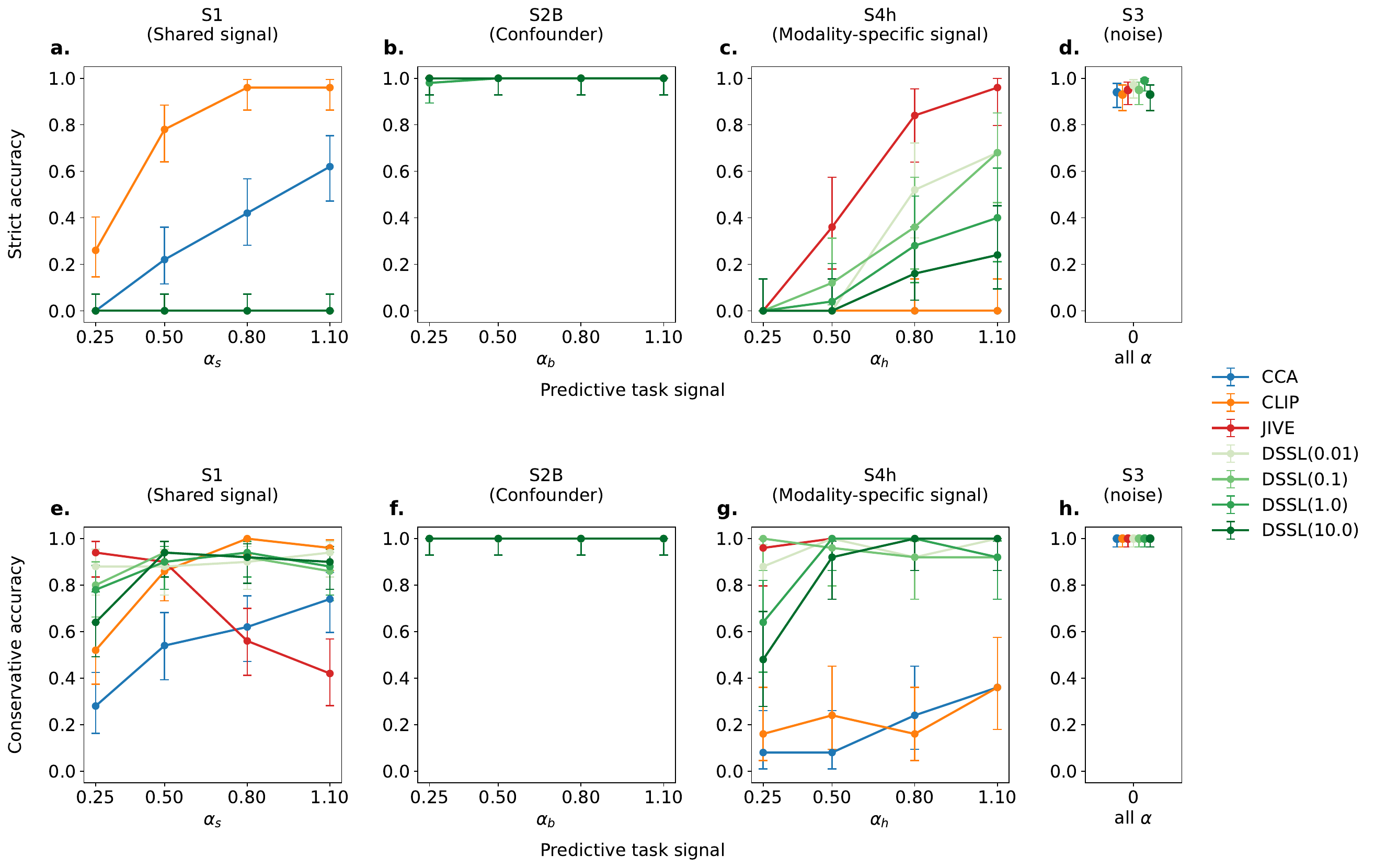}
\caption{\textbf{Detection rate versus predictive task signal, batch-dominant regime} ($\beta_b = 1.5$). Same panel layout as Figure~\ref{fig:alpha_curves}. S2B detection approaches 100\% at strong signal; S1 localization degrades for factorized models because batch signal contaminates the shared component. S1: Scenario~1, S2B: Scenario~2 (direct confounding), S4h: Scenario~4 (H\&E-specific latent), S3: Scenario~3. $N_{\text{eval}} = 1000$, 95\% Clopper--Pearson confidence intervals, 100 evaluations per point (S1, S2B, S3) and 50 per point (S4h).}
\label{fig:app-batch-dominant}
\end{figure}

\begin{figure}[H]
\centering
\includegraphics[width=\linewidth]{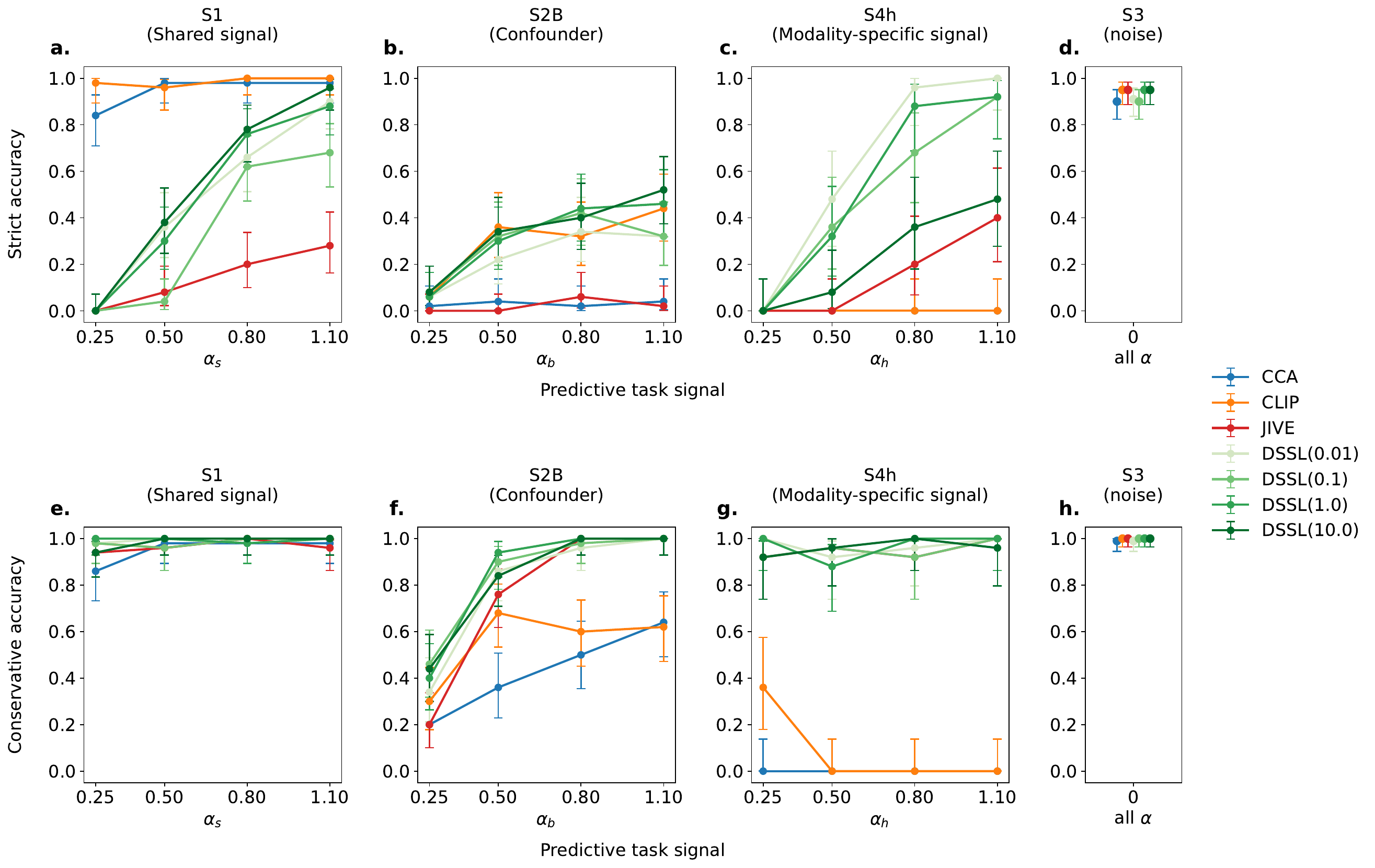}
\caption{\textbf{Detection rate versus predictive task signal, modality-dominant regime} ($\beta_h = \beta_r = 2.0$, $\beta_b = 0.75$). Same panel layout as Figure~\ref{fig:alpha_curves}. S4 detection improves relative to baseline because the stronger modality-specific signal aids localization. S1: Scenario~1, S2B: Scenario~2 (direct confounding), S4h: Scenario~4 (H\&E-specific latent), S3: Scenario~3. $N_{\text{eval}} = 1000$, 95\% Clopper--Pearson confidence intervals, 100 evaluations per point (S1, S2B, S3) and 50 per point (S4h).}
\label{fig:app-modality-dominant}
\end{figure}

Figures~\ref{fig:app-shared-dominant}--\ref{fig:app-modality-dominant} replicate the main-body analysis (Figure~\ref{fig:alpha_curves}) for the three non-baseline measurement regimes, showing strict and conservative detection rates stratified by signal strength.

The shared-dominant regime ($\beta_s = 2.0$, Figure~\ref{fig:app-shared-dominant}) improves S1 detection for factorized models (stronger shared signal aids localization; Figure~\ref{fig:app-shared-dominant}a) but S2B strict detection collapses to near zero across all models because the shared signal overwhelms the batch contribution in the representation (Figure~\ref{fig:app-shared-dominant}b). However, S2B conservative accuracy for factorized models remains high (60--100\% at strong signal; Figure~\ref{fig:app-shared-dominant}f), indicating that the framework returns indeterminate rather than misclassifying confounded signal as shared signal.
The batch-dominant regime ($\beta_b = 1.5$, Figure~\ref{fig:app-batch-dominant}) drives S2B strict detection to near 100\% across all signal strengths (Figure~\ref{fig:app-batch-dominant}b) because the stronger batch contribution improves probe-shift alignment. However, S1 strict accuracy for factorized models collapses to near zero (Figure~\ref{fig:app-batch-dominant}a) because batch signal contaminates the shared component ($z_c$), preventing $\Delta_{\text{shared}}$ from localizing shared biology. The modality-dominant regime ($\beta_h = \beta_r = 2.0$, Figure~\ref{fig:app-modality-dominant}) improves S4 detection relative to baseline (Figure~\ref{fig:app-modality-dominant}c) because the stronger modality-specific signal aids localization.
Across all regimes, disentanglement strength within DisentangledSSL modulates the trade-off between factorization quality and sensitivity: weaker configurations ($\beta_{\text{ssl}} = 0.01$) behave more like entangled models, while stronger configurations ($\beta_{\text{ssl}} = 10.0$) achieve cleaner factorization but can degrade S4 detection at high signal, possibly because aggressive compression of the shared latent discards information needed for localization.

\subsection{Detection Rate vs.\ Evaluation Sample Size}
\label{app:further-results:neval}

\begin{figure}[H]
\centering
\includegraphics[width=\linewidth]{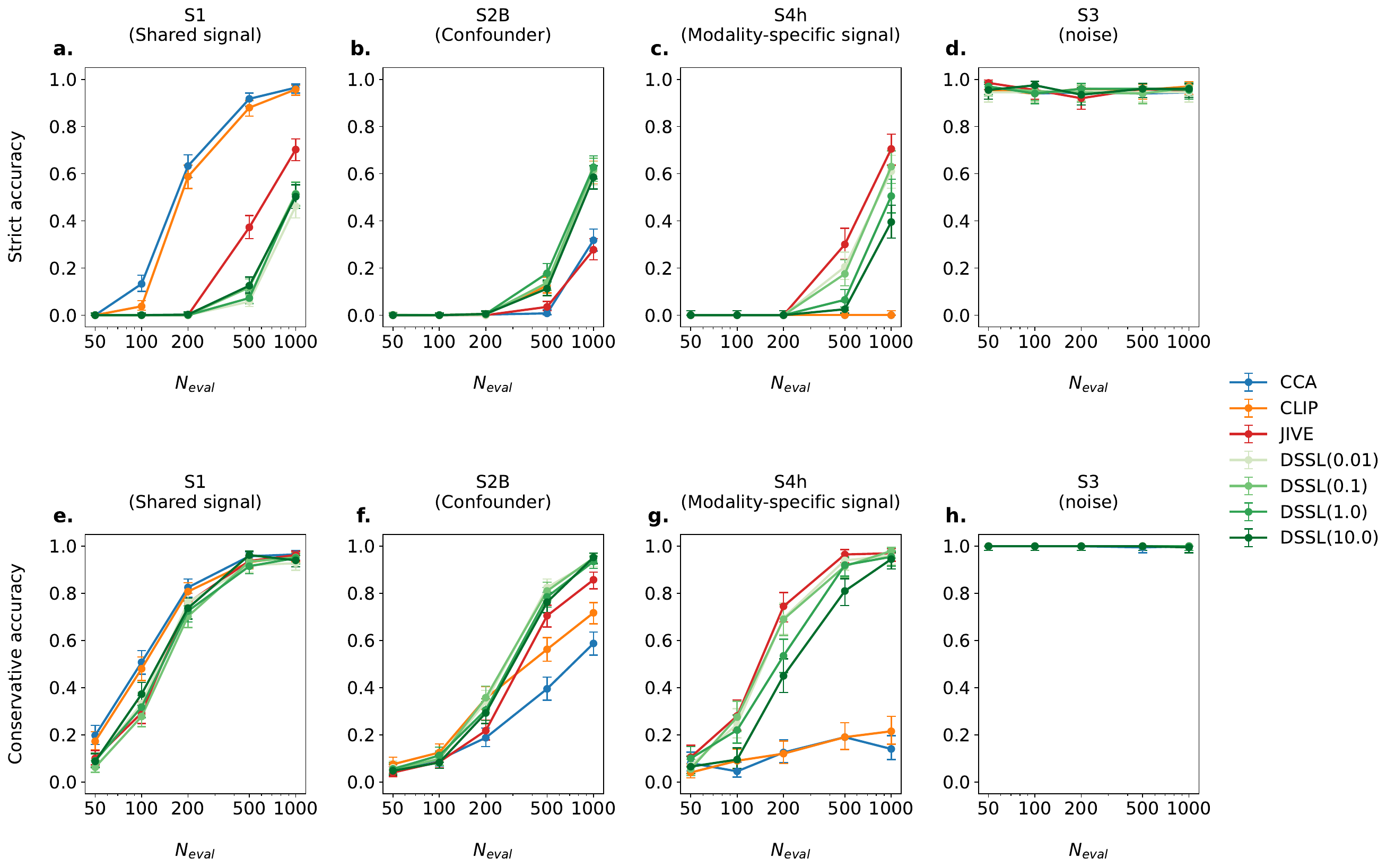}
\caption{\textbf{Detection rate versus evaluation sample size per model.} Same panel layout as Figure~\ref{fig:alpha_curves} (top row: strict accuracy, bottom row: conservative accuracy) but with $N_{\text{eval}}$ on the x-axis, pooled across all $\alpha$ values. Most metrics are still improving at $N_{\text{eval}} = 1000$, confirming that larger evaluation cohorts would improve diagnostic reliability. S1: Scenario~1, S2B: Scenario~2 (direct confounding), S4h: Scenario~4 (H\&E-specific latent), S3: Scenario~3. Baseline regime, no proxy, 95\% Clopper--Pearson confidence intervals.}
\label{fig:app-neval}
\end{figure}

Figure~\ref{fig:app-neval} shows detection rates versus $N_{\text{eval}}$ on the baseline regime, pooled across all signal strengths ($\alpha \in \{0.25, 0.50, 0.80, 1.10\}$). S2B and S4h strict accuracy curves are still rising at $N_{\text{eval}} = 1000$ (Figure~\ref{fig:app-neval}b,c), confirming that larger evaluation cohorts would improve diagnostic reliability for these scenarios. Conservative accuracy saturates earlier than strict for most scenario-model combinations (Figure~\ref{fig:app-neval}e--g), reflecting the framework's transition from indeterminate to confident classification as sample size grows. Note that this synthetic experiment uses equal-sized cohorts (A$' = $ B $=$ C $= N_{\text{eval}}$). The TCGA power curve (Figure~\ref{fig:tcga_power_curve}) reveals a more nuanced picture on real data where the primary bottleneck is probe quality in the reference cohorts (A$'$/B), and the deployment cohort C can be substantially smaller while still achieving reliable detection.

\subsection{FSCR by Measurement Regime}
\label{app:further-results:fscr-regime}

\begin{figure}[H]
\centering
\includegraphics[width=\linewidth]{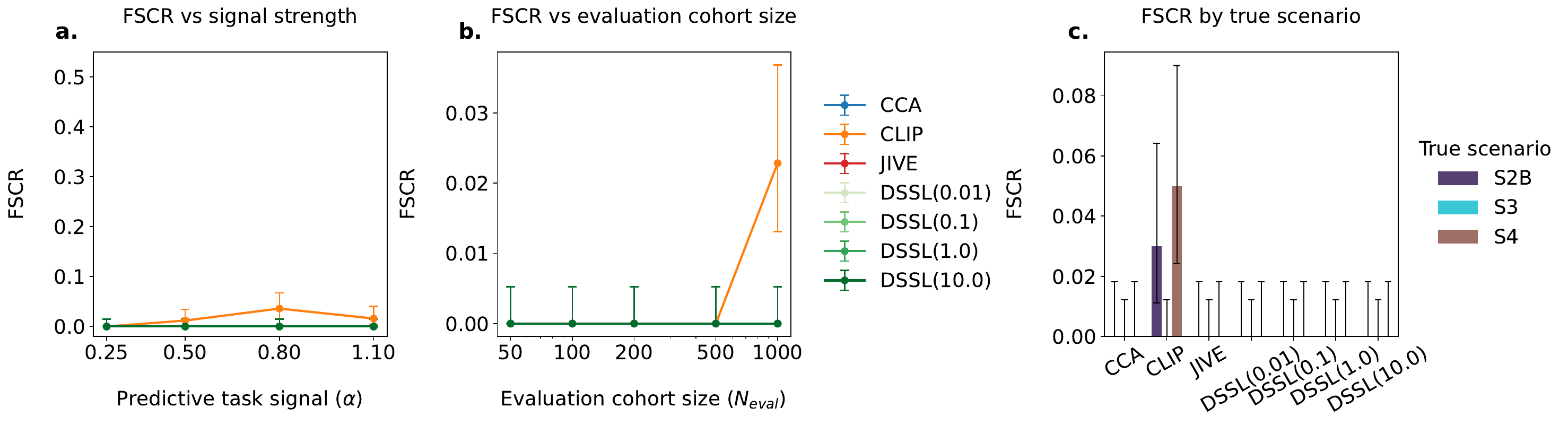}
\caption{\textbf{False shared claim rate (FSCR), shared-dominant regime} ($\beta_s = 2.0$, $\beta_b = 0.75$). Same panel layout as Figure~\ref{fig:fscr}. (a)~FSCR versus predictive task signal. (b)~FSCR versus $N_{\text{eval}}$. (c)~FSCR decomposed by true scenario. No proxy. 95\% Clopper--Pearson confidence intervals. 500 evaluations per point in (a), 1{,}400 per point in (b), 400 (S2B, S4) and 600 (S3) per bar in (c).}
\label{fig:app-fscr-shared}
\end{figure}

\begin{figure}[H]
\centering
\includegraphics[width=\linewidth]{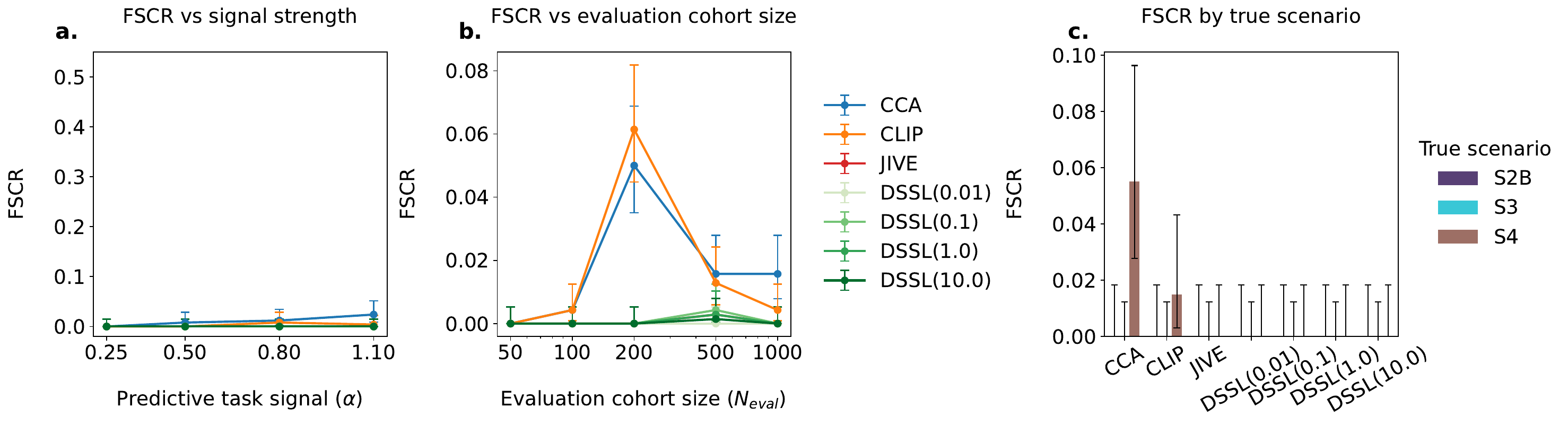}
\caption{\textbf{False shared claim rate (FSCR), batch-dominant regime} ($\beta_b = 1.5$). Same panel layout as Figure~\ref{fig:fscr}. (a)~FSCR versus predictive task signal. (b)~FSCR versus $N_{\text{eval}}$. (c)~FSCR decomposed by true scenario. No proxy. 95\% Clopper--Pearson confidence intervals. 500 evaluations per point in (a), 1{,}400 per point in (b), 400 (S2B, S4) and 600 (S3) per bar in (c).}
\label{fig:app-fscr-batch}
\end{figure}

\begin{figure}[H]
\centering
\includegraphics[width=\linewidth]{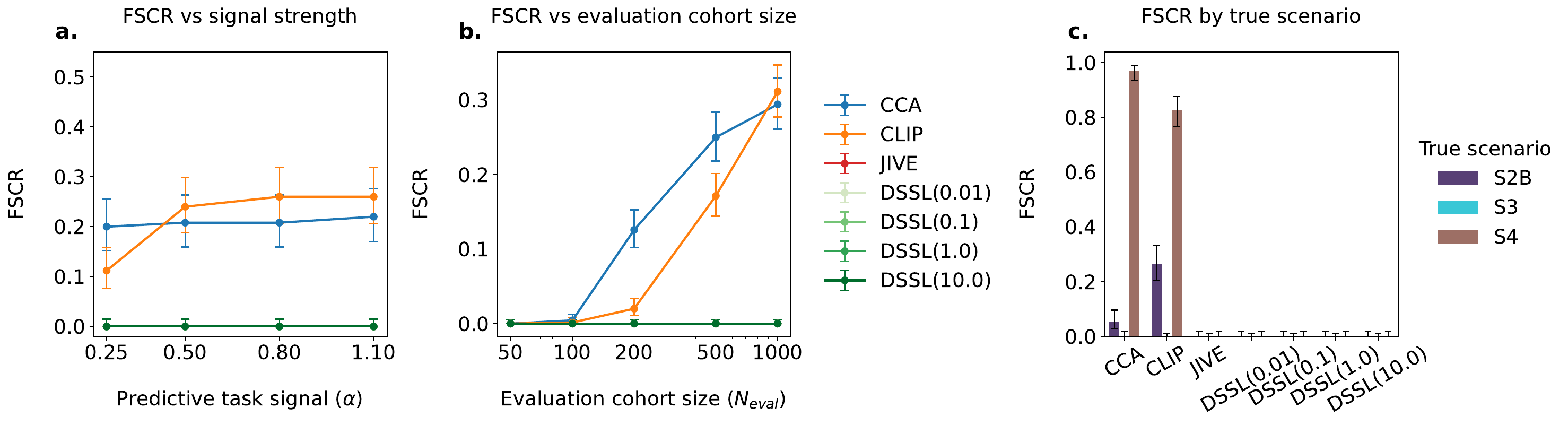}
\caption{\textbf{False shared claim rate (FSCR), modality-dominant regime} ($\beta_h = \beta_r = 2.0$, $\beta_b = 0.75$). Same panel layout as Figure~\ref{fig:fscr}. (a)~FSCR versus predictive task signal. (b)~FSCR versus $N_{\text{eval}}$. (c)~FSCR decomposed by true scenario. No proxy. 95\% Clopper--Pearson confidence intervals. 500 evaluations per point in (a), 1{,}400 per point in (b), 400 (S2B, S4) and 600 (S3) per bar in (c).}
\label{fig:app-fscr-modality}
\end{figure}

Figures~\ref{fig:app-fscr-shared}--\ref{fig:app-fscr-modality} replicate the FSCR analysis (Figure~\ref{fig:fscr}) for the three non-baseline regimes. FSCR patterns differ across regimes in interpretable ways.
In the shared-dominant regime (Figure~\ref{fig:app-fscr-shared}), FSCR for entangled models is lower than baseline because the stronger shared signal makes S2B tasks relatively less detectable (the shared signal overwhelms the batch contribution in the representation, so fewer S2B cases pass $P_{\text{transfer}}$ confidently enough to be misclassified as S1).
In the batch-dominant regime (Figure~\ref{fig:app-fscr-batch}), FSCR on S2B is near zero for all models because $D_{\text{quantile}}^{\text{task}}$ is reliably significant when batch signal is strong and the S2 check intercepts confounded signal before it can be misclassified as S1.
In the modality-dominant regime (Figure~\ref{fig:app-fscr-modality}), FSCR is elevated relative to baseline for entangled models, driven primarily by S4 misclassification. Stronger modality-specific signal provides more detectable predictive structure that entangled models attribute to shared signal.

\clearpage
\subsection{Proxy Detection by Condition}
\label{app:further-results:proxy}

\begin{figure}[H]
\centering
\includegraphics[width=\linewidth]{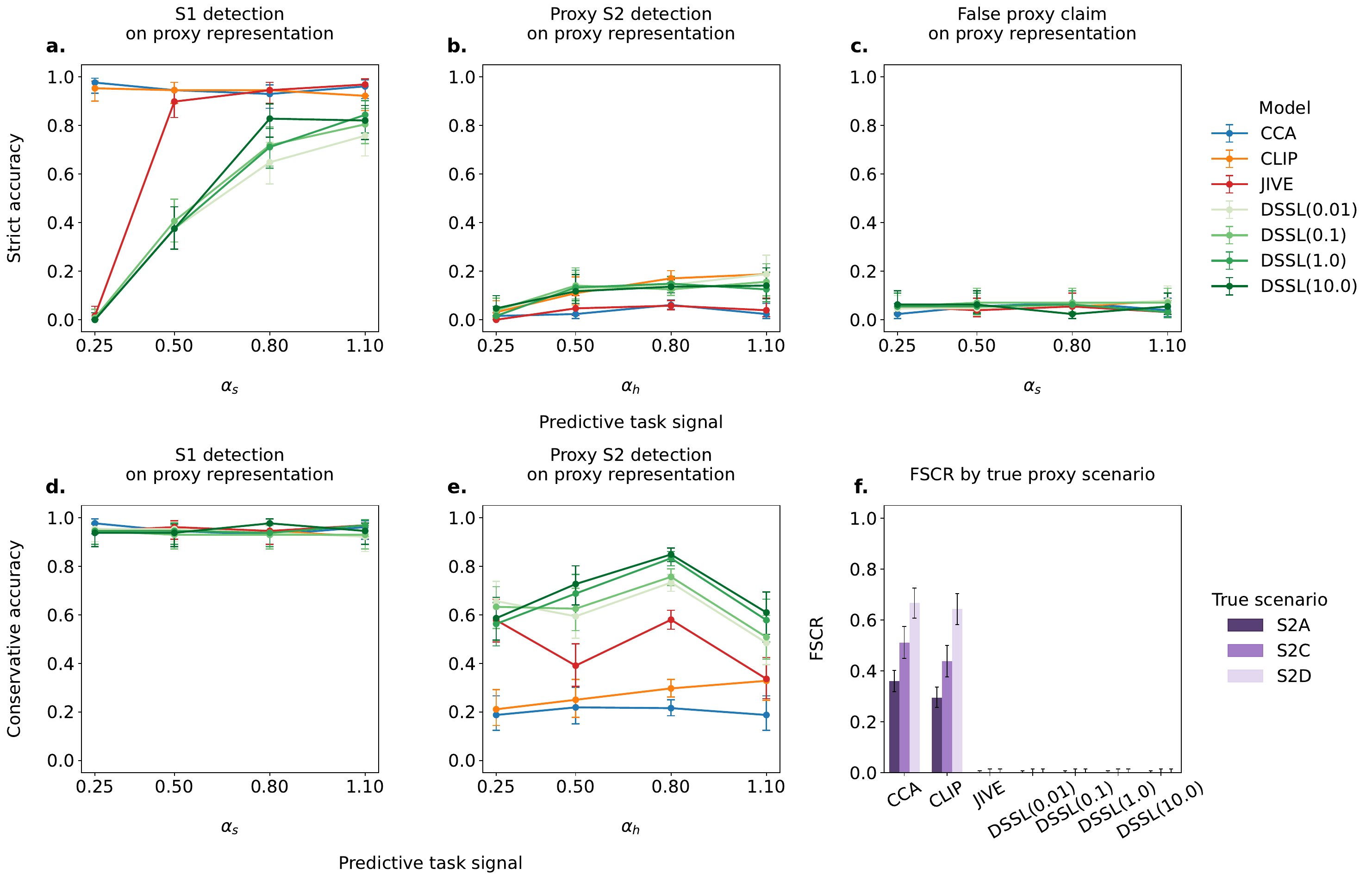}
\caption{\textbf{DECAT behavior on proxy-contaminated representations.} (a/d)~S1 detection rate on proxy-contaminated representations versus $\alpha_s$ (shared signal strength). (b/e)~Proxy S2 detection rate versus $\alpha_h$ (H\&E-specific signal strength through the proxy-contaminated pathway). (c)~False proxy claim rate versus $\alpha_s$, measuring how often genuine shared biology is falsely flagged as spurious. (f)~FSCR decomposed by proxy subtype. S2A: misaligned single-modality proxy, S2C: aligned single-modality proxy, S2D: dual-modality proxy. Top row: strict rates. Bottom row: conservative rates (a/d, b/e). Baseline regime, $N_{\text{eval}} = 1000$, pooled across all 8 proxy configurations, 95\% Clopper--Pearson confidence intervals. (a/d) and (c): 128 evaluations per point. (b/e): 128 evaluations per point except $\alpha_h = 0.80$ (640 evaluations). (f): 512 (S2A), 256 (S2C), 256 (S2D) evaluations per model.}
\label{fig:proxy_summary}
\end{figure}

Figure~\ref{fig:proxy_summary} evaluates DECAT on representations learned from proxy-entangled data.
S1 detection is largely preserved relative to clean data (Figure~\ref{fig:proxy_summary}a), with most models showing drops of less than 5\%. Conservative S1 accuracy remains above 90\% for all models (Figure~\ref{fig:proxy_summary}d).
Proxy S2 strict detection remains low at 5--20\% (Figure~\ref{fig:proxy_summary}b), consistent with the geometric challenge of detecting instability along proxy-contaminated directions. Conservative rates reach 60--75\% for factorized models at strong signal (Figure~\ref{fig:proxy_summary}e), indicating that instability is detected but the evidence is insufficient for confident S2 classification.
The false proxy claim rate (Figure~\ref{fig:proxy_summary}c) stays below 15\% across all models and signal strengths.
The FSCR decomposition by proxy subtype (Figure~\ref{fig:proxy_summary}f) reveals that dual-modality proxy (S2D) induces the highest false S1 rates for entangled models, because proxy signal entering both modalities along the same direction mimics shared structure.
All factorized models (JIVE and DSSL) maintain near-zero FSCR across all proxy subtypes (Figure~\ref{fig:proxy_summary}f), confirming that the $\Delta_{\text{shared}}$ localization gate prevents false shared claims even under proxy contamination.

\begin{figure}[H]
\centering
\includegraphics[width=\linewidth]{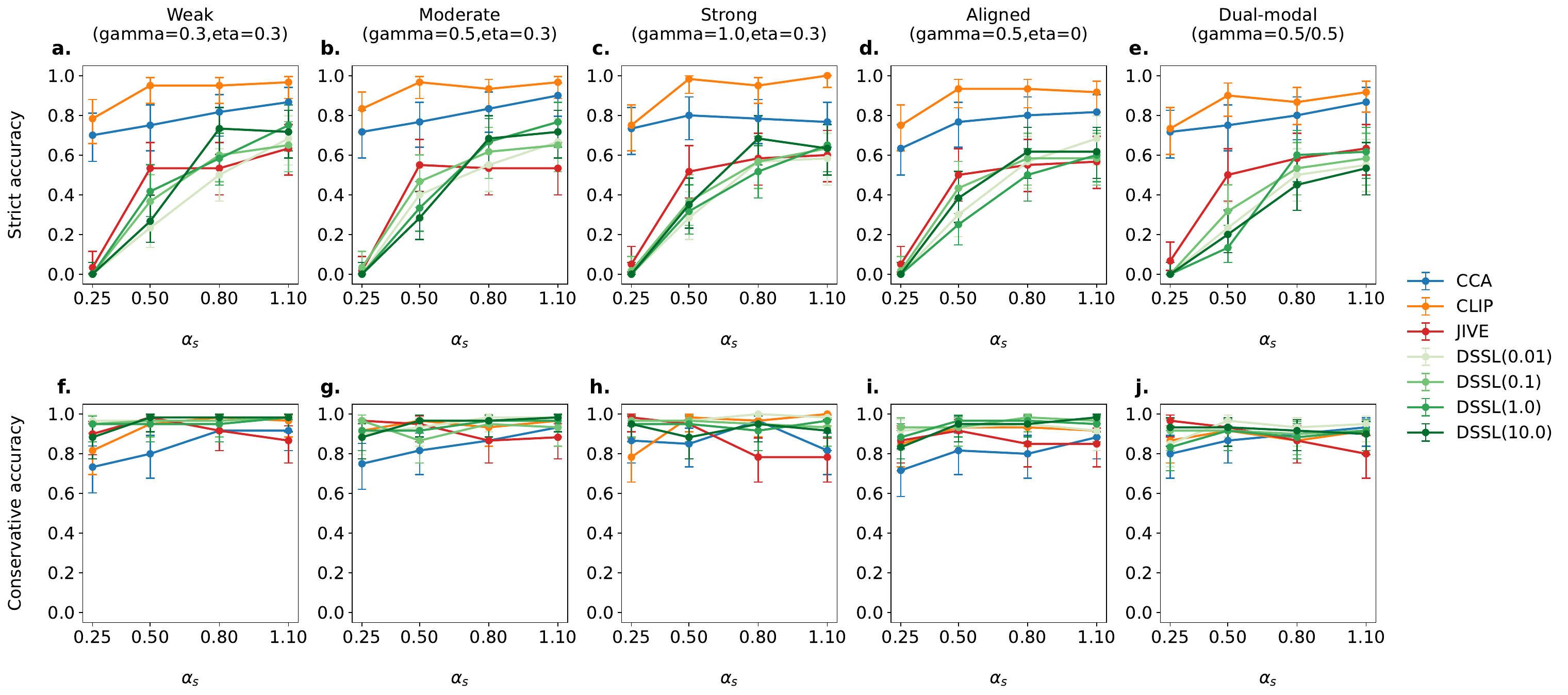}
\caption{\textbf{S1 detection on proxy-contaminated representations, stratified by proxy condition.} Columns: proxy conditions varying in strength ($\gamma$) and alignment ($\eta$). Same format as Figure~\ref{fig:alpha_curves} (top: strict, bottom: conservative).}
\label{fig:app-proxy-s1-condition}
\end{figure}

\begin{figure}[H]
\centering
\includegraphics[width=\linewidth]{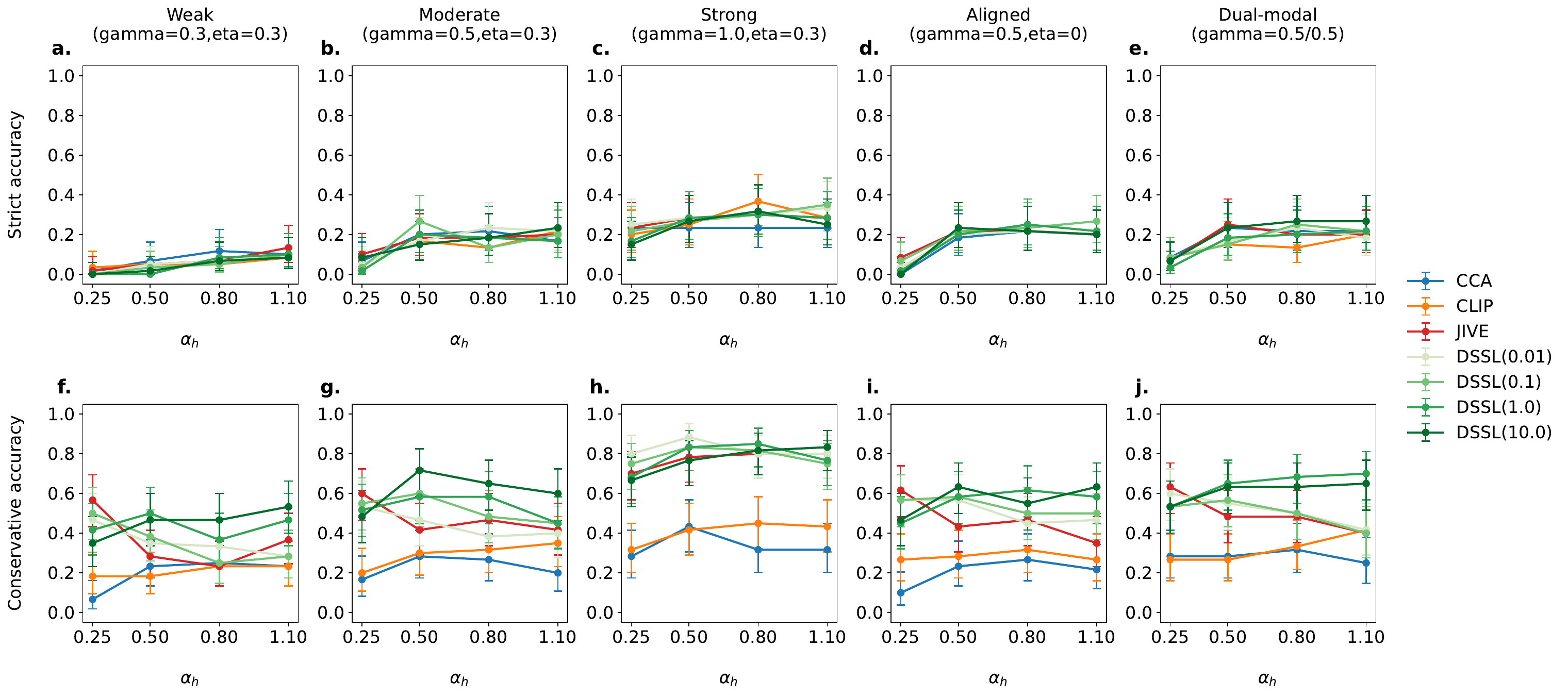}
\caption{\textbf{Proxy S2 detection stratified by proxy condition.} Columns: proxy conditions. Detection rates remain low (5--35\%) across all conditions, with stronger proxy ($\gamma = 1.0$) producing the highest detection rates.}
\label{fig:app-proxy-s2a-condition}
\end{figure}

Figures~\ref{fig:app-proxy-s1-condition} and~\ref{fig:app-proxy-s2a-condition} stratify the proxy results from Figure~\ref{fig:proxy_summary} by individual proxy condition ($\gamma$, $\eta$), providing a detailed view of how proxy strength and alignment geometry affect S1 detection and proxy S2 detection respectively.

\subsection{Cross-Modality Resolution}
\label{app:further-results:cross-modality}

\begin{figure}[H]
\centering
\includegraphics[width=\linewidth]{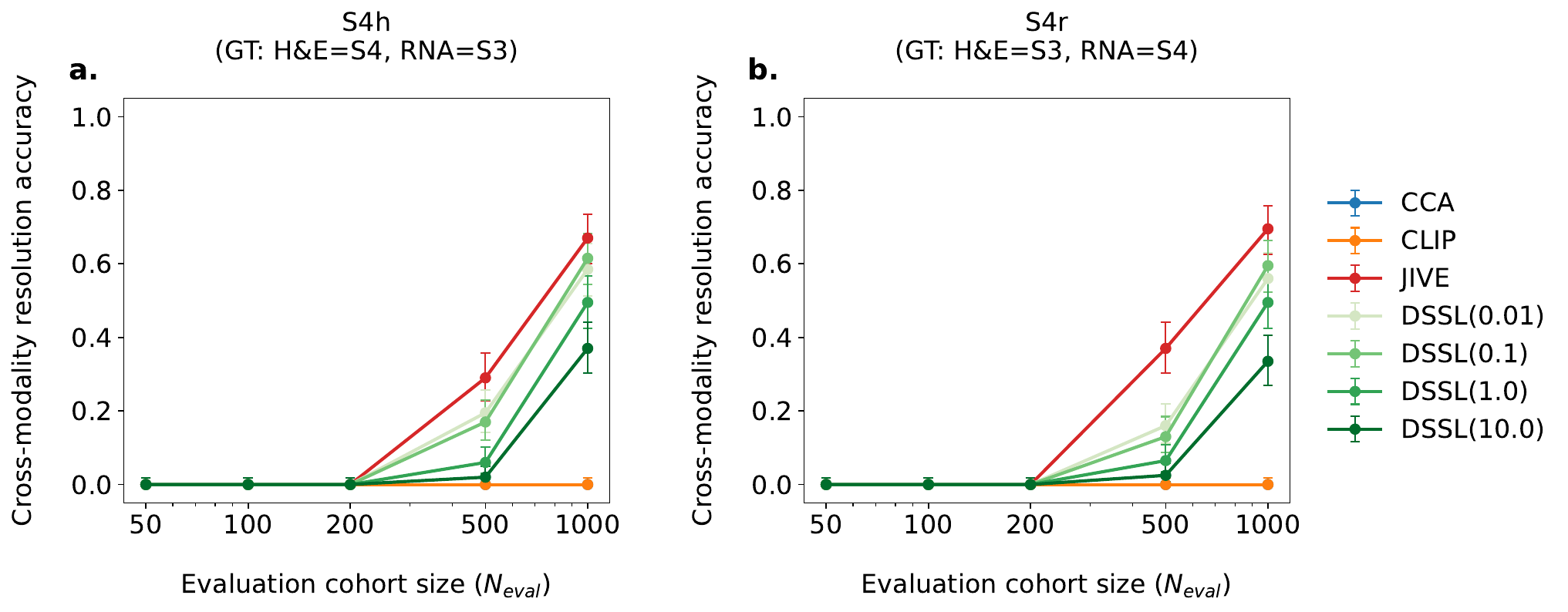}
\caption{\textbf{Cross-modality resolution accuracy versus $N_{\text{eval}}$.} Probability that both modalities are correctly classified simultaneously. (a)~S4h (H\&E=S4, RNA=S3). (b)~S4r (H\&E=S3, RNA=S4). Factorized models achieve meaningful resolution; entangled models cannot resolve different scenarios across modalities.}
\label{fig:app-cross-modality-resolution}
\end{figure}

Figure~\ref{fig:app-cross-modality-resolution} shows cross-modality resolution accuracy, the probability that both modalities are correctly classified simultaneously when they have different ground-truth scenarios (S4h: H\&E=S4, RNA=S3 and vice versa for S4r). These are pure global scenario configurations (not mixed-mechanism), where each modality independently has a well-defined ground truth. Factorized models (JIVE at 60--68\%, DSSL variants at 35--62\%) achieve meaningful cross-modality resolution at $N_{\text{eval}} = 1000$, while entangled models remain near zero because they cannot localize modality-specific signal. Resolution accuracy is still rising at $N_{\text{eval}} = 1000$ for all factorized models.

\clearpage
\subsection{Asymmetric Mixed and Transition Scenario Behavior}
\label{app:further-results:mixed}

\begin{figure}[H]
\centering
\includegraphics[width=\linewidth]{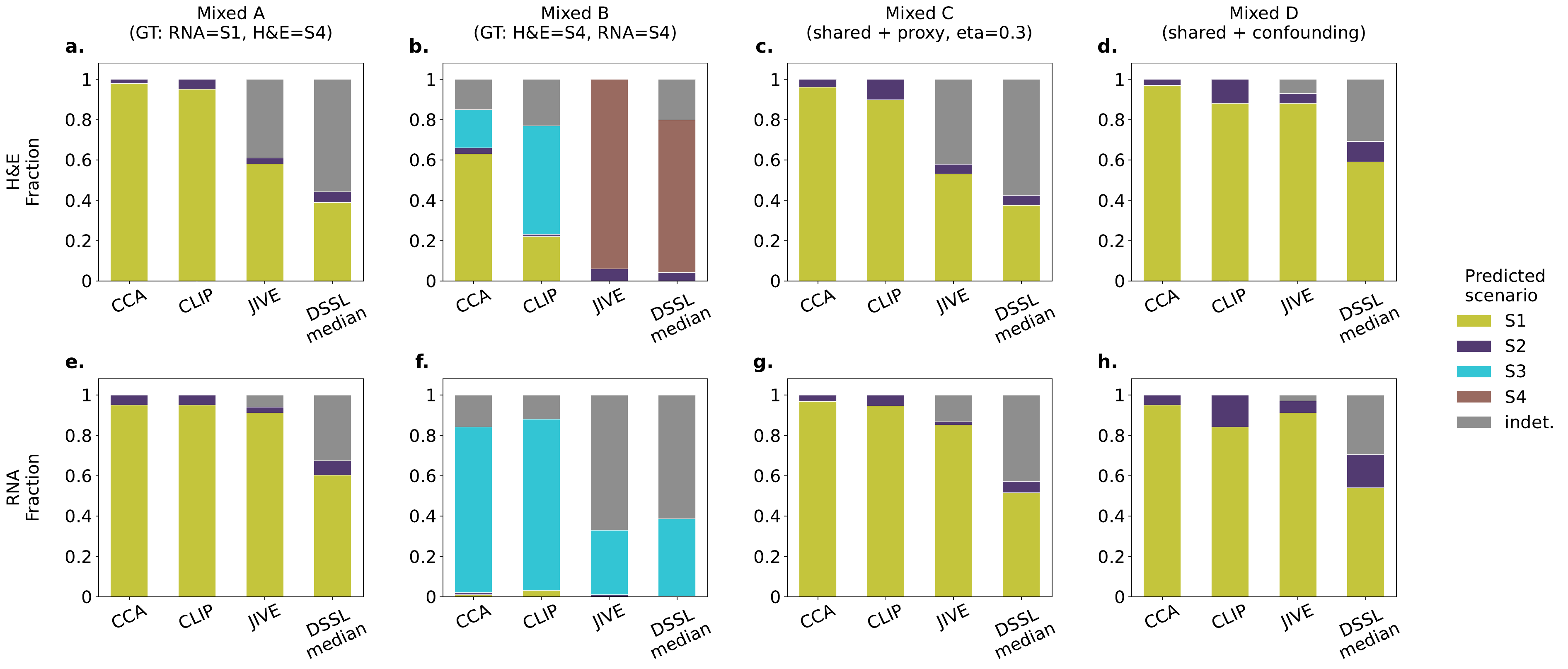}
\caption{\textbf{Per-modality scenario classification on asymmetric mixed-signal tasks.} Baseline regime, $N_{\text{eval}} = 1000$. Top row: H\&E modality. Bottom row: RNA modality. Columns: (a/e)~Mixed~A, (b/f)~Mixed~B, (c/g)~Mixed~C, (d/h)~Mixed~D. All mixed configurations use asymmetric signal strengths ($\alpha_s > \alpha_h$ or $\alpha_s > \alpha_b$). Bars show predicted scenario distribution per model.}
\label{fig:app-mixed-modality}
\end{figure}

When multiple signal sources co-exist in the same task (mixed configurations), DECAT classifies each modality independently. Figure~\ref{fig:app-mixed-modality} shows per-modality scenario classification for mixed-type configurations on the baseline regime at $N_{\text{eval}} = 1000$.

For Mixed~A ($\alpha_s > \alpha_h$, ground truth: RNA=S1, H\&E=S4; Figure~\ref{fig:app-mixed-modality}a/e), RNA is correctly classified as S1 in the majority of cases across all models. However, H\&E is also routed to S1 rather than S4 because the shared signal ($\alpha_s \in \{0.8, 1.1\}$) dominates the modality-specific signal ($\alpha_h \in \{0.3, 0.5\}$) in the learned representation. For Mixed~B ($\alpha_h + \alpha_r$, no shared signal, ground truth: both S4; Figure~\ref{fig:app-mixed-modality}b/f), H\&E is correctly classified as S4 by factorized models, but the RNA-specific signal is too weak for reliable localization, producing mostly S3 or indeterminate classifications for RNA. Mixed~C ($\alpha_s + \alpha_h$ with proxy $\gamma_h > 0$, $\eta = 0.3$, ground truth: RNA=S1, H\&E=S2; Figure~\ref{fig:app-mixed-modality}c/g) tests whether DECAT can distinguish shared biology from proxy-driven signal across modalities. Mixed~D ($\alpha_s + \alpha_b$, ground truth: both shared biology and confounding co-exist; Figure~\ref{fig:app-mixed-modality}d/h) tests behavior when a prediction is partially real and partially confounded. In this case DECAT should ideally return S2 (since confounding is present) or indeterminate, but not S1. Strict mixed-scenario resolution accuracy (both modalities simultaneously correct) is near zero for Mixed~A and~B due to signal asymmetries.

The inability to resolve Mixed~A as S1/S4 reflects the asymmetric signal design ($\alpha_s \gg \alpha_h$), in which shared signal dominates the H\&E representation and the modality-specific contribution is too weak for $\Delta_{\text{shared}}$ to localize. In deployment, distinguishing S1 from S4 per modality matters because it determines whether a single-modality prediction is independently supported by cross-modal biology or relies on modality-specific features.

\begin{figure}[H]
\centering
\includegraphics[width=\linewidth]{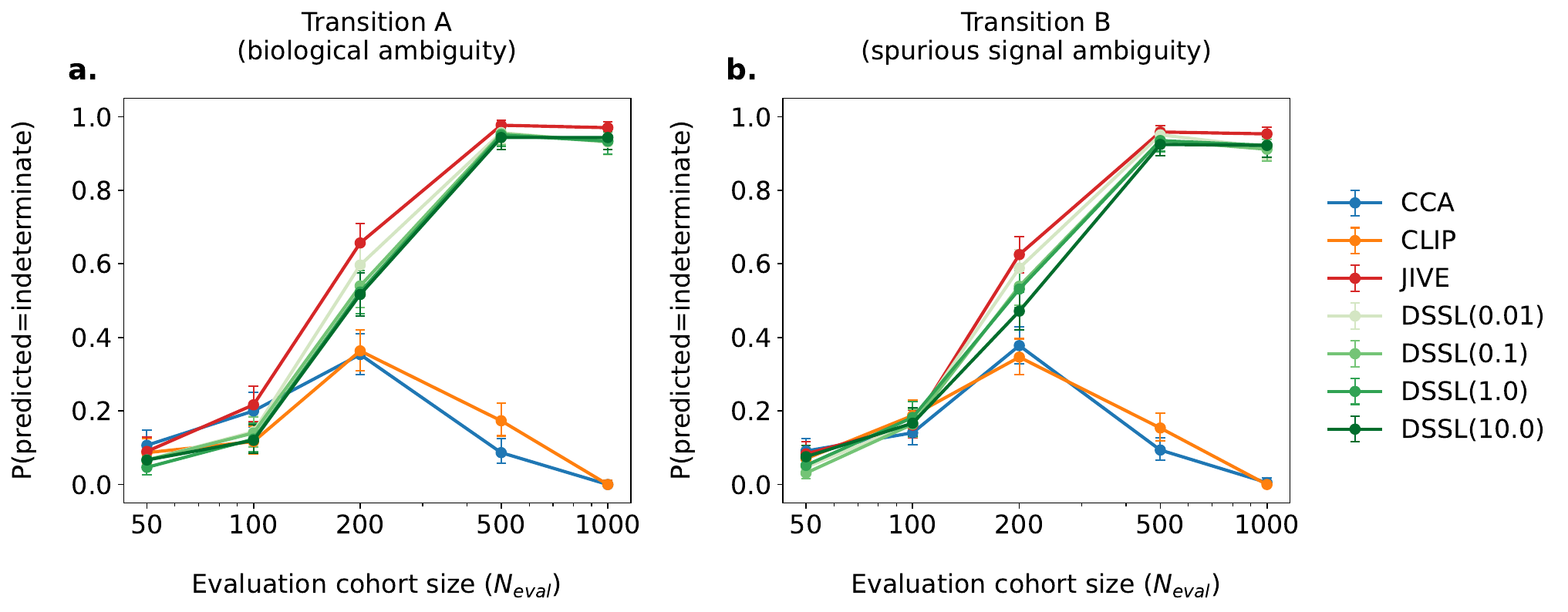}
\caption{\textbf{Indeterminate sensitivity on transition configurations.} (a)~Transition~A (biological ambiguity, no proxy or confounding). (b)~Transition~B (confounded ambiguity, with proxy and confounding overlaid). Factorized models increasingly return indeterminate with more data (correct behavior on non-identifiable tasks) while entangled models become more overconfident.}
\label{fig:app-transition}
\end{figure}

Figure~\ref{fig:app-transition} shows indeterminate sensitivity on transition configurations, which are designed to be fundamentally non-identifiable ($\alpha_s \approx \alpha_h \approx \alpha_r > 0$, with no dominant signal source). Factorized models (JIVE, DSSL) correctly return indeterminate in 90--95\% of cases at $N_{\text{eval}} = 1000$, with sensitivity increasing as sample size grows. Larger samples make $\Delta_{\text{shared}}$ decisive enough to reveal that signal does not cleanly localize to either component, correctly routing to indeterminate.
Entangled models (CCA, CLIP) show a non-monotonic pattern: indeterminate sensitivity rises briefly at moderate $N_{\text{eval}}$ (200--500) then falls to 0--10\% at $N_{\text{eval}} = 1000$.
At small $N_{\text{eval}}$, insufficient power means the decision tree cannot classify at all, producing indeterminate by default.
At moderate $N_{\text{eval}}$, some metrics are significant while others remain ambiguous, generating genuine indeterminate hedging.
At large $N_{\text{eval}}$, entangled models have sufficient power for the signal gate, $P_{\text{transfer}}$, and $D_{\text{quantile}}^{\text{task}}$ to reach significance, and the non-factorized pathway uses $A_{\text{norm}}$ for the S1/S4 distinction, committing to a specific scenario even when the ground truth is ambiguous.
This is the same mechanism that drives increasing FSCR with sample size for entangled models (Figure~\ref{fig:fscr}): more data provides more power, but without factorization that power produces overconfidence rather than appropriate uncertainty.

\subsection{Sensitivity Analyses}
\label{app:further-results:sensitivity}

This section consolidates the sensitivity analyses of the decision procedure and the simulator assumptions into a single summary. Sensitivity to measurement regime, evaluation cohort size, and proxy contamination is reported in Appendices~\ref{app:further-results:power}, \ref{app:further-results:neval}, and~\ref{app:further-results:proxy}, and the minimum cohort size for reliable S2 detection on TCGA in Appendix~\ref{app:further-tcga-results:power}. Here we report an ablation of the decision procedure, nonlinear observation processes, and cohort shift alignment. All analyses use the baseline regime, $N_{\text{eval}} = 1000$, 50 seeds $\times$ 7 models, pooled across all signal strengths $\alpha \in \{0.25, 0.50, 0.80, 1.10\}$. The S3 null calibration at 94--96\% across all conditions (ablation variants, nonlinear conditions) demonstrates that the permutation tests are consistently calibrated regardless of operating conditions.

\paragraph{Decision procedure ablation.}
We disable each metric in the decision tree and re-classify all evaluations, reporting strict accuracy (correct scenario assigned). We remove one metric at a time by overriding its value before classification while leaving all other metrics unchanged. $-D_{\text{quantile}}^{\text{task}}$ sets $D_{\text{quantile}}^{\text{task}}$ inside its null, preventing S2 detection. $-\Delta_{\text{shared}}$ treats every model as entangled, removing the localization stage. $-A_{\text{norm}}$ sets $A_{\text{norm}}$ below its null, removing the shared-structure confirmation required for S1. $-P_{\text{transfer}}$ sets the lower confidence bound of $P_{\text{transfer}}$ below its null, so no signal is treated as transferring.

Each metric prevents a specific failure mode (Table~\ref{tab:app-ablation}).
$D_{\text{quantile}}^{\text{task}}$ is necessary for S2 (removing it eliminates 100\% of spurious signal detection). For entangled models, removing it increases FSCR from 16\% to 29\% (S2 cases misrouted to S1).
$\Delta_{\text{shared}}$ is the sole S4 discriminator (for factorized models, removing it reduces S4 accuracy from 31\% to 0\%, as modality-specific signal is misassigned to S1 and FSCR rises from 0\% to 25\%, replicating the entangled-model failure mode in Section~\ref{sec:synthetic-results}). It has no effect on entangled models, which do not use localization.
$P_{\text{transfer}}$ is required for any positive scenario assignment.
Finally, $A_{\text{norm}}$ is required for S1 assignment, preventing shared-signal claims when cross-modal structure is absent.
S3 calibration (95--96\%) is unaffected by any single ablation, confirming the permutation nulls are robust.
No single metric is redundant, and the ordering prioritizes conservative assignment (S2 checked before S1) to minimize false shared claims.

\begin{table}[h]
\centering
\small
\caption{\textbf{Decision procedure ablation.} Strict accuracy per ground-truth scenario and FSCR when one metric is disabled. Bold: scenarios affected by the ablation. Entangled: mean of CCA and CLIP. Factorized: mean of JIVE and DSSL (averaged over its four configurations). Baseline regime, $N_{\text{eval}} = 1000$, 50 seeds, pooled across signal strengths.}
\label{tab:app-ablation}
\begin{tabular}{lllccccc}
\toprule
& \textbf{Condition} & \textbf{Effect} & \textbf{S1} & \textbf{S2B} & \textbf{S3} & \textbf{S4} & \textbf{FSCR} \\
\midrule
\multirow{5}{*}{\rotatebox{90}{\textbf{Entangled}}}
& Full                                 & ---                              & 0.961 & 0.461 & 0.958 & 0.000 & 0.163 \\
& $-D_{\text{quantile}}^{\text{task}}$ & S2 cannot be assigned            & 1.000 & \textbf{0.000} & 0.958 & 0.000 & 0.293 \\
& $-\Delta_{\text{shared}}$            & No effect (not used)             & 0.961 & 0.461 & 0.958 & 0.000 & 0.163 \\
& $-A_{\text{norm}}$                   & S1 cannot be assigned            & \textbf{0.000} & 0.461 & 0.958 & 0.173 & 0.000 \\
& $-P_{\text{transfer}}$               & S1, S2, S4 cannot be assigned    & \textbf{0.000} & \textbf{0.000} & 0.958 & \textbf{0.000} & 0.000 \\
\midrule
\multirow{5}{*}{\rotatebox{90}{\textbf{Factorized}}}
& Full                                 & ---                              & 0.600 & 0.443 & 0.951 & 0.314 & 0.000 \\
& $-D_{\text{quantile}}^{\text{task}}$ & S2 cannot be assigned            & 0.629 & \textbf{0.000} & 0.951 & 0.327 & 0.000 \\
& $-\Delta_{\text{shared}}$            & S4 cannot be assigned; S1 relaxed & \textbf{0.953} & 0.443 & 0.951 & \textbf{0.000} & 0.251 \\
& $-A_{\text{norm}}$                   & S1 cannot be assigned            & \textbf{0.000} & 0.443 & 0.951 & 0.314 & 0.000 \\
& $-P_{\text{transfer}}$               & S1, S2, S4 cannot be assigned    & \textbf{0.000} & \textbf{0.000} & 0.951 & \textbf{0.000} & 0.000 \\
\bottomrule
\end{tabular}
\end{table}

\paragraph{Nonlinear observation processes.}
DECAT's metrics operate on learned representations, which are already nonlinear transforms of the observations, and the simulator's linearity applies only to the observation-generating process (Appendix~\ref{app:simulator:measurement}). To directly test robustness to nonlinear observation processes, we apply fixed random 2-layer MLPs (ReLU, variance-normalized to the linear baseline, sampled once per run and per modality) to the data-generating process under three conditions: nonlinear observations (the entire measurement model is nonlinear), nonlinear confounding (only the batch term enters nonlinearly), and both nonlinear (biology and batch pass through separate MLPs).

Under fully nonlinear conditions (the most realistic scenario), family-mean S2B detection is preserved at baseline levels (0.48 vs.\ 0.46 entangled; 0.45 vs.\ 0.44 factorized), although individual models converge (CCA and JIVE increase from 0.28--0.32 to 0.50--0.52; CLIP and DSSL decrease from 0.61 to 0.40--0.45), and S3 calibration remains at 94--96\% (Table~\ref{tab:app-nonlinear}). FSCR is unchanged for factorized models ($\leq 0.1\%$) and rises from 16\% to 22\% for entangled models.
The cost is reduced localization for factorized models (S4: 0.31 to 0.13), consistent with the higher indeterminate rates observed for DSSL on real TCGA data.
Under isolated nonlinear confounding (unlikely in practice), S2B detection degrades (0.15 vs.\ 0.46 entangled; 0.15 vs.\ 0.44 factorized) but errors route to indeterminate rather than false S1 claims.

\begin{table}[h]
\centering
\small
\caption{\textbf{Nonlinear observation processes.} Strict accuracy per ground-truth scenario and FSCR. MLP architectures: input dimension $d_h$, hidden width 512, output dimension $d_h$ (nonlinear observations); input dimension $k$, hidden width 128, output dimension $d_h$ (nonlinear confounding); the both-nonlinear condition uses both. CCA and CLIP are entangled; JIVE and DSSL are factorized. DSSL is averaged over its four configurations. Family means reported in the text are the means of the two models in each family. Entangled models do not assign S4. Baseline regime, $N_{\text{eval}} = 1000$, 50 seeds, pooled across signal strengths.}
\label{tab:app-nonlinear}
\begin{tabular}{llccccc}
\toprule
\textbf{Model} & \textbf{Condition} & \textbf{S1} & \textbf{S2B} & \textbf{S3} & \textbf{S4} & \textbf{FSCR} \\
\midrule
\multirow{4}{*}{CCA}
& Linear (baseline)       & 0.965 & 0.318 & 0.945 & --- & 0.224 \\
& Nonlinear observations  & 0.970 & 0.510 & 0.925 & --- & 0.254 \\
& Nonlinear confounding   & 0.973 & 0.110 & 0.945 & --- & 0.190 \\
& Both nonlinear          & 0.968 & 0.515 & 0.935 & --- & 0.269 \\
\cmidrule(lr){1-7}
\multirow{4}{*}{CLIP}
& Linear (baseline)       & 0.958 & 0.605 & 0.970 & --- & 0.102 \\
& Nonlinear observations  & 0.983 & 0.495 & 0.955 & --- & 0.194 \\
& Nonlinear confounding   & 0.975 & 0.185 & 0.955 & --- & 0.144 \\
& Both nonlinear          & 0.968 & 0.453 & 0.935 & --- & 0.176 \\
\midrule
\multirow{4}{*}{JIVE}
& Linear (baseline)       & 0.703 & 0.278 & 0.945 & 0.359 & 0.000 \\
& Nonlinear observations  & 0.375 & 0.505 & 0.960 & 0.159 & 0.000 \\
& Nonlinear confounding   & 0.725 & 0.105 & 0.970 & 0.360 & 0.000 \\
& Both nonlinear          & 0.425 & 0.498 & 0.930 & 0.194 & 0.000 \\
\cmidrule(lr){1-7}
\multirow{4}{*}{DSSL}
& Linear (baseline)       & 0.498 & 0.608 & 0.956 & 0.269 & 0.000 \\
& Nonlinear observations  & 0.134 & 0.458 & 0.955 & 0.053 & 0.002 \\
& Nonlinear confounding   & 0.616 & 0.193 & 0.943 & 0.205 & 0.000 \\
& Both nonlinear          & 0.178 & 0.403 & 0.941 & 0.071 & 0.002 \\
\bottomrule
\end{tabular}
\end{table}

\paragraph{Cohort shift alignment.}
$D_{\text{quantile}}^{\text{task}}$ detects instability along the outcome-predictive direction. As noted in Appendix~\ref{app:simulator-limitations}, if a confounder is orthogonal to the cohort shift it does not constitute a generalization risk for cohorts shifted in that direction. In practice, this means that a prediction that appears safe for one deployment cohort or context is not necessarily safe for all deployment contexts.
To demonstrate the effect of cohort shift alignment directly, we sweep $\rho_b$ (the alignment between the cohort shift direction and the confounding direction; Eq.~\ref{eq:app-batchalign}) from 1.0 (fully aligned, main experiment) through 0.5 (partial) to 0.0 (random shift) for the S1 and S2B configurations, holding the shift magnitude $M$ fixed.

At $\rho_b = 0$ (random shift), S2B detection drops to 4--5\% for both model families (Table~\ref{tab:app-rhob}). For entangled models, the S2B to S1 misclassification rate rises from 22\% to 72\%: these confounded cases satisfy S1 criteria because $D_{\text{quantile}}^{\text{task}}$ cannot detect instability when the cohort shift is orthogonal to the confounding direction. Factorized models never misclassify S2B as S1 at any $\rho_b$; 86\% of their S2B cases route to indeterminate.
S1 accuracy is unaffected for both families (entangled 0.96--0.98; factorized 0.59--0.60), confirming batch shift direction is irrelevant when no confounding is present.
This is consistent with Figure~\ref{fig:app-dq-trajectories}a, which validates on ground-truth latents that with lower values of $\rho_b$, S2B trajectories collapse into the S1/S4 region of the $P_{\text{transfer}}$ vs.\ $D_{\text{quantile}}^{\text{task}}$ metric space.

\begin{table}[h]
\centering
\small
\caption{\textbf{Cohort shift alignment.} S1 and S2B strict accuracy, and the fraction of S2B cases classified as S1 or indeterminate, as the cohort shift rotates away from the confounding direction. $\rho_b = 1$ is the main experiment. Entangled: mean of CCA and CLIP. Factorized: mean of JIVE and DSSL (averaged over its four configurations). Baseline regime, $N_{\text{eval}} = 1000$, 50 seeds, pooled across signal strengths.}
\label{tab:app-rhob}
\begin{tabular}{llcccc}
\toprule
\textbf{Model} & $\rho_b$ & \textbf{S1 acc} & \textbf{S2B acc} & \textbf{S2B to S1} & \textbf{S2B to indeterminate} \\
\midrule
\multirow{3}{*}{Entangled}
& 1.0 (aligned) & 0.961 & 0.461 & 22\% & 19\% \\
& 0.5 (partial) & 0.965 & 0.298 & 44\% & 13\% \\
& 0.0 (random)  & 0.978 & 0.041 & 72\% & 10\% \\
\midrule
\multirow{3}{*}{Factorized}
& 1.0 (aligned) & 0.600 & 0.443 & 0\%  & 46\% \\
& 0.5 (partial) & 0.592 & 0.279 & 0\%  & 62\% \\
& 0.0 (random)  & 0.598 & 0.047 & 0\%  & 86\% \\
\bottomrule
\end{tabular}
\end{table}

\section{Further TCGA Experimental Results}
\label{app:further-tcga-results}

\subsection{TCGA Detection Rate Curves and CCA Variate Extreme-C}
\label{app:further-tcga-results:detection}
\label{app:further-tcga-results:cca-extreme}

\begin{figure}[H]
\centering
\includegraphics[width=\linewidth]{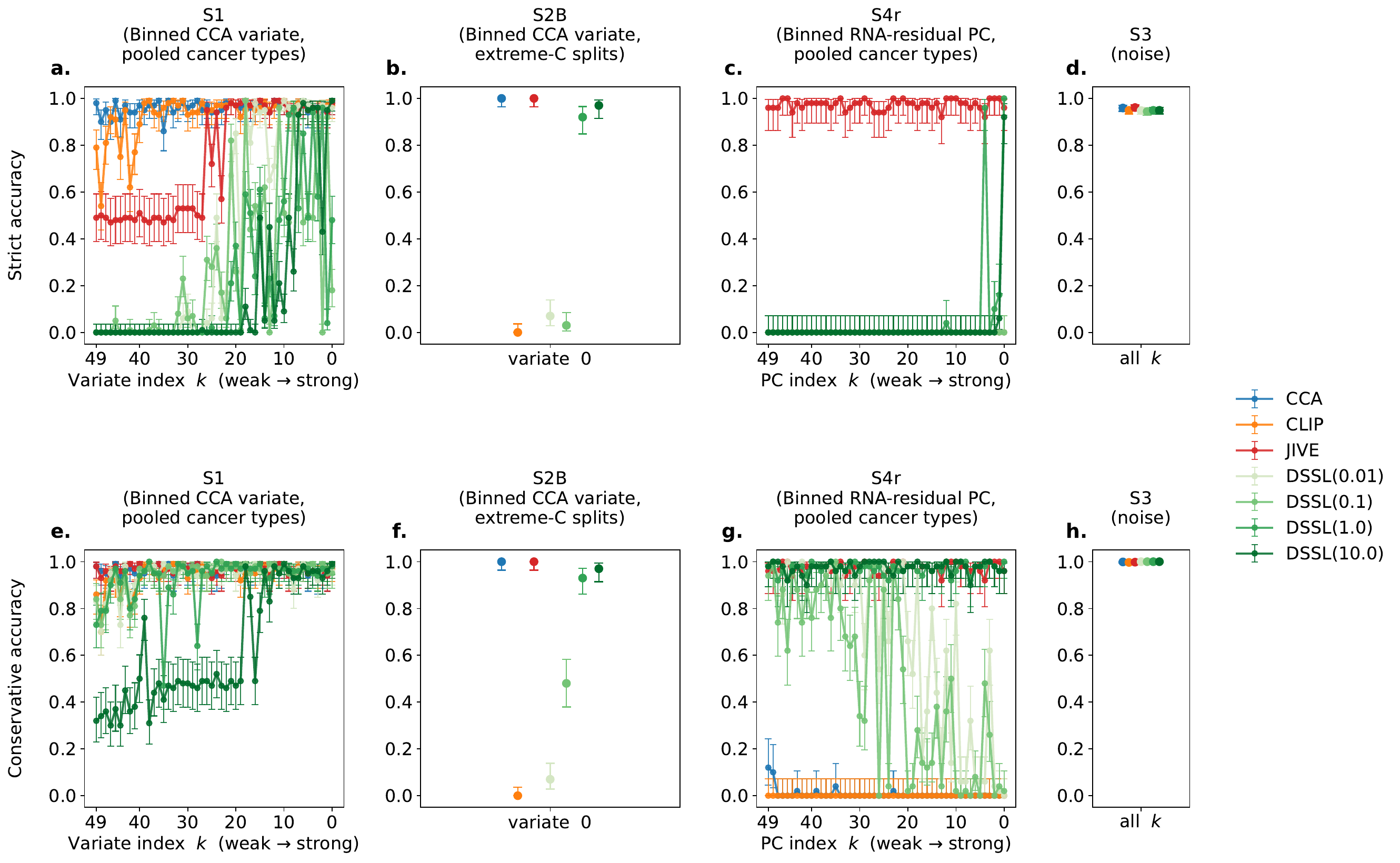}
\caption{\textbf{TCGA detection rates by variate index}. Top row: strict accuracy (correct scenario assigned). Bottom row: conservative accuracy (correct or indeterminate). All continuous labels are binarized at the pan-cancer median. 95\% Clopper--Pearson confidence intervals.
(a/e)~S1: CCA canonical variates ($k{=}0$--$49$), pooled random splits. Variate~0 carries the strongest shared signal (maximal cross-modal correlation) and performance degrades for weaker variates.
(b/f)~S2B: CCA variate~0 only, extreme-C splits. Only variate~0 is shown because the extreme-C cohort was constructed by enriching C for cancer types with extreme variate-0 scores. Variates 1--49 are orthogonal to this axis by construction (CCA orthogonality constraint), so the cohort shift is invisible to their probe directions.
(c/g)~S4r: JIVE RNA-residual PCs, pooled random splits, RNA modality only. JIVE RNA-residual PCs are orthogonal to the shared CCA subspace by construction. Entangled models return 0\% S4 at all indices.
(d/h)~S3: random binary labels, pooled random splits. All models return S3 at 94--97\% strict accuracy.}
\label{fig:tcga_detection_curves}
\end{figure}

Figure~\ref{fig:tcga_detection_curves} shows detection rates across all designed label types on TCGA. CCA variate~0 (the direction of maximal cross-modal correlation) serves as a positive control for S1 detection under pooled evaluation, and is classified as S1 at ${\geq}$97\% (Figure~\ref{fig:tcga_detection_curves}a). Under extreme-C splits, it is reclassified as S2 at ${\sim}$100\% by CCA and JIVE but at 0\% by CLIP (Figure~\ref{fig:tcga_detection_curves}b). For the CCA variate extreme-C design, the extreme pool $E$ consists of patients above the 75th percentile of CCA variate~0 score. Because CCA variate~0 is strongly correlated with cancer type (high $\eta^2$), this enrichment implicitly creates a cancer-type-biased Cohort~C without requiring cancer-type labels. The 0\% strict accuracy for S2 by CLIP may be because CLIP's contrastive training objective distributes confound-correlated variance across a high-dimensional nonlinear manifold, making the cohort shift undetectable along the linear probe direction used by $D_{\text{quantile}}^{\text{task}}$. A nonlinear probe extension may address this (Section~\ref{sec:discussion}). DSSL at moderate-to-strong disentanglement ($\beta{\geq}1.0$) detected S2 at 92--97\%, while weak disentanglement ($\beta{\leq}0.1$) detected S2 near 0\%, mirroring entangled behavior.

\subsection{TCGA False Shared Claim Rate}
\label{app:further-tcga-results:fscr}

\begin{figure}[H]
\centering
\includegraphics[width=\linewidth]{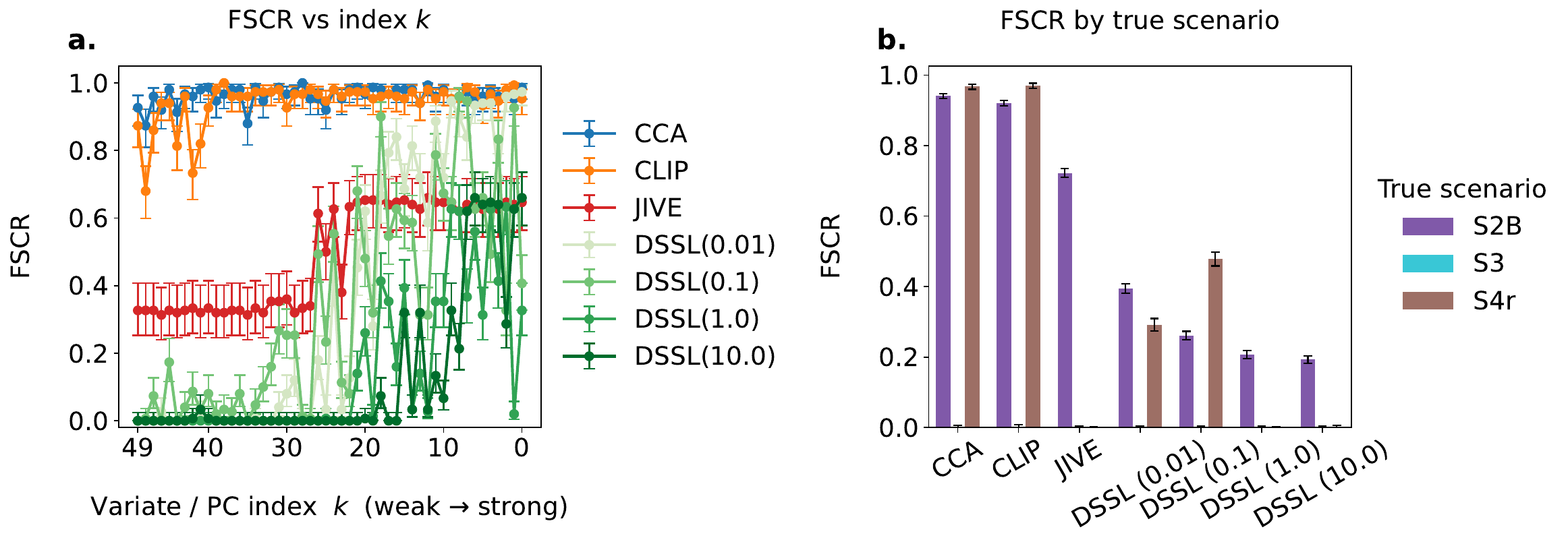}
\caption{\textbf{False shared claim rate (FSCR) on TCGA pooled random splits.} (a)~FSCR by variate index for S4r tasks (RNA-modality residual PCs). (b)~FSCR decomposed by ground-truth scenario (S2B, S3, S4r). Entangled models (CCA, CLIP) systematically misclassify non-shared signal as shared. Factorized models (JIVE, DSSL at $\beta{\geq}1.0$) eliminate false S4r claims.}
\label{fig:tcga_fscr}
\end{figure}

Figure~\ref{fig:tcga_fscr} decomposes FSCR on TCGA pooled random splits by ground-truth scenario and model. Ground-truth S4r labels are JIVE RNA-residual PCs (modality-specific by construction, regardless of evaluation design). Ground-truth S2B labels are CCA variates, which are confounded with cancer type but the confounding is only detectable under extreme-C evaluation. Under the pooled random evaluation shown here, $D_{\text{quantile}}^{\text{task}}$ is rarely significant because cohort compositions are similar (no systematic enrichment of label-extreme patients in C), so most S2B tasks are classified as S1, counting as FSCR even though the framework's behavior is expected given the evaluation design.

JIVE and DSSL at strong disentanglement ($\beta{\geq}1.0$) exhibited zero or near-zero FSCR on S4r tasks, while DSSL at weak disentanglement ($\beta{\leq}0.1$) showed nonzero S4r FSCR because insufficient bottleneck pressure allows modality-specific signal to leak into the shared latent. JIVE showed substantial FSCR on S2B tasks (72\%), where its shared component (identical to CCA by construction) captures cancer-type-driven cross-modal correlation (different cancer types have correlated morphological and transcriptomic profiles, creating cross-modal agreement that reflects tissue-of-origin rather than within-type biology). Because JIVE's shared component is the CCA projection, claiming S1 for CCA variate labels under pooled random evaluation is architecturally expected rather than a genuine false claim. DSSL at weak disentanglement ($\beta{\leq}0.1$) also exhibited substantial FSCR (29--32\%), driven primarily by S2B false claims, while strong disentanglement ($\beta{\geq}1.0$) reduced FSCR to 12\%. This exceeds the near-zero observed synthetic FSCR for factorized models and likely reflects the correlation between cancer-type confounding and genuine shared biology in TCGA. The synthetic simulator enforces orthogonality between shared signal and batch ($z_s \perp b$), whereas in real data the information bottleneck at weak $\beta$ is insufficient to exclude confound-correlated variance from the shared latent.
In summary, the S4r component of FSCR represents unambiguous misclassification (modality-specific signal wrongly called shared), while the S2B component reflects the fundamental limitation that confounding cannot be detected without a composition shift in the evaluation cohorts. The extreme-C design (Section~\ref{sec:tcga}, Figure~\ref{fig:tcga_alpha_mixture_curve}) provides the composition shift needed to convert S2B FSCR into correct S2 detection.

\subsection{Power Curve for S2 Detection}
\label{app:further-tcga-results:power}

\begin{figure}[H]
\centering
\includegraphics[width=\linewidth]{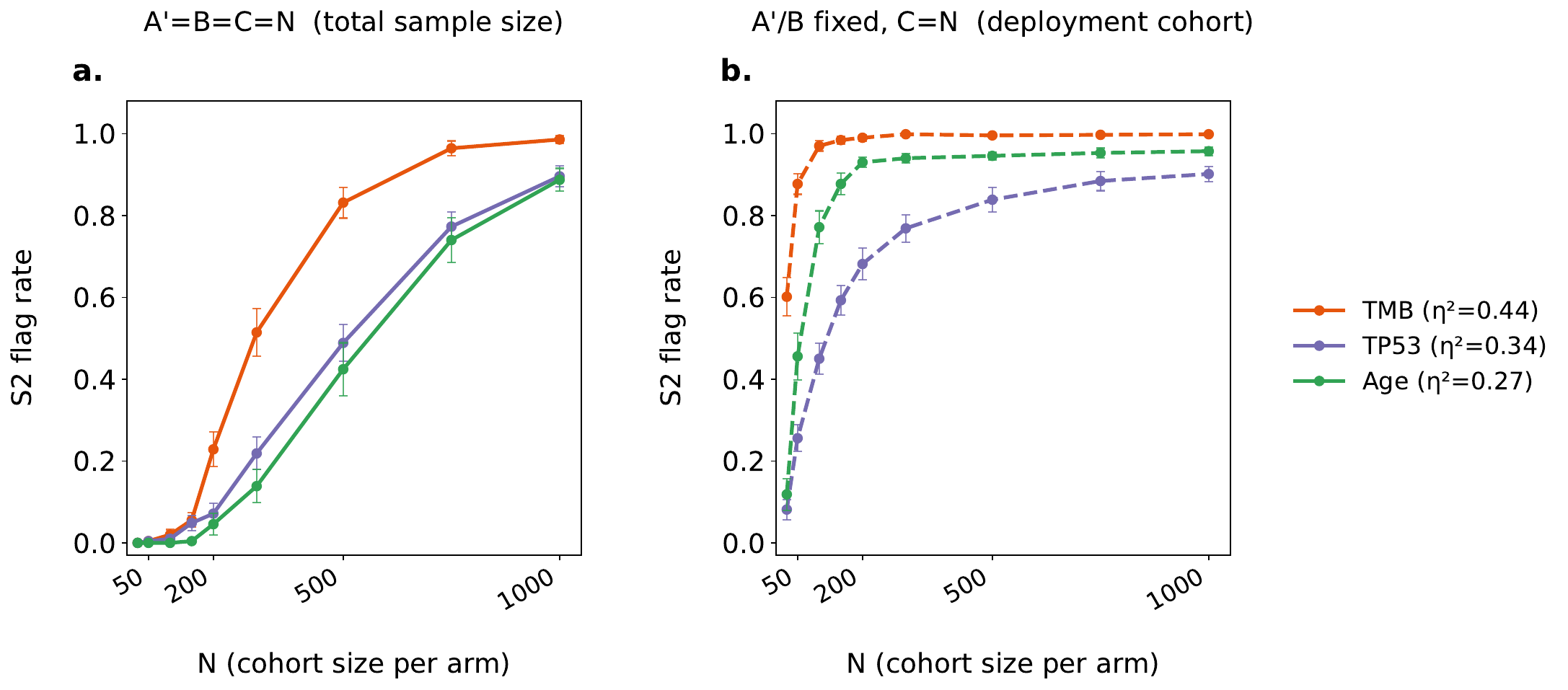}
\caption{\textbf{Power curve for S2 detection (C drawn entirely from extreme pool E, $\alpha{=}1$).} S2 detection rate versus evaluation cohort size $N$ under two designs: (a)~vary-C (A$'$/B held at full size, only $N_C$ varied) and (b)~equal-N (all cohorts varied together, A$'{=}$B${=}$C${=}N$). The gap between panels shows the value of large reference cohorts for probe quality. For example, under vary-C, TMB reaches 80\% detection at $N_C{\approx}50$, whereas under equal-N the same detection requires $N{\approx}500$. E is the top 25\% of patients by label value for TMB and Age (${\sim}4 \times N_C$ evaluation patients needed), or the cancer types with TP53 prevalence above 65\% or below 15\% for TP53 (${\sim}3 \times N_C$).}
\label{fig:tcga_power_curve}
\end{figure}

Figure~\ref{fig:tcga_power_curve} measures the minimum cohort size required for reliable S2 detection under the extreme-C design (C drawn entirely from the label-extreme pool E, i.e.\ $\alpha{=}1$), pooled across all seven models, with two cohort-sizing strategies. For TMB and Age, the extreme pool E is defined as patients above the 75th percentile of the label value (top 25\%), requiring ${\sim}4 \times N_C$ evaluation patients to populate C. For TP53, E is defined as cancer types with TP53 mutation prevalence above 65\% or below 15\% (18 of 32 types; 36\% of patients), because TP53 status is binary and a patient-level quantile is not meaningful. This yields a ${\sim}3 \times N_C$ multiplier. The vary-C design (Figure~\ref{fig:tcga_power_curve}a) holds A$'$/B at full size (${\sim}$1{,}465 patients each) and varies only $N_C$, isolating the deployment cohort size requirement. The equal-N design (Figure~\ref{fig:tcga_power_curve}b) varies all cohorts together (A$'{=}$B${=}$C${=}N$), revealing the additional bottleneck from probe quality when reference cohorts are small.

\subsection{Confounding Within a Single Cancer Type}
\label{app:further-tcga-results:within-brca}

To test confounding mechanisms beyond cancer-type composition, we applied DECAT within BRCA patients only (N=1{,}038, eliminating cancer-type variation entirely) on 4 labels (TP53, CDH1, ER, PR) across 5 FMs, using candidate confounders identified by Dawood et al.~\cite{dawoodConfoundingFactorsBiases2026} through stratification analysis. These labels span prevalences from 15\% (CDH1 mutation) to 77\% (ER positivity). The candidate confounders are never provided to DECAT. Cohorts follow the $\alpha$-mixture design of Section~\ref{sec:tcga}: B is a random split of the evaluation pool, and C draws a fraction $\alpha$ from the extreme pool E (top 25\% of the C pool by FM probe score) and the remainder from outside E. $\alpha = 0$ therefore excludes E entirely, and the natural composition corresponds to $\alpha \approx 0.25$. The sweep does not introduce confounding: C contains only real patients from the evaluation pool with unchanged labels and embeddings, and $\alpha$ changes only which patients are included. It therefore tests whether the probe's predictions remain stable under a shift in deployment composition along the probe's own predictive direction. Because C is shifted along this direction, $D_{\text{quantile}}^{\text{task}}$ confirms the shift, and the S2 verdict depends on whether the predictive ranking still replicates in the shifted cohort ($P_{\text{transfer}}$). Each $\alpha \in \{0.0, 0.2, \ldots, 1.0\}$ is evaluated over 50 splits.

DECAT flags composition-dependent instability for TP53 (S2 in up to 68\% of splits), CDH1 (up to 56\%), and ER (up to 54\%) without access to the candidate confounders (Table~\ref{tab:app-within-brca}), testing fundamentally different confounding mechanisms (molecular correlates, histological identity) than the pan-cancer analysis. Post-hoc stratification by the candidate confounder confirms AUROC drops of 0.10--0.40 in the affected subgroup for TP53, CDH1, and PR across all 5 FMs; for ER, the CDH1+ stratum contains too few ER$-$ patients for a stable estimate. S2 rates vary across FMs because of $P_{\text{transfer}}$: $D_{\text{quantile}}^{\text{task}}$ is significant in 78--100\% of splits for every FM, whereas for TITAN, UNI, and (except for ER) OpenMidnight the A$'$- and B-trained probes rarely agree on the patient ranking in C, and the verdict is indeterminate (insufficient replication; Appendix~\ref{app:further-tcga-results:indeterminate}). Within a single cancer type, S2 rates track probe strength (peak S2 vs.\ cross-validated AUROC on A$'$, Spearman $\rho = 0.67$), so differences across labels and FMs partly reflect the power of $P_{\text{transfer}}$ rather than confounder strength alone. DECAT does not detect a specific known confounder but rather flags composition-dependent instability along the predictive direction, regardless of upstream source.

\begin{table}[h]
\centering
\small
\caption{\textbf{Within-BRCA confounding.}
DECAT columns: percentage of 50 splits in which $D_{\text{quantile}}^{\text{task}}$ and
$P_{\text{transfer}}$ are significant (exceed their permutation nulls) and in which the verdict is S2 or indeterminate, measured at each
FM's peak-S2 $\alpha$ ($\alpha{=}0$ excludes the extreme pool from C;
natural composition is $\alpha{\approx}0.25$; for FMs with 0\% peak S2, $\alpha{=}0.6$ is used).
S2 requires both metrics to be significant; when only $D_{\text{quantile}}^{\text{task}}$ is significant, the verdict is indeterminate ($\leq$8\% transferable in
all rows).
Post-hoc columns: AUROC is the overall probe AUROC within BRCA; worst stratum is the AUROC
within the candidate-confounder stratum with the largest drop (stratum listed under the
confounder), and $\Delta$ the drop from overall.
Candidate confounders from Dawood et al.~\cite{dawoodConfoundingFactorsBiases2026} are used
only for post-hoc stratification.
$^\dagger$Worst stratum and $\Delta$ computable in 8 of 50 splits ($\approx$5 ER$-$ patients
in CDH1+).}
\label{tab:app-within-brca}
\setlength{\tabcolsep}{3pt}
\begin{tabular}{lllccccccccc}
\toprule
& & & & \multicolumn{4}{c}{\textbf{DECAT (\%)}} & \multicolumn{3}{c}{\textbf{Post-hoc stratification}} \\
\cmidrule(lr){5-8} \cmidrule(lr){9-11}
\textbf{Label} & \shortstack[l]{\textbf{Confounder}\\($\eta^2$)} & \textbf{FM}
  & \shortstack{\textbf{Peak}\\$\boldsymbol{\alpha}$} & \textbf{S2} & $D_{\text{quantile}}^{\text{task}}$ & $P_{\text{transfer}}$ & \textbf{Indet.}
  & \textbf{AUROC} & \shortstack{\textbf{Worst}\\\textbf{stratum}} & $\boldsymbol{\Delta}$ \\
\midrule
\multirow{5}{*}{TP53} & \multirow{5}{*}{\shortstack[l]{ER status\\(0.320)\\worst: ER$-$}}
  & TITAN        & 0.6 & 10 & 96 & 10 & 90 & 0.829 & 0.646 & 0.183 \\
 & 
  & CONCH        & 0.6 & 68 & 96 & 70 & 30 & 0.788 & 0.657 & 0.131 \\
 & 
  & UNI          & --- & 0 & 98 & 0 & 100 & 0.790 & 0.641 & 0.149 \\
 & 
  & H-Optimus-0  & 0.6 & 28 & 94 & 30 & 70 & 0.827 & 0.687 & 0.140 \\
 & 
  & OpenMidnight & 0.0 & 2 & 78 & 2 & 98 & 0.806 & 0.688 & 0.118 \\
\midrule
\multirow{5}{*}{CDH1} & \multirow{5}{*}{\shortstack[l]{Histology\\(0.532)\\worst: lobular}}
  & TITAN        & 0.6 & 8 & 96 & 8 & 92 & 0.891 & 0.488 & 0.404 \\
 & 
  & CONCH        & 0.6 & 56 & 96 & 60 & 40 & 0.846 & 0.509 & 0.338 \\
 & 
  & UNI          & 0.6 & 8 & 98 & 10 & 90 & 0.787 & 0.538 & 0.249 \\
 & 
  & H-Optimus-0  & 0.6 & 30 & 98 & 30 & 70 & 0.829 & 0.525 & 0.305 \\
 & 
  & OpenMidnight & 0.6 & 2 & 92 & 2 & 98 & 0.822 & 0.512 & 0.310 \\
\midrule
\multirow{5}{*}{ER} & \multirow{5}{*}{\shortstack[l]{CDH1 mutation\\(0.043)\\worst: CDH1+$^\dagger$}}
  & TITAN        & 0.0 & 6 & 96 & 6 & 94 & 0.856 & 0.837 & 0.018 \\
 & 
  & CONCH        & 0.0 & 52 & 84 & 60 & 40 & 0.812 & 0.677 & 0.135 \\
 & 
  & UNI          & 0.6 & 12 & 100 & 12 & 88 & 0.810 & 0.686 & 0.124 \\
 & 
  & H-Optimus-0  & 0.0 & 54 & 96 & 56 & 44 & 0.834 & 0.704 & 0.130 \\
 & 
  & OpenMidnight & 0.0 & 30 & 86 & 30 & 70 & 0.830 & 0.639 & 0.191 \\
\midrule
\multirow{5}{*}{PR} & \multirow{5}{*}{\shortstack[l]{ER status\\(0.482)\\worst: ER$-$}}
  & TITAN        & 0.0 & 2 & 90 & 2 & 98 & 0.728 & 0.595 & 0.133 \\
 & 
  & CONCH        & 0.0 & 22 & 78 & 28 & 72 & 0.702 & 0.553 & 0.149 \\
 & 
  & UNI          & 0.0 & 2 & 100 & 2 & 98 & 0.677 & 0.563 & 0.114 \\
 & 
  & H-Optimus-0  & 0.0 & 32 & 96 & 32 & 68 & 0.699 & 0.566 & 0.134 \\
 & 
  & OpenMidnight & --- & 0 & 100 & 0 & 100 & 0.685 & 0.586 & 0.099 \\
\bottomrule
\end{tabular}
\end{table}

\subsection{Institution Effects}
\label{app:further-tcga-results:institution}

Cancer types in TCGA are contributed by different institutions with different protocols and patient demographics, so cancer-type confounding is entangled with institution, scanner, staining, and tissue processing effects. To assess this, we stratified the probe AUROC by tissue source site (TSS) instead of cancer type (pair-weighted mean over TSS strata with at least 5 positive and 5 negative patients; 50 pooled random splits; Table~\ref{tab:app-tss}). Across all five FMs, AUROC drops from 0.84--0.88 to 0.60--0.69 for TMB ($\Delta = 0.19$--$0.24$) and from 0.65--0.70 to 0.55--0.60 for age ($\Delta = 0.07$--$0.12$), and by 0.03--0.05 for TP53. Because each TSS code contributes a single cancer type, this collapse includes the cancer-type effect. Within cancer types, the probe-score extreme pool (top 10\% of held-out patients) also concentrates specific institutions beyond what label prevalence explains: summed within-type Cram\'er's $V$ exceeds a label-stratified permutation null (1{,}000 permutations; median $p$ over 50 splits) for age ($p \leq 0.002$) and TMB ($p \leq 0.012$) across all five FMs, and for TP53 in four of five FMs ($p \leq 0.04$; CONCH $p = 0.20$).

\begin{table}[h]
\centering
\small
\caption{\textbf{AUROC collapse under tissue source site stratification.} Pan-cancer probe AUROC and pair-weighted mean AUROC within TSS strata (${\geq}5$ positive and ${\geq}5$ negative patients), averaged over 50 pooled random splits.}
\label{tab:app-tss}
\begin{tabular}{lccccccccc}
\toprule
& \multicolumn{3}{c}{\textbf{TMB}} & \multicolumn{3}{c}{\textbf{TP53}} & \multicolumn{3}{c}{\textbf{Age}} \\
\cmidrule(lr){2-4} \cmidrule(lr){5-7} \cmidrule(lr){8-10}
\textbf{FM} & Pan & TSS & $\Delta$ & Pan & TSS & $\Delta$ & Pan & TSS & $\Delta$ \\
\midrule
TITAN        & 0.881 & 0.689 & 0.192 & 0.822 & 0.787 & 0.035 & 0.705 & 0.597 & 0.108 \\
CONCH        & 0.866 & 0.660 & 0.206 & 0.798 & 0.753 & 0.045 & 0.689 & 0.574 & 0.115 \\
UNI          & 0.839 & 0.633 & 0.206 & 0.775 & 0.728 & 0.047 & 0.651 & 0.576 & 0.075 \\
H-Optimus-0  & 0.845 & 0.645 & 0.200 & 0.782 & 0.740 & 0.043 & 0.668 & 0.585 & 0.082 \\
OpenMidnight & 0.841 & 0.604 & 0.238 & 0.780 & 0.746 & 0.034 & 0.647 & 0.554 & 0.094 \\
\bottomrule
\end{tabular}
\end{table}

\subsection{Indeterminate Outcomes on TCGA}
\label{app:further-tcga-results:indeterminate}

Table~\ref{tab:app-indet} reports how often DECAT returns indeterminate on TCGA. On pooled random splits, DECAT returns indeterminate in 4--5\% of cases for entangled models, but this does not indicate reliability: 92--94\% of their S2B tasks are misclassified as S1 (Figure~\ref{fig:tcga_fscr}). For factorized models, indeterminate rates are higher (JIVE 24\%, DSSL 51--69\%), but fewer S2B tasks are misclassified as S1 (19--21\% for DSSL at $\beta_{\text{ssl}} \geq 1.0$). On S3 (random-label) tasks, indeterminate rates are 4--6\% for all models. Among unimodal FMs on pooled splits, DECAT returns no indeterminate outcomes for CONCH and TITAN, compared with 35--37\% for H-Optimus-0, UNI, and OpenMidnight.

Indeterminate indicates that signal is present but cannot be confidently assigned. DECAT returns all metric values alongside the classification, allowing practitioners to diagnose the source: $P_{\text{transfer}}$ below its null suggests insufficient replication, unresolved $\Delta_{\text{shared}}$ suggests inconclusive localization, and a significant $D_{\text{quantile}}^{\text{task}}$ without a significant $P_{\text{transfer}}$ indicates instability that cannot be distinguished from a weak probe. In all cases, indeterminate is preferable to a confident misclassification.

\begin{table}[h]
\centering
\small
\caption{\textbf{Indeterminate rates on TCGA (\%).} Multimodal models on pooled random splits, by ground-truth task type (H\&E and RNA modalities pooled).}
\label{tab:app-indet}
\begin{tabular}{lccccc}
\toprule
\textbf{Model} & S1 & S2B & S3 & S4r & All \\
\midrule
CCA                     & 0.5  & 0.5  & 4.0 & 13.8 & 4.8 \\
CLIP                    & 1.6  & 1.6  & 4.8 & 8.9  & 4.0 \\
JIVE                    & 23.1 & 22.1 & 3.7 & 30.1 & 23.7 \\
DSSL ($\beta{=}0.01$)   & 53.9 & 53.6 & 5.2 & 69.9 & 55.7 \\
DSSL ($\beta{=}0.1$)    & 67.7 & 68.2 & 5.6 & 59.1 & 61.4 \\
DSSL ($\beta{=}1.0$)    & 71.2 & 70.3 & 5.1 & 79.3 & 69.3 \\
DSSL ($\beta{=}10$)     & 43.5 & 43.4 & 5.1 & 75.7 & 51.0 \\
\bottomrule
\end{tabular}
\end{table}

\clearpage
\section{Compute and Reproducibility}
\label{app:compute}

\paragraph{Hardware.}
All experiments were run on a single node with 8$\times$NVIDIA A10 GPUs, 178 CPU cores, and 740~GB RAM.

\paragraph{Compute budget.}
Main experiment total wall time is approximately 22 hours on the above hardware. Representation training ($\sim$15h) comprises 2{,}555 representation trainings (365 runs $\times$ 7 model configurations) parallelized across 8 GPUs. Closed-form models (CCA, JIVE) contribute negligible time and the majority is CLIP and DisentangledSSL training. Evaluation ($\sim$7h) comprises 294{,}525 evaluation calls (589{,}050 modality-task records) via a CPU process pool with up to 100 concurrent workers.

\paragraph{Code and data availability.}
The DECAT implementation, experiment scripts, and notebooks reproducing all figures and tables are available at \url{https://github.com/dst53/DECAT} under the Apache License 2.0. All synthetic data used in the synthetic validation experiments is regenerable from seeds. For the TCGA validation experiments, publicly available whole-slide images and bulk RNA-seq data were downloaded from the Genomic Data Commons (GDC). Somatic mutation labels were derived from the MC3 consensus mutation callset (mc3.v0.2.8.PUBLIC.maf.gz).